\documentclass[10pt,journal,compsoc]{IEEEtran}
\IEEEoverridecommandlockouts
\usepackage{mydef}
\usepackage{cite}
\usepackage{amsmath,amssymb,amsfonts,bm}
\usepackage{algorithmic}
\usepackage{graphicx}
\usepackage{balance}
\usepackage{enumitem}
\usepackage{textcomp}
\usepackage{bbding} 
\usepackage{xcolor}
\usepackage{threeparttable}
\usepackage{algorithm}
\usepackage{multirow}
\usepackage{url}
\usepackage{svg}
\usepackage{booktabs}
\usepackage{array}
\usepackage{pifont}
\usepackage{caption}
\usepackage{colortbl}
\usepackage{balance}
\usepackage{float,color}
\usepackage[colorlinks, linkcolor=blue, citecolor=blue, urlcolor=blue]{hyperref}

\def\BibTeX{{\rm B\kern-.05em{\sc i\kern-.025em b}\kern-.08em
    T\kern-.1667em\lower.7ex\hbox{E}\kern-.125emX}}
\begin{document}
\def\method{}
\newcommand{\R}[1]{{\color{black}{\sc} #1}}
\newcommand{\paratitle}[1]{\vspace{0.5ex}\noindent\textbf{\textit{#1}}}

\def \x {\mathbf{x}}
\def \X {{\mathbf{X}}}
\def \Y {\mathbf{Y}}
\def \y {\mathbf{y}}
\def \Z {\mathbf{Z}} \def \z {\mathbf{z}}

\def \ca {{\mathcal A}}
\def \cb {{\mathcal B}} \def \cd {{\mathcal D}} \def \cf {{\mathcal F}}
\def \cw {{\mathcal W}} \def \ck {{\mathcal K}} \def \cq {{\mathcal q}}
\def \cx {{\mathcal X}} \def \cm {{\mathcal M}} \def \cn {{\mathcal N}}
\def \cl {{\mathcal L}} \def \cR {{\mathcal R}} \def \cy {{\mathcal Y}}
\def \cu {{\mathcal U}} \def \ch {{\mathcal H}} \def \cp {{\mathcal P}}
\def \co {{\mathcal O}} \def \cs {{\mathcal S}} \def \ce {{\mathcal E}}
\def \cz {{\mathcal Z}} \def \cc {{\mathcal C}} \def \cg {{\mathcal G}}
\def \ct {{\mathcal T}} \def \ci {{\mathcal I}}
\def \proof {\noindent \emph{Proof.}\ \ }
\def \cv {{\mathcal V}}

\def\eg{\emph{e.g}.} \def\Eg{\emph{E.g}.}
\def\ie{\emph{i.e}.} \def\Ie{\emph{I.e}.}
\def\cf{\emph{c.f}.} \def\Cf{\emph{C.f}.}
\def\st{\emph{s.t}.} \def\St{\emph{S.t}.}
\def\etc{\emph{etc}.} \def\vs{\emph{vs}.}
\def\wrt{w.r.t.} \def\dof{d.o.f.}
\def\etal{\emph{et al}.}

%
\title{Large Language Model Agent: A Survey on Methodology, Applications and Challenges}
%
%
%
%

\author{
Junyu Luo, Weizhi Zhang, Ye Yuan, Yusheng Zhao, Junwei Yang, Yiyang Gu, Bohan Wu, Binqi Chen,\\ Ziyue Qiao, Qingqing Long, Rongcheng Tu, Xiao Luo, Wei Ju, Zhiping Xiao, Yifan Wang, Meng Xiao, \\ Chenwu Liu,  Jingyang Yuan, Shichang Zhang, Yiqiao Jin, Fan Zhang, Xian Wu,  Hanqing Zhao,\\
Dacheng Tao,~\IEEEmembership{Fellow,~IEEE}, 
Philip S. Yu,~\IEEEmembership{Fellow,~IEEE} and Ming Zhang

\IEEEcompsocitemizethanks{
\IEEEcompsocthanksitem Junyu Luo, Ye Yuan, Yusheng Zhao, Junwei Yang, Yiyang Gu, Bohan Wu, Binqi Chen, Wei Ju, Chenwu Liu, Jingyang Yuan, and Ming Zhang are with the School of Computer Science and PKU-Anker LLM Lab, Peking University, Beijing, China. (e-mail: luojunyu@stu.pku.edu.cn, mzhang\_cs@pku.edu.cn)
\IEEEcompsocthanksitem Weizhi Zhang and P.S. Yu are with the Department of Computer Science, University of Illinois at Chicago, Chicago, USA. 
\IEEEcompsocthanksitem Ziyue Qiao is with the School of Computing and Information Technology, Great Bay University, Guangdong, China. 
\IEEEcompsocthanksitem Qingqing Long and Meng Xiao are with the Computer Network Information Center, Chinese Academy of Sciences, Beijing, China. 
\IEEEcompsocthanksitem Rongcheng Tu, Hanqing Zhao, and Dacheng Tao are with Nanyang Technological University, Singapore. 
\IEEEcompsocthanksitem Xiao Luo is with the Department of Computer Science, University of California, Los Angeles, USA. 
\IEEEcompsocthanksitem Zhiping Xiao is with Paul G. Allen School of Computer Science and Engineering, University of Washington, Seattle, USA. 
\IEEEcompsocthanksitem Yifan Wang is with the School of Information Technology $\&$ Management, University of International Business and Economics, Beijing, China. 
\IEEEcompsocthanksitem Shichang Zhang is with Harvard University, Cambridge, USA. 
\IEEEcompsocthanksitem Yiqiao Jin is with Georgia Institute of Technology, Atlanta, USA.
\IEEEcompsocthanksitem Fan Zhang and Xian Wu are with Jarvis Research Center, Tencent YouTu Lab, Shenzhen, China. 
}
}

\IEEEtitleabstractindextext{%

\begin{abstract}
The era of intelligent agents is upon us, driven by revolutionary advancements in large language models. Large Language Model (LLM) agents, with goal-driven behaviors and dynamic adaptation capabilities, potentially represent a critical pathway toward artificial general intelligence. This survey systematically deconstructs LLM agent systems through a methodology-centered taxonomy, linking architectural foundations, collaboration mechanisms, and evolutionary pathways. We unify fragmented research threads by revealing fundamental connections between agent design principles and their emergent behaviors in complex environments. Our work provides a unified architectural perspective, examining how agents are constructed, how they collaborate, and how they evolve over time, while also addressing evaluation methodologies, tool applications, practical challenges, and diverse application domains. By surveying the latest developments in this rapidly evolving field, we offer researchers a structured taxonomy for understanding LLM agents and identify promising directions for future research. The collection is available at \url{https://github.com/luo-junyu/Awesome-Agent-Papers}.
\end{abstract}

}

\maketitle
\section{Introduction}
\IEEEPARstart{A}{}rtificial Intelligence is entering a pivotal era with the emergence of LLM agents—intelligent entities powered by large language models (LLMs) capable of perceiving environments, reasoning about goals, and executing actions~\cite{xi2025rise}. Unlike traditional AI systems that merely respond to user inputs, modern LLM agents actively engage with their environments through continuous learning, reasoning, and adaptation. This shift represents a technological advancement and a fundamental reimagining of human-machine relationships. Commercial LLM agent systems (\eg, DeepResearch, DeepSearch, and Manus) exemplify this paradigm shift—autonomously executing complex tasks that once required human expertise, from in-depth research to computer operation, while adapting to specific user needs.

Compared to traditional agent systems~\cite{wooldridge1995intelligent}, LLM-based agents have achieved generational across multiple dimensions, including knowledge sources~\cite{zheng2024large}, generalization capabilities~\cite{lotfi2023non}, and interaction modalities~\cite{fei2024multimodal}. Today's agents represent a qualitative leap driven by the convergence of three key developments: \ding{182} unprecedented reasoning capabilities of LLMs~\cite{huang2022towards}, \ding{183} advancements in tool manipulation and environmental interaction~\cite{wang2024tool}, and \ding{184} sophisticated memory architectures that support longitudinal experience accumulation~\cite{zhang2024survey,zhao2023depth}. This convergence has transformed theoretical constructs into practical systems, increasingly blurring the boundary between assistants and collaborators. This shift fundamentally arises from LLMs' role as \textit{general-purpose task processors}, unifying perception, decision-making, and action within semantic space through generative architectures, thereby forming human-like cognitive loops~\cite{sumers2023cognitive}.

Our study presents a novel examination of agent systems through a unified taxonomy that connects agent construction, collaboration mechanisms, and evolutionary pathways. We offer a comprehensive perspective tracing on how agents are defined, how they function individually or collectively, and how they evolve over time. Beyond clarifying the current landscape, our work not only clarifies the current landscape but identifies emerging patterns that signal future developments. The rapid advancement of agent technologies necessitates timely surveys to provide researchers with an up-to-date taxonomy for understanding this dynamic field.

\begin{figure*}[ht]
\centering
\includegraphics[width=\textwidth]{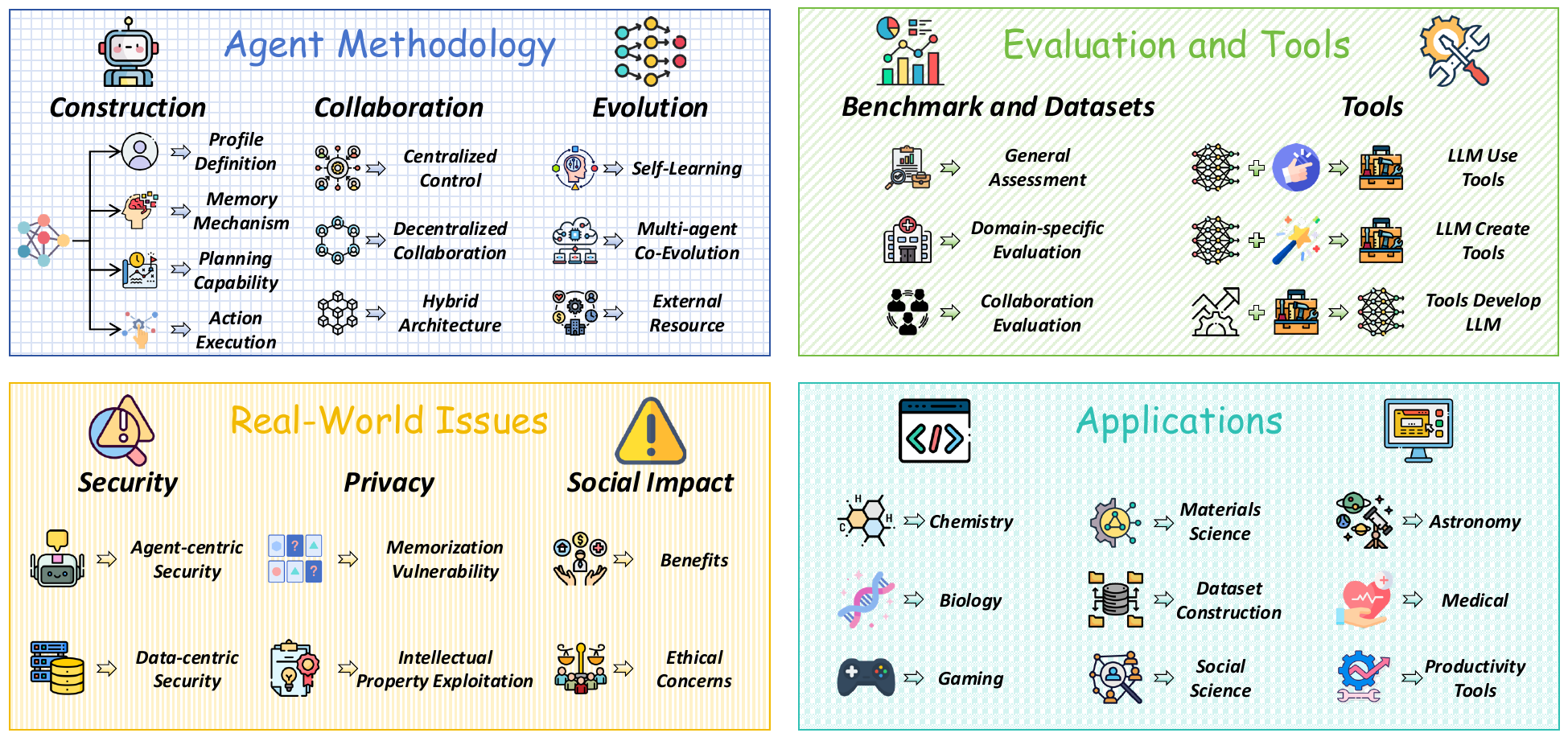}
\caption{An overview of the LLM agent ecosystem organized into four interconnected dimensions: \ding{182} Agent Methodology, covering the foundational aspects of construction, collaboration, and evolution; \ding{183} Evaluation and Tools, presenting benchmarks, assessment frameworks, and development tools; \ding{184} Real-World Issues, addressing critical concerns around security, privacy, and social impact; and \ding{185} Applications, highlighting diverse domains where LLM agents are being deployed. We provide a structured framework for understanding the complete lifecycle of modern LLM-based agent systems.}
\label{fig:overview}
\end{figure*}

Figure~\ref{fig:overview} presents our organizational framework for understanding the LLM agent ecosystem. At its core, our methodology-centered approach examines the technical foundations of agent systems through three interconnected dimensions: construction (how agents are defined and built), collaboration (how they interact and work together), and evolution (how they learn and improve over time). This tripartite foundation is complemented by practical considerations, including evaluation methodologies, development tools, real-world challenges related to security and ethics, and diverse application domains. This framework shapes the structure of our survey, enabling a systematic exploration of each dimension while highlighting their interconnections.

\paratitle{Distinction from Previous Surveys.}
Despite several surveys exploring various aspects of AI agents in recent years, our study makes a distinctive contribution through its methodological focus and comprehensive analysis of LLM agent architectures. Previous surveys have primarily focused on specific applications (\eg, gaming~\cite{hu2024survey, xu2024survey}), deployment environments~\cite{xu2024unleashing, qu2025mobile}, multi-modality~\cite{durante2024agent} or security~\cite{wang2024large}, while others have provided broad overviews without a detailed methodological taxonomy~\cite{xi2025rise, wang2024survey}. Recent works also have examined LLM-based agents compared to traditional AI agents~\cite{zhao2023depth}, multi-agent interaction~\cite{li2024survey}, workflows~\cite{li2024review}, and cooperative decision-making mechanisms~\cite{jin2025comprehensive}. In contrast to these works, our survey stands out through:
\begin{enumerate}
    \item \textbf{Methodology-centered taxonomy:} We propose a systematic taxonomy that deconstructs LLM agent systems into their fundamental methodological components, including role definition, memory mechanisms, planning capabilities, and action execution~\cite{ma2024survey}.
    
    \item \textbf{Build-Collaborate-Evolve framework:} We analyze three interconnected dimensions of LLM agents - construction, collaboration, and evolution - offering a more holistic understanding than previous approaches~\cite{guo2024large, masterman2024landscape}. This integrated architectural perspective highlights the continuity between individual LLM agent design and collaborative systems, whereas prior studies have often examined these aspects separately~\cite{cheng2024exploring, guo2024large}.
    
    \item \textbf{Frontier applications and real-world focus:} Beyond addressing theoretical concepts, our work examines cutting-edge tools, communication protocols, and diverse applications on LLM agents. We provide comprehensive analysis of pressing real-world challenges including security, privacy, and ethics. This forward-looking perspective is particularly valuable as agent technologies transition from research to widespread implementation.
\end{enumerate}

Our survey provides researchers and practitioners with a more structured taxonomy for understanding, comparing, and advancing research of LLM agents from different perspectives.
As LLM agent systems increasingly integrate into various critical domains, understanding their architectural foundations becomes essential not only for researchers but also for policy scholars, industry practitioners, and society at large. This survey aims to provide this foundation while charting a path forward for this rapidly evolving field.


\definecolor{softblue}{RGB}{220,230,242}    
\definecolor{softgreen}{RGB}{226,239,218}   
\definecolor{softpurple}{RGB}{229,224,236}  
\definecolor{softyellow}{RGB}{255,242,204}  
\definecolor{softred}{RGB}{242,220,219}     
\definecolor{softgray}{RGB}{240,240,240}     
\definecolor{softgold}{RGB}{235,190,115}     

\tikzstyle{leaf}=[draw=black, 
    rounded corners,minimum height=1em,
    text width=22.50em, edge=black!10, 
    text opacity=1, 
    align=left,
    fill opacity=.3,  text=black,font=\scriptsize,
    inner xsep=5pt, inner ysep=3pt,
    ]
\tikzstyle{leaf1}=[draw=black, 
    rounded corners,minimum height=1em,
    text width=6.5em, edge=black!10, 
    text opacity=1, align=center,
    fill opacity=.5,  text=black,font=\scriptsize,
    inner xsep=3pt, inner ysep=3pt,
    ]
\tikzstyle{leaf2}=[draw=black, 
    rounded corners,minimum height=1em,
    text width=6.5em, edge=black!10, 
    text opacity=1, align=center,
    fill opacity=.8,  text=black,font=\scriptsize,
    inner xsep=3pt, inner ysep=3pt,
    ]
\tikzstyle{leaf3}=[draw=black, 
    rounded corners,minimum height=1em,
    text width=6em, edge=black!10, 
    text opacity=1, align=center,
    fill opacity=.8,  text=black,font=\scriptsize,
    inner xsep=3pt, inner ysep=3pt,
]
\begin{figure*}[t]
\centering
\begin{forest}
  for tree={
  forked edges,
  grow=east,
  reversed=true,
  anchor=center,
  parent anchor=east,
  child anchor=west,
  base=center,
  font=\scriptsize,
  rectangle,
  draw=black, 
  edge=black!50, 
  rounded corners,
  minimum width=2em,
  s sep=5pt,
  inner xsep=3pt,
  inner ysep=1pt
  },
  where level=1{text width=4.5em}{},
  where level=2{text width=6em,font=\scriptsize}{},
  where level=3{font=\scriptsize}{},
  where level=4{font=\scriptsize}{},
  where level=5{font=\scriptsize}{},
  [Large Language Model Agent,rotate=90,anchor=north,inner xsep=8pt,inner ysep=3pt,edge=black!50,draw=black
    [Profile Definition \\ \S \ref{sec:profile-def}, edge=black!50, leaf3,
      [Human-Curated Static Profiles , leaf1, fill=softgreen,
          [Camel\cite{li2023camel}{, }AutoGen\cite{wu2023autogen}{, }MetaGPT\cite{hong2024metagpt}{, }ChatDev\cite{qian2024chatdev}{, }AFlow\cite{zhang2025aflow},
            ,leaf,fill=softgreen]
      ]
      [Betch-Generated Dynamic Profiles, leaf1, fill=softgreen,
        [Generative Agents\cite{park2023generative}{, }RecAgent\cite{wang2025user}{, }DSPy\cite{khattab2024dspy},leaf,fill=softgreen]
      ]
    ]
    [Memory Mechanism \\ \S \ref{sec:mem-mechanism}, edge=black!50, leaf3, 
      [Short-Term Memory, leaf1, fill=myyellow,
        [ReAct \cite{yao2023react}{, }ChatDev \cite{qian2024chatdev}{, }Graph of Thoughts \cite{besta2024graphofthoughts}{, }AFlow \cite{zhang2025aflow},leaf,fill=myyellow]
      ]
      [Long-Term Memory, leaf1, fill=myyellow,
        [Voyager \cite{wang2023voyager}{, }GITM \cite{zhu2023ghost}{, }ExpeL \cite{zhao2024expel}{, }Reflexion \cite{shinn2023reflexion}{, }TPTU \cite{ruan2023tptu}{, }OpenAgents \cite{xie2023openagents}{, }Lego-Prover \cite{wang2024lego}{, }MemGPT \cite{packer2023memgpt} ,leaf,fill=myyellow]
      ]
      [Knowledge Retrieval as Memory, leaf1, fill=myyellow,
        [RAG \cite{lewis2020retrieval}{, }GraphRAG \cite{edge2024graphrag}{, }Chain of Agnets \cite{zhang2024chainofagents}{, }IRCoT \cite{trivedi2022interleaving}{, }Llatrieval \cite{li2024llatrieval}{, }KG-RAR \cite{wu2025graphaugreasoning}{, }DeepRAG \cite{guan2025deeprag},leaf,fill=myyellow]
      ]
    ]
    [Planning Capability \\ \S \ref{sec:plan-capability}, edge=black!50, leaf3, 
      [Task Decomposition Strategies, leaf1, fill=softblue,
        [Plan-and-solve Prompting\cite{wang2023plan}{, }Distributed Problem Solving and Planning \cite{durfee2001distributed}{, }ReAct \cite{yao2023react}{, }Chain-of-discussion \cite{tao2024chain}{, }Tree-planner \cite{hu2023tree}{, }ReAcTree \cite{choireactree}{, }ToT \cite{long2023large}{, }ReST-MCTS* \cite{zhang2024rest}{, }LLM-MARS \cite{lykov2023llm}{, }LLM as BT-planner \cite{ao2024llm}{, }ConceptAgent \cite{rivera2024conceptagent} ,leaf,fill=softblue]
      ]
      [Feedback-Driven Iteration, leaf1, fill=softblue,
        [BrainBody-LLM \cite{bhat2024grounding}{, }TrainerAgent \cite{li2023traineragent}{, }RASC \cite{wan2024dynamic}{, }REVECA \cite{seo2024llm}{, }AdaPlanner \cite{sun2023adaplanner}{, }AIFP \cite{jafaripour2025adaptive} ,leaf,fill=softblue]
      ]
    ]
    [Action Execution \\ \S \ref{sec:action-exec}, edge=black!50, leaf3, 
      [Tool Utilization, leaf1, fill=softgray,
        [TRICE \cite{qiao2023making}{, }GPT4Tools \cite{yang2023gpt4tools}{, }EASYTOOL\cite{yuan2024easytool}{, }AvaTaR \cite{wu2025avatar}{, },leaf,fill=softgray]
      ]
      [Physical Interaction, leaf1, fill=softgray,
        [DriVLMe \cite{huang2024drivlme}{, }ReAd \cite{zhang2024towards}{, }Collaborative Voyager \cite{colle2024improving} ,leaf,fill=softgray]
      ]
    ]
    [Agent Collaboration \\ \S \ref{sec:agent-collab}, edge=black!50, leaf3, 
      [Centralized Control \\ \S \ref{sec:central-control}, leaf1, fill=softyellow,
        [Coscientist \cite{boiko2023autonomous}{, }LLM-Blender \cite{jiang2023llmlingua}{, }MetaGPT \cite{hong2024metagpt}{, }AutoAct \cite{qiao2024autoact}{, }Meta-Prompting \cite{suzgun2024meta}{, }Wjudge \cite{khan2024debating} ,leaf,fill=softyellow]
      ]
      [Decentralized Collaboration \\ \S \ref{sec:decentral-collab}, leaf1, fill=softyellow,
        [GAgents \cite{park2023generative}{, }CAMEL \cite{li2023camel}{, }MedAgents \cite{tang2023medagents}{, }ReConcile \cite{chen2023reconcile}{, }MAD \cite{liang2023encouraging}{, }MADR \cite{kim2024can}{, }MDebate \cite{du2023improving}{, }AutoGen \cite{wu2023autogen} ,leaf,fill=softyellow]
      ]
      [Hybrid Architecture \\ \S \ref{sec:hybrid-arch}, leaf1, fill=softyellow,
        [KnowAgent \cite{zhu2024knowagent}{, }WKM \cite{qiao2024agent}{, }Textgrad \cite{fang2024refining} ,leaf,fill=softyellow]
      ]
    ]
    [Agent Evolution \\ \S \ref{sec:agent-evol}, edge=black!50, leaf3, 
      [Autonomous Optimization and Self-Learning \\ \S \ref{sec:auto-opt-self-learn}, leaf1, fill=softred,
        [SE \cite{zhong2023self}{, }Evolutionary Optimization \cite{akiba2025evolutionary}{, }DiverseEvol \cite{wu2023self}{, }SELF-REFINE \cite{madaan2023self}{, }STaR~\cite{zelikman2024star}{, }V-STaR~\cite{hosseiniv}{, }Self-Verification~\cite{weng2023large}{, }Self-Rewarding~\cite{yuan2024selfrewardinglanguagemodels}{, }RLCD~\cite{yang2024rlcd}{, }RLC \cite{panglanguage} ,leaf,fill=softred]
      ]
      [Multi-Agent Co-Evolution \\ \S \ref{sec:multi-agent-co-evol}, leaf1, fill=softred,
        [ProAgent \cite{zhang2024proagent}{, }CORY \cite{ma2024coevolving}{, }CAMEL~\cite{li2023camel}{, }Red-Team LLMs \cite{ma2023evolving}{, }Multi-Agent Debate \cite{du2023improving}{, }MAD \cite{liang2024encouraging} ,leaf,fill=softred]
      ]
      [Evolution via External Resources \\ \S \ref{sec:evol-external-resource}, leaf1, fill=softred,
        [KnowAgent \cite{zhu2024knowagent}{, }WKM \cite{qiao2024agent}{, }CRITIC \cite{goucritic}{, }STE~\cite{song2024trial}{, }SelfEvolve \cite{jiang2023selfevolve} ,leaf,fill=softred]
      ]
    ]
    ]
  ]
\end{forest}
\caption{A taxonomy of large language model agent methodologies.}
\label{fig:taxonomy_data_efficient_llm_post_training}
\end{figure*}
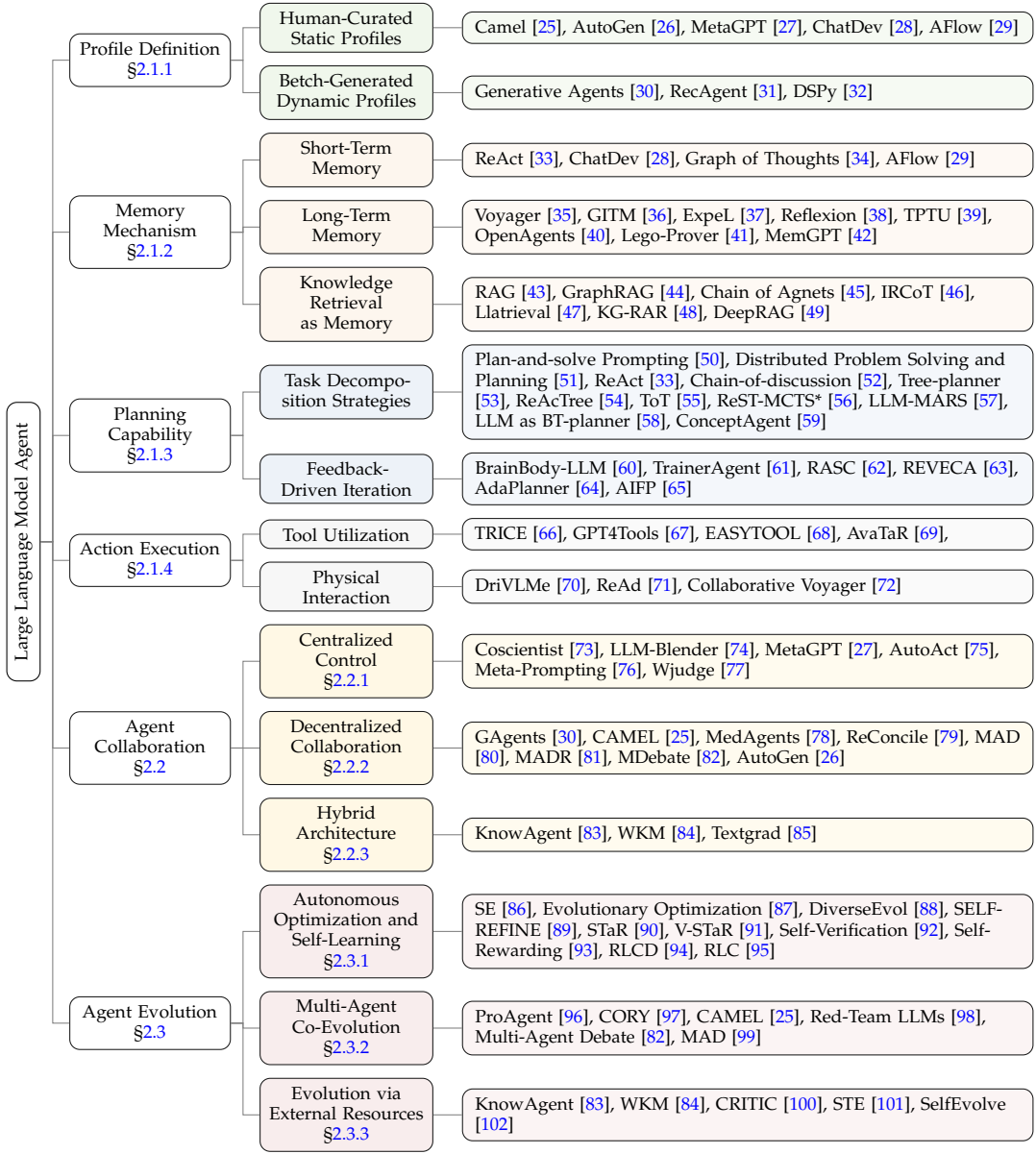


\section{Agent Methodology}\label{sec:method}

This section presents a comprehensive framework for understanding LLM-based agent systems through three interconnected dimensions: construction, collaboration, and evolution. As illustrated in Figure~\ref{fig:taxonomy_data_efficient_llm_post_training}, we first examine agent construction~(Section~\ref{sec:agent-construction}), which establishes the foundational components including profile definition, memory mechanisms, planning capabilities, and action execution. We then explore collaboration paradigms~(Section~\ref{sec:agent-collab}) that enable multiple agents to work together through centralized control, decentralized cooperation, or hybrid architectures. Finally, we investigate evolution mechanisms~(Section~\ref{sec:agent-evol}) that allow agents to improve over time through autonomous optimization, multi-agent co-evolution, and external resource integration. This three-dimensional framework provides a systematic approach to analyzing the full lifecycle of LLM agent systems.

\subsection{Agent Construction}\label{sec:agent-construction}

Agent construction serves as the foundational phase in developing LLM-based autonomous systems, encompassing the systematic design of core components that enable goal-directed behaviors. This process prioritizes four interdependent pillars: profile definition (\ref{sec:profile-def}), memory mechanism (\ref{sec:mem-mechanism}), planning capability (\ref{sec:plan-capability}), and action execution (\ref{sec:action-exec}). These components collectively form a recursive optimization loop, where memory informs planning, execution outcomes update memory, and contextual feedback refines agent profiles. The construction paradigm emphasizes modular interoperability while preserving system-wide coherence, enabling subsequent collaboration and evolutionary adaptation mechanisms, which will be discussed in later sections.

\subsubsection{Profile Definition}
\label{sec:profile-def}
Profile definition establishes an agent's operational identity by configuring its intrinsic attributes and behavioral patterns~\cite{li2023camel,wu2023autogen}. Current methodologies encompass two approaches: \textit{human-curated static profiles} ensure domain-specific consistency through manual specification, while \textit{batch-generated dynamic profiles} adaptively modulate operational parameters to stochastically yield a batch of agent initializations. These mechanisms collectively govern an agent's decision boundaries and interaction protocols while maintaining alignment with predefined objectives.

\paratitle{Human-Curated Static Profiles.}
This approach establishes fixed agent profiles through manual specification by domain experts, embedding explicit rules and domain-specific knowledge. It ensures strict adherence to predefined behavioral guidelines and task requirements enabling standardized communication protocols among agents. This is particularly effective in scenarios demanding high interpretability and regulatory compliance. Such frameworks typically employ coordinated interactions between predefined agent components to achieve complex functionalities through structured communication patterns.
Representative implementations demonstrate two key paradigms: systems like Camel~\cite{li2023camel}, AutoGen~\cite{wu2023autogen}, and OpenAgents~\cite{xie2023openagents} orchestrate human-agent collaboration through predefined conversational roles (e.g., user proxy and assistant), enabling task execution through structured dialogues. Meanwhile, frameworks such as MetaGPT~\cite{hong2024metagpt}, ChatDev~\cite{qian2024chatdev}, and AFlow~\cite{zhang2025aflow} showcase role-based coordination patterns. ChatDev specializes in code development by coordinating static technical roles (e.g., product managers and programmers) with deterministic interaction protocols, while MetaGPT and AFlow extend this paradigm to general task solving through structured role orchestration.

\paratitle{Batch-Generated Dynamic Profiles.}
This paradigm employs parameterized initialization to systematically generate diverse agent profiles that emulate human societal behaviors. By injecting controlled variations into personality traits, knowledge backgrounds, or value systems during agent creation (e.g., through template-based prompting or latent space sampling), the framework produces heterogeneous populations capable of exhibiting complex social dynamics. Such parameter-driven diversity is essential for simulating realistic human-agent interactions in applications ranging from social behavior studies to emergent group intelligence simulations. This is demonstrated in systems for human behavior simulation~\cite{park2023generative} and simulated user data collection~\cite{wang2025user} where different profile configurations directly shape collective interaction patterns. Moreover, DSPy~\cite{khattab2024dspy} can further optimize the parameters of the agent profile initialization.

\subsubsection{Memory Mechanism}
\label{sec:mem-mechanism}
Memory mechanisms equip agents with the ability to store, organize, and retrieve information across temporal dimensions. Short-term memory maintains transient contextual data for immediate task execution, while long-term memory preserves structured experiential knowledge for persistent reference. Integrating knowledge retrieval mechanisms further optimizes information accessibility with Retrieval-Augmented Generation~(RAG) techniques~\cite{lewis2020retrieval}.

\paratitle{Short-Term Memory.}
Short-term memory retains agent-internal dialog histories and environmental feedback to support context-sensitive task execution. This mechanism is widely implemented in frameworks such as ReAct~\cite{yao2023react} for thinking with reflection, ChatDev~\cite{qian2024chatdev} for software development, Graph of Thoughts~\cite{besta2024graphofthoughts} for solving elaborate problems, and AFlow~\cite{zhang2025aflow} for workflow automation, demonstrating its versatility across domains. While this mechanism enables detailed reasoning through interactive exchanges, its transient nature limits knowledge retention beyond immediate contexts—intermediate reasoning traces often dissipate after task completion and cannot be directly transferred to new scenarios. Furthermore, due to LLMs' context window limitations, practical implementations require active information compression (e.g., summarization or selective retention) and impose many constraints on multi-turn interaction depth to prevent performance degradation.

\paratitle{Long-Term Memory.}
Long-term memory systematically archives agents' intermediate reasoning trajectories and synthesizes them into reusable tools for future invocation. This process transforms ephemeral cognitive efforts into persistent operational assets through three dominant paradigms: \ding{182} skill libraries that codify procedural knowledge (e.g., Voyager's automated skill discovery in Minecraft~\cite{wang2023voyager} and GITM's text-based knowledge base~\cite{zhu2023ghost}), \ding{183} experience repositories that store success/failure patterns (e.g., ExpeL's distilled experience pool~\cite{zhao2024expel} and Reflexion's trial-optimized memory~\cite{shinn2023reflexion}), and \ding{184} tool synthesis frameworks that evolve capabilities through combinatorial adaptation (e.g., TPTU’s adaptive tool composition~\cite{ruan2023tptu} and OpenAgents' self-expanding toolkit~\cite{xie2023openagents}). Cross-domain implementations, such as Lego-Prover's theorem bank~\cite{wang2024lego} and MemGPT's tiered memory architecture~\cite{packer2023memgpt}, further demonstrate how structured long-term storage enhances reasoning efficiency through strategic knowledge reuse.

\paratitle{Knowledge Retrieval as Memory.}
This paradigm diverges from agent-internal memory generation by integrating external knowledge repositories into generation processes, effectively expanding agents' accessible information boundaries. Current implementations exhibit three dominant approaches: \ding{182} Static knowledge grounding through text corpora (RAG~\cite{lewis2020retrieval}) or structured knowledge graphs (GraphRAG~\cite{edge2024graphrag}), \ding{183} Interactive retrieval that integrates agent dialogues with external queries, as demonstrated in Chain of Agents~\cite{zhang2024chainofagents} where short-term inter-agent communications trigger contextualized knowledge fetching, and \ding{184} Reasoning-integrated retrieval, exemplified by IRCoT~\cite{trivedi2022interleaving} and Llatrieval~\cite{li2024llatrieval}, which interleave step-by-step reasoning with dynamic knowledge acquisition. Advanced variants like KG-RAR~\cite{wu2025graphaugreasoning} further construct task-specific subgraphs during reasoning, while DeepRAG~\cite{guan2025deeprag} introduces fine-tuned retrieval decision modules to balance parametric knowledge and external evidence. These hybrid architectures enable agents to transcend training data limitations while maintaining contextual relevance, establishing knowledge retrieval as critical infrastructure for scalable memory systems.

\subsubsection{Planning Capability}
\label{sec:plan-capability}
Planning capabilities are a critical aspect of LLM agents' abilities, enabling them to navigate through complex tasks and problem-solving scenarios with high accuracy~\cite{huang2024understanding}. Effective planning is essential for deploying LLM agents in real-world applications, where they must handle a diverse range of complex tasks and scenarios.
The planning capability of an LLM agent can be viewed from two perspectives: task decomposition and feedback-driven iteration.

\paratitle{Task Decomposition Strategies.}
Task decomposition represents a basic approach to enhancing LLM planning capabilities by breaking down complex problems into more manageable subtasks. Although solving an entire problem may be challenging for LLM agents, they can more easily handle subtasks and then integrate the results to address the full problem. Task decomposition strategies fall into two main categories: single-path chaining and multi-path tree expansion.

Single-path chaining is a simple method with the simplist version as zero-shot chain-of-thought~\cite{kojima2022large, wei2022chain}. It first asks the agent to devise a plan, which consists of a sequence of subtasks that are built upon one another. Subsequently, the agent is asked to solve the subtasks in the order they are presented~\cite{wei2022chain, wang2023plan}. This plan-and-solve paradigm~\cite{durfee2001distributed} is straightforward and easy to implement. However, it may suffer from a lack of flexibility and error accumulation during chaining, as the agent is required to follow the pre-defined plan without any deviation during the problem-solving procedure. Therefore, one line of work proposes to adopt dynamic planning that only generates the next subtask based on the current situation of the agent~\cite{wei2022chain, yao2023react}. This enables the agent to receive environmental feedback and adjust its plan accordingly, enhancing its robustness and adaptability. Moreover, another line of work proposes to use multiple chain-of-thoughts to improve the robustness of the planning process. This is similar to ensemble methods, involving self-consistency~\cite{wang2022self, wan2024dynamic}, majority voting~\cite{li2024enhancing}, and agent discussion~\cite{tao2024chain} to combine multiple chains. By combining the wisdom of multiple chains, the agent can make more accurate decisions and reduce the risk of error accumulation.

A more complicated method is to use trees instead of chains as the planning data structure, where multiple possible reasoning paths exist when the agent is planning, and the agent is allowed to backtrack with information from feedback~\cite{hu2023tree, choireactree}. Long et al. \cite{long2023large} propose a tree-of-thought~(ToT) method that explores the solution space through a tree-like thought process. This allows the LLMs to backtrack to previous states, which makes it possible for the model to correct its previous mistakes, enabling applications to various complicated tasks that involve the "trial-error-correct" process. In more realistic scenarios, the agent can gather feedback from the environment or humans and dynamically adjust its reasoning path, potentially incorporating reinforcement learning~\cite{zhang2024rest, jiang2024technical}. This enables the agent to make more informed decisions in real-world applications using advanced algorithms such as Monte Carlo Tree Search~\cite{browne2012survey}, facilitating use cases in robotics~\cite{lykov2023llm, ao2024llm, rivera2024conceptagent} and game-playing~\cite{guo2024can, liu2024large}.

\paratitle{Feedback-Driven Iteration.}
Feedback-driven iteration is a crucial aspect of LLM planning capabilities, enabling the agent to learn from the feedback and enhance its performance over time. Feedback can originate from various sources, such as environmental input, human guidance, model introspection, and multi-agent collaboration.

Environmental feedback is one of the most common types of feedback in robotics~\cite{bhat2024grounding}, generated by the environment in which the embodied agent operates. 
Human feedback, another crucial type, comes from user interactions or manually labeled data prepared in advance~\cite{li2023traineragent, laleh2024survey}.
Model introspection provides an additional source of feedback, which is generated by the agent itself~\cite{wan2024dynamic}.
Multi-agent collaboration also serves as a feedback mechanism, where multiple agents work together to solve a problem and exchange insights~\cite{laleh2024survey, seo2024llm}.
These sources of feedback help evaluate the agent's performance and thus guide its planning. For instance, the agent can use feedback to update (regenerate) its plan, adjust its reasoning path, or even modify its goal. This iterative process continues until a satisfactory plan is achieved~\cite{sun2023adaplanner, jafaripour2025adaptive}.

\subsubsection{Action Execution}
\label{sec:action-exec}
With the planning capability, it is important for the LLMs to have the ability to execute the planned actions in the real world. Action execution is a critical aspect of LLM agents' abilities, as good plans are useless if the agent cannot execute them effectively. Action execution involves two aspects: tool utilization~\cite{shen2024llm}, and physical interaction~\cite{kim2024understanding}.

\textbf{Tool utilization}~\cite{shen2024llm} is an important aspect of LLM action execution, enabling a wide range of abilities such as precise calculation of numbers, up-to-date information understanding, and proficient code generation. The tool use ability involves two aspects: tool use decision and tool selection. The tool-use decision is the process of deciding whether to use a tool to solve a problem. When the agent is generating content with less confidence or facing problems related to specific tool functions, the agent should decide to use specific tools~\cite{qiao2023making, yang2023gpt4tools}. Tool selection is another important aspect of tool utilization, involving the understanding of tools and the agent's current situation~\cite{yuan2024easytool, wu2025avatar}. For example, Yuan et al.~\cite{yuan2024easytool} propose simplifying the tool documentation to better understand the available tools, enabling a more accurate selection of tools.

\textbf{Physical interaction}\cite{kim2024understanding} is a fundamental aspect of embodied LLM agents. Their ability to perform specific actions in the real world and interpret environmental feedback is crucial. When deployed in real-world settings, LLM agents must comprehend various factors to execute actions accurately. These factors include robotic hardware\cite{kim2024understanding}, social knowledge~\cite{huang2024drivlme}, and interactions with other LLM agents~\cite{zhang2024towards, colle2024improving}.
\subsection{Agent Collaboration}
\label{sec:agent-collab}

\begin{table}[t]
    \centering
    \caption{A summary of agent collaboration methods.}
    \label{table:collab_compare}
    \resizebox{0.485\textwidth}{!}{
    \begin{tabular}{l l l}
        \toprule
        \textbf{Category} & \textbf{Method} & \textbf{Key Contribution} \\ \midrule
        \multirow{6}{*}{\textbf{Centralized Control}} 
        & Coscientist~\cite{boiko2023autonomous} &    Human-centralized experimental control \\
        & LLM-Blender~\cite{jiang2023llmlingua} &     Cross-attention response fusion \\
        & MetaGPT~\cite{hong2024metagpt} & Role-specialized workflow management \\

        & AutoAct~\cite{qiao2024autoact} & Triple-agent task differentiation \\
        & Meta-Prompting~\cite{suzgun2024meta} & Meta-prompt task decomposition \\
        & WJudge~\cite{khan2024debating} & Weak-discriminator validation \\ \midrule
        
        \multirow{8}{*}{\textbf{Decentralized Collaboration}} 
        & MedAgents~\cite{tang2023medagents} & Expert voting consensus \\ 
        & ReConcile~\cite{chen2023reconcile} & Multi-agent answer refinement \\
        & METAL~\cite{li2025metal} & Domain-specific revision agents \\
        & DS-Agent~\cite{guo2024ds} & Database-driven revision \\
        & MAD~\cite{liang2023encouraging} & Structured anti-degeneration protocols \\
        & MADR~\cite{kim2024can} & Verifiable fact-checking critiques \\
        & MDebate~\cite{du2023improving} & Stubborn-collaborative consensus \\
        & AutoGen~\cite{wu2023autogen} & Group-chat iterative debates \\
        \midrule

        \multirow{6}{*}{\textbf{Hybrid Architecture}}
        & CAMEL~\cite{li2023camel} & Grouped role-play coordination \\
        & AFlow~\cite{zhang2025aflow} & Three-tier hybrid planning \\
        & EoT~\cite{yin2023exchange} & Multi-topology collaboration patterns \\ 
        & DiscoGraph~\cite{li2021learning} & Pose-aware distillation \\
        & DyLAN~\cite{liu2024dynamic} & Importance-aware topology \\
        & MDAgents~\cite{kim2024mdagents} & Complexity-aware routing \\ \bottomrule
    \end{tabular}
    }
\end{table}

Collaboration among LLM agents plays a crucial role in extending their problem-solving capabilities beyond individual reasoning.
Effective collaboration enables agents to leverage distributed intelligence, coordinate actions, and refine decisions through multi-agent interactions~\cite{wu2023autogen, ying2023inferring}.
We categorize existing collaboration paradigms into three fundamental architectures: \textit{centralized control}, \textit{decentralized cooperation}, and \textit{hybrid architectures}.
These paradigms differ in their decision hierarchies, communication topologies, and task allocation mechanisms, each offering distinct advantages for specific application scenarios.

\subsubsection{Centralized Control}
\label{sec:central-control}

Centralized control architectures employ a hierarchical coordination mechanism where a central controller organizes agent activities through task allocation and decision integration, while other sub-agents can only communicate with the controller.
This paradigm features two implementation strategies: \textit{explicit controller} systems utilize dedicated coordination modules (often implemented as separate LLM agents) to decompose tasks and assign subgoals, while \textit{differentiation-based} systems achieve centralized control by using prompts to guide the meta agent in assuming distinct sub-roles.
The centralized approach excels in mission-critical scenarios requiring strict coordination, such as industrial automation~\cite{vyas2024autonomous} and scientific research~\cite{boiko2023autonomous}.

\paratitle{Explicit Controller Systems.} Multiple related works have been developed to explicitly implenment centralized architectures.
The Coscientist~\cite{boiko2023autonomous} exemplifies the explicit controller paradigm, where a human operator serves as the central controller.
It establishes standardized scientific experimental workflows, allocates specialized agents and tools to distinct experimental phases, and maintains direct control over the final execution plan.
LLM-Blender~\cite{jiang2023llmlingua} explicitly creates a controller that employs a cross-attention encoder for pairwise comparison to identify the best responses,
and then fuses the top-ranked responses, enhancing their strengths while mitigating weaknesses.
MetaGPT~\cite{hong2024metagpt} simulates real-world software development workflows, direclty assigning specialized managers to control distinct functional roles and phases.

\paratitle{Differentiation-based Systems.}
AutoAct~\cite{qiao2024autoact} exemplifies the differentiation-based paradigm, which implicitly differentiates the meta-agent into three sub-agents—plan-agent, tool-agent, and reflect-agent—to break down the complex ScienceQA task.
Meta-Prompting~\cite{suzgun2024meta} decomposes complex tasks into domain-specific subtasks through carefully crafted meta-prompts. A single model acts as a coordinator, dynamically assigning subtasks to specialized sub-agents guided by task-oriented prompts. The centrol manager then integrates all intermediate outputs to produce the final solution.
These works predominantly employ highly capable agents as central controllers to optimize task allocation and decision aggregation. However, WJudge~\cite{khan2024debating} demonstrates that even controllers with limited discriminative power can also significantly enhance the overall performance of agent systems.

\subsubsection{Decentralized Collaboration}
\label{sec:decentral-collab}

In contrast to centralized architectures where a single control node often becomes a bottleneck due to handling all inter-agent communication, task scheduling, and contention resolution, decentralized collaboration enables direct node-to-node interaction through self-organizing protocols.
This paradigm can be further categorized into two distinct approaches: \textit{revision-based systems} and \textit{communication-based systems}.

\paratitle{Revision-based Systems.}
In this paradigm, agents only observe finalized decisions generated by peers and iteratively refine a shared output through structured editing protocols.
This approach typically produces more standardized and deterministic outcomes.
For instance, MedAgents~\cite{tang2023medagents} employs predefined domain-specific expert agents that sequentially propose and modify decisions independently, with consensus achieved through final voting.
ReConcile~\cite{chen2023reconcile} coordinates agents to iteratively refine answers through mutual response analysis, confidence evaluation, and human-curated exemplars.
METAL~\cite{li2025metal} introduces specialized text and visual revision agents for chart generation tasks, demonstrating how domain-specific refinement improves output quality.
Notably, revision signals may originate not only from agent interactions but also from external knowledge bases~\cite{guo2024ds, dell2022data}, enabling hybrid refinement strategies.

\paratitle{Communication-based Systems.}
Compared to revision-based approaches, communication-based methods feature more flexible organizational structures, allowing agents to directly engage in dialogues and observe peers' reasoning processes.
This makes them particularly suitable for modeling dynamic scenarios such as human social interactions~\cite{park2023generative}. Key implementations include:
MAD~\cite{liang2023encouraging} employs structured communication protocols to address the "degeneration-of-thought" problem, where agents overly fixate on initial solutions.
MADR~\cite{kim2024can} enhances this by enabling agents to critique implausible claims, refine arguments, and generate verifiable explanations for fact-checking.
MDebate~\cite{du2023improving} optimizes consensus-building through strategic alternation between stubborn adherence to valid points and collaborative refinement.
AutoGen~\cite{wu2023autogen} implements a group-chat framework that supports multi-agent participation in iterative debates for decision refinement.

\subsubsection{Hybrid Architecture}
\label{sec:hybrid-arch}

Hybrid architectures strategically combine centralized coordination and decentralized collaboration to balance controllability with flexibility, optimize resource utilization, and adapt to heterogeneous task requirements. This approach introduces two implementation patterns: \textit{static systems} with predefined coordination rules and \textit{dynamic systems} featuring self-optimizing topologies.

\paratitle{Static Systems.}
Static systems predefine fixed patterns for combining different collaboration modalities. Representative implementations include:
CAMEL~\cite{li2023camel} partitions agents into intra-group decentralized teams for role-playing simulations, while maintaining inter-group coordination through centralized governance.
AFlow~\cite{zhang2025aflow} employs a three-tier hierarchy consisting of centralized strategic planning, decentralized tactical negotiation, and market-driven operational resource allocation.
EoT~\cite{yin2023exchange} formalizes four collaboration patterns (\textsc{Bus}, \textsc{Star}, \textsc{Tree}, \textsc{Ring}) to align network topologies with specific task characteristics.

\paratitle{Dynamic Systems.} 
Recent innovations introduce neural topology optimizers that dynamically reconfigure collaboration structures based on real-time performance feedback, enabling automatic adaptation to changing conditions.
Key implementations demonstrate this paradigm:
DiscoGraph~\cite{li2021learning} introduces trainable pose-aware collaboration through a teacher-student framework. The teacher model with holistic-view inputs guides the student model via feature map distillation, while matrix-valued edge weights enable adaptive spatial attention across agents.
DyLAN~\cite{liu2024dynamic} first utilizes the Agent Importance Score to identify the most contributory agents and then dynamically adjusts the collaboration structure to optimize task completion.
MDAgents~\cite{kim2024mdagents} dynamically assigns collaboration structures based on the task at hand. It first performs a complexity check to classify tasks as low, moderate, or high complexity.
Simple tasks are handled by a single agent, while more complex tasks are addressed through hierarchical collaboration.

\begin{table}[t]
    \centering
    \caption{A summary of agent evolution methods.}
    \label{tab:evolution}
    \resizebox{0.485\textwidth}{!}{
    \begin{tabular}{l l l}
        \toprule
        \textbf{Category} & \textbf{Method} & \textbf{Key Contribution} \\ \midrule
        \multirow{3}{*}{\textbf{Self-Supervised Learning}} 
        & SE~\cite{zhong2023self} & Adaptive token masking for pretraining \\
        & Evolutionary Optimization~\cite{akiba2025evolutionary} & Efficient model merging and adaptation \\
        & DiverseEvol~\cite{wu2023self} & Improved instruction tuning via diverse data \\ 
        \midrule
        
        \multirow{4}{*}{\textbf{Self-Reflection \& Self-Correction}} 
        & SELF-REFINE~\cite{madaan2023self} & Iterative self-feedback for refinement \\
        & STaR~\cite{zelikman2024star} & Bootstrapping reasoning with few rationales \\
        & V-STaR~\cite{hosseiniv} & Training a verifier using DPO\\
        & Self-Verification~\cite{weng2023large} & Backward verification for correction \\
        \midrule
        
        \multirow{3}{*}{\textbf{Self-Rewarding \& RL}} 
        & Self-Rewarding~\cite{yuan2024selfrewardinglanguagemodels} & LLM-as-a-Judge for self-rewarding \\
        & RLCD~\cite{yang2024rlcd} & Contrastive distillation for alignment \\
        & RLC~\cite{panglanguage} & Evaluation-generation gap for optimization \\ \midrule
        
        \multirow{3}{*}{\textbf{Cooperative Co-Evolution}} 
        & ProAgent~\cite{zhang2024proagent} & Intent inference for teamwork \\
        & CORY~\cite{ma2024coevolving} & Multi-agent RL fine-tuning \\
        & CAMEL~\cite{li2023camel} & Role-playing framework for cooperation \\
        \midrule
        
        \multirow{3}{*}{\textbf{Competitive Co-Evolution}} 
        & Red-Team LLMs~\cite{ma2023evolving} & Adversarial robustness training \\
        & Multi-Agent Debate~\cite{du2023improving} & Iterative critique for refinement \\
        & MAD~\cite{liang2024encouraging} & Debate-driven divergent thinking \\ \midrule
        
        \multirow{2}{*}{\textbf{Knowledge-Enhanced Evolution}} 
        & KnowAgent~\cite{zhu2024knowagent} & Action knowledge for planning \\
        & WKM~\cite{qiao2024agent} & Synthesizing prior and dynamic knowledge \\
        \midrule
        
        \multirow{3}{*}{\textbf{Feedback-Driven Evolution}} 
        & CRITIC~\cite{goucritic} & Tool-assisted self-correction \\
        & STE~\cite{song2024trial} & Simulated trial-and-error for tool learning \\
        & SelfEvolve~\cite{jiang2023selfevolve} & Automated debugging and refinement \\ \bottomrule
    \end{tabular}
    }
\end{table}


\subsection{Agent Evolution}
\label{sec:agent-evol}

LLM Agents are evolving through various mechanisms that enable autonomous improvement, multi-agent interaction, and external resource integration. This section explores three key dimensions of agent evolution: autonomous optimization and self-learning, multi-agent co-evolution, and evolution via external resources. These mechanisms collectively enhance model adaptability, reasoning, and performance in complex environments. We summarize the methods in Table~\ref{tab:evolution}.

\subsubsection{Autonomous Optimization and Self-Learning}
\label{sec:auto-opt-self-learn}

Autonomous optimization and self-learning allow LLMs to improve their capabilities without extensive supervision. This includes self-supervised learning, self-reflection, self-correction, and self-rewarding mechanisms that enable models to explore, adapt, and refine their outputs dynamically.

\paratitle{Self-Supervised Learning and Adaptive Adjustment.}  
Self-supervised learning enables LLMs to improve using unlabeled or internally generated data, reducing reliance on human annotations. For example, self-evolution learning (SE)~\cite{zhong2023self} enhances pretraining by dynamically adjusting token masking and learning strategies. Evolutionary optimization techniques facilitate efficient model merging and adaptation, improving performance without extensive additional resources \cite{akiba2025evolutionary}. DiverseEvol~\cite{wu2023self} refines instruction tuning by improving data diversity and selection efficiency. 
These advancements contribute to the autonomous adaptability of LLMs, enabling more efficient learning and generalization across tasks.

\paratitle{Self-Reflection and Self-Correction.}  
Self-reflection and self-correction enable LLMs to iteratively refine their outputs by identifying and addressing errors. For instance, 
SELF-REFINE~\cite{madaan2023self} applies iterative self-feedback to improve generated responses without external supervision. In reasoning tasks, STaR~\cite{zelikman2024star} and V-STaR~\cite{hosseiniv} train models to verify and refine their own problem-solving processes, reducing reliance on labeled data. 
Additionally, self-verification techniques enable models to retrospectively assess and correct their outputs, leading to more reliable decision-making \cite{weng2023large}. 
These approaches collectively enhance LLM agents’ ability to self-reflect and self-correct, reducing hallucinations and improving reasoning quality.

\paratitle{Self-Rewarding and Reinforcement Learning.}  
Self-rewarding and reinforcement learning approaches enable LLMs to enhance performance by generating internal reward signals. 
Self-generated rewards help models refine decision-making, with techniques ensuring stable and consistent learning improvements \cite{yuan2024selfrewardinglanguagemodels}. Contrastive distillation further enables models to align themselves through self-rewarding mechanisms \cite{yang2024rlcd}. Additionally, RLC~\cite{panglanguage} leverages the evaluation-generation gap via reinforcement learning strategies, facilitating self-improvement. These methods enhance LLM adaptability by integrating self-rewarding strategies and reinforcement learning paradigms.

\subsubsection{Multi-Agent Co-Evolution}
\label{sec:multi-agent-co-evol}

Multi-agent co-evolution enables LLMs to improve through interactions with other agents. This involves cooperative learning, where agents share information and coordinate actions, as well as competitive co-evolution, where agents engage in adversarial interactions to refine strategies and enhance performance.

\paratitle{Cooperative and Collaborative Learning.}  
Multi-agent collaboration enhances LLMs by enabling knowledge sharing, joint decision-making, and coordinated problem-solving. 
For instance, ProAgent~\cite{zhang2024proagent} enables LLM-based agents to adapt dynamically in cooperative tasks by inferring teammates' intentions and updating beliefs, enhancing zero-shot coordination. CORY~\cite{ma2024coevolving} extends RL fine-tuning into a cooperative multi-agent framework, where LLMs iteratively improve through role-exchange mechanisms, enhancing policy optimality and stability. 
CAMEL~\cite{li2023camel} develops a role-playing framework where communicative agents collaborate autonomously using inception prompting, improving coordination and task-solving efficiency in multi-agent settings.
These approaches contribute to more efficient, adaptable, and intelligent multi-agent LLM systems.

\paratitle{Competitive and Adversarial Co-Evolution.}  
Competitive co-evolution strengthens LLMs through adversarial interactions, debate, and strategic competition. For example, Red-team LLMs~\cite{ma2023evolving} dynamically evolve in adversarial interactions, continuously challenging LLMs to uncover vulnerabilities and mitigate mode collapse, leading to more robust safety alignment.
Du et al. propose a multi-agent debate framework~\cite{du2023improving} to enhance reasoning by having multiple LLMs critique and refine each other's arguments over multiple rounds, improving factuality and reducing hallucinations. Furthermore, the MAD framework~\cite{liang2024encouraging} structures debates among agents in a tit-for-tat manner, encouraging divergent thinking and refining logical reasoning in complex tasks. These competitive co-evolution strategies drive LLMs to develop stronger reasoning, resilience, and strategic adaptability in a multi-agent adversarial manner.

\subsubsection{Evolution via External Resources}
\label{sec:evol-external-resource}

External resources enhance the evolution of agents by providing structured information and feedback. Knowledge-enhanced evolution integrates structured knowledge to improve reasoning and decision-making, while external feedback-driven evolution leverages real-time feedback from tools and environments to refine model performance.

\paratitle{Knowledge-Enhanced Evolution.}  
LLMs can evolve by integrating structured external knowledge, improving reasoning, decision-making, and task execution. For example, KnowAgent~\cite{zhu2024knowagent} improves LLM-based planning by integrating action knowledge, constraining decision paths, and mitigating hallucinations, leading to more reliable task execution. The world knowledge model (WKM)~\cite{qiao2024agent} enhances agent planning by synthesizing expert and empirical knowledge, providing global priors and dynamic local knowledge to guide decision-making. 
These approaches collectively improve the evolution of LLM by incorporating diverse and structured external information.
 
\paratitle{External Feedback-Driven Evolution.}  
LLMs can refine their behavior by leveraging external feedback from tools, evaluators, and humans to improve performance iteratively. 
For example, CRITIC~\cite{goucritic} allows LLMs to validate and revise their outputs through tool-based feedback, improving accuracy and reducing inconsistencies.  
STE~\cite{song2024trial} enhances tool learning by simulating trial-and-error, imagination, and memory, enabling more effective tool use and long-term adaptation.  
SelfEvolve~\cite{jiang2023selfevolve} adopts a two-step framework where LLMs generate and debug code using feedback from execution results, enhancing performance without human intervention. 
These approaches enable LLMs to evolve iteratively by integrating structured feedback, improving adaptability and robustness.

\section{Evaluation and Tools}\label{sec:dataset}

\begin{figure}[t]
\centering
\includegraphics[width=0.5\textwidth]{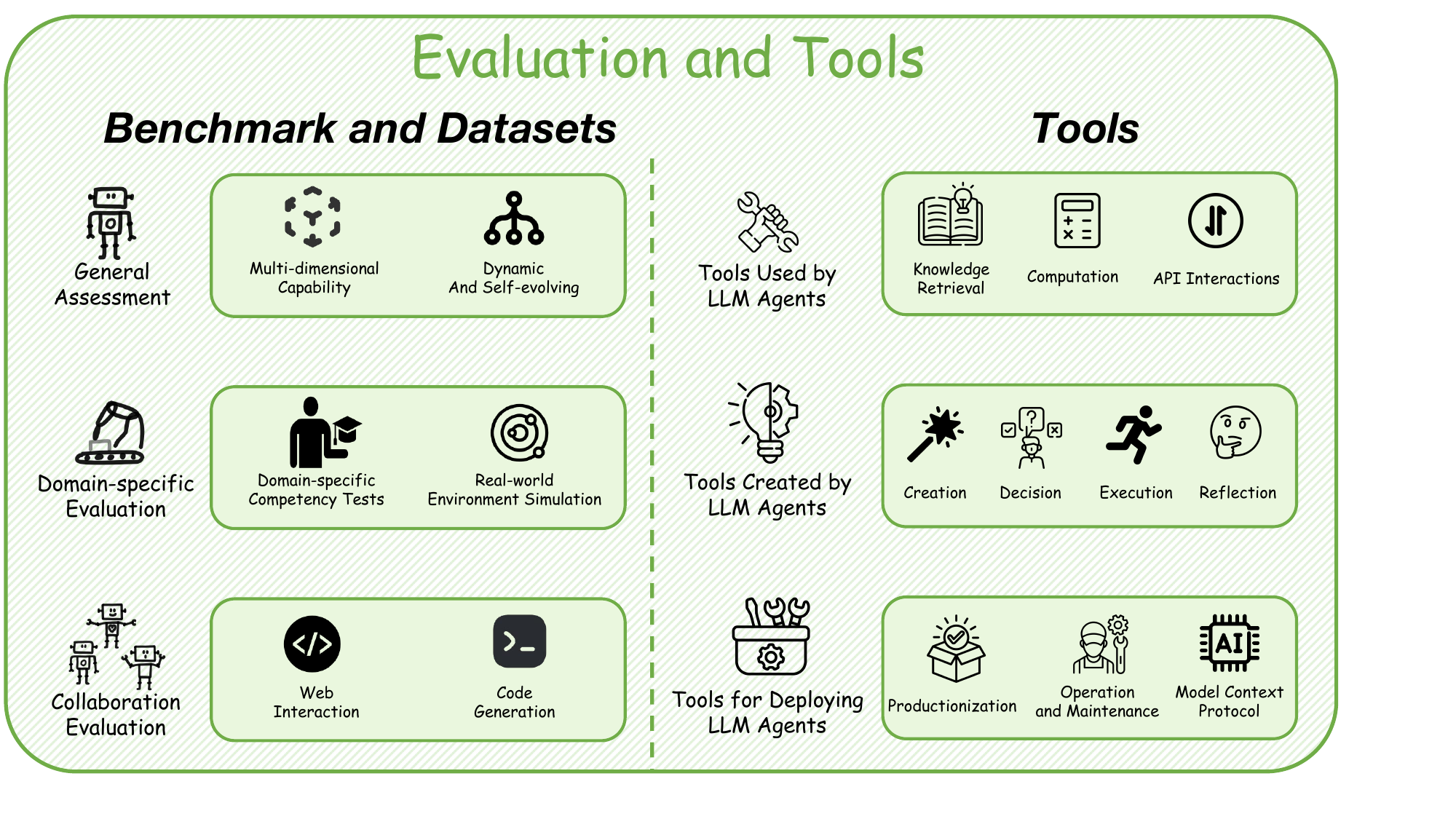}
\caption{An overview of evaluation benchmarks and tools for LLM agents. The left side shows various evaluation frameworks categorized by general assessment, domain-specific evaluation, and collaboration evaluation. The right side illustrates tools used by LLM agents, tools created by agents, and tools for deploying agents.}
\vspace{-3mm}
\label{fig:eval-tools}
\end{figure}

As LLM agents continue to evolve in complexity and capability, robust evaluation frameworks and specialized tools have become essential components of the agent ecosystem. This section explores the comprehensive landscape of benchmarks, datasets, and tools that enable the development, assessment, and deployment of LLM agents. We first examine evaluation methodologies in Section~\ref{sec:benchmark_datasets}, covering general assessment frameworks, domain-specific evaluation systems, and collaborative evaluation approaches. We then discuss the tools ecosystem in Section~\ref{sec:tools}, including tools used by LLM agents, tools created by agents themselves, and infrastructure for deploying agent systems.

\subsection{Evaluation Benchmarks and Datasets}
\label{sec:benchmark_datasets}

The evolution of LLM agents has driven the creation of specialized benchmarks that systematically evaluate agent capabilities across technical dimensions and application domains. These frameworks address three key requirements: general assessment frameworks, domain-specific scenario simulation, and collaborative evaluation of complex systems.

\subsubsection{General Assessment Frameworks}
The evolution of intelligent agents requires evaluation frameworks to move beyond simple success-rate metrics to comprehensive cognitive analysis. Recent advances focus on building adaptive and interpretable assessment systems capable of capturing the subtle interplay between reasoning depth, environmental adaptability, and task complexity. 

\paratitle{Multi-Dimensional Capability Assessment.}
Modern benchmarks are increasingly adopting a hierarchical paradigm that dissects agent intelligence across various dimensions of reasoning, planning, and problem solving. AgentBench \cite{liu2023agentbench} builds a unified test field across eight interactive environments, revealing the advantages of a commercial LLM in complex reasoning. Mind2Web \cite{deng2023mind2web} extends this paradigm to web interaction scenarios, proposing the first generalist agent for evaluating 137 real-world websites with different tasks spanning 31 domains. This open environment benchmark enables multi-dimensional capability assessment through real web-based challenges. This is in line with MMAU \cite{yin2024mmau}, which enhances explainability through granular capability mapping and breaks down agent intelligence into five core competencies by more than 3,000 cross-domain tasks. BLADE~\cite{gu2024blade} extends evaluation to scientific discovery by tracking the analytical decision patterns of expert validation workflows. VisualAgentBench \cite{liu2024visualagentbench} further extends this approach to multimodal foundation agents, establishing a unified benchmark across materialized interactions, GUI operations, and visual design tasks, and rigorously testing the LLM's ability to handle the dynamics of the complex visual world. Embodied Agent Interface \cite{li2025embodied} introduces modular inference components (object interpretation, subobject decomposition, etc.) to provide fine-grained error classification for embedded systems. CRAB \cite{xu2024crab} offers cross-platform testing with graphics-based assessment and a unified Python interface. These frameworks emphasize the shift from a single measure of success to multifaceted cognitive analysis.

\paratitle{Dynamic and Self-Evolving Evaluation Paradigms.}
Next-generation framework addresses baseline obsolescence through adaptive generation and human-AI collaboration. BENCHAGENTS \cite{butt2024benchagents} automatically creates benchmarks through LLM agents for planning, validating, and measuring designs, enabling rapid capacity expansion. Benchmark self-evolving \cite{wang2024benchmark} introduces six refactoring operations to dynamically generate test instances for short-cut biases. Revisiting Benchmark \cite{wang2024revisiting} proposed TestAgent with reinforcement learning for domain adaptive assessment. Other methods such as Seal-Tools \cite{wu2024seal} (1,024 nested instances of tool calls) and CToolEval \cite{guo2024ctooleval} (398 Chinese APIs across 14 domains), complement static datasets and standardize tool usage evaluation.

\subsubsection{Domain-Specific Evaluation System}
The increasing specialization of agent applications requires evaluation systems tailored to domain-specific knowledge and environmental constraints. Researchers are developing dual-axis frameworks that combine vertical competency testing for professional scenarios with horizontal validation in real-world simulated environments.

\paratitle{Domain-Specific Competency Tests.}
Several key application areas are specifically benchmarked with scenario-driven assessments. For example, healthcare applications are rigorously tested by MedAgentBench \cite{jiang2025medagentbench} and AI Hospital \cite{fan2024ai}. Specifically, MedAgentBench contains tasks designed by 300 clinicians in an FHIR-compliant environment, while the AI hospital simulates clinical workflows through multi-agent collaboration. The autonomous driving system benefits from LaMPilot \cite{ma2024lampilot}, which connects the LLM to the autonomous driving architecture through code generation benchmarks. Data science capabilities are evaluated by DSEval \cite {zhang2024benchmarking} and DA-Code \cite {huang2024code}, covering lifecycle management from data debate to model deployment, while DCA-Bench~\cite{huang2024dca} evaluates dataset curation agents based on real-world quality issues. TravelPlanner \cite{xie2024travelplanner} provides a sandbox environment for travel planning scenarios. It contains 1225 planning tasks that require multi-step reasoning, tool integration, and constraint balancing under realistic conditions (e.g., budget and time). Machine learning engineering capabilities, measured by MLAgant-Bench \cite{huang2023benchmarking} and MLE-Bench \cite{chan2024mle}, simulate kaggle-like challenges that require optimization of an end-to-end pipeline. Security-focused AgentHarm \cite{andriushchenkoagentharm} curated 440 malicious agent tasks in 11 hazard categories, and systematically assessed LLM abuse risk for the first time in a multi-step tool usage scenario. These domain-specific benchmarks reveal significant performance gaps compared to general testing in practical applications.

\paratitle{Real-World Environment Simulation.}
Several benchmarks bridge the simulation to reality gap with real interactive environments. OSWorld \cite{xie2025osworld} builds the first scalable real-computer ecosystem that supports 369 multi-application tasks across Ubuntu/Windows/macOS. TurkingBench \cite{xu2024tur} evaluates 158 micro-tasks using a crowdsourcing-derived HTML interface, and LaMPilot \cite{ma2024lampilot} introduces an executable code generation benchmark for autonomous driving scenarios. OmniACT \cite{kapoor2024omniact} provides 32K web/desktop automation instances with basic requirements for visualization. EgoLife \cite{yang2025egolife} advances real-world simulation through a 300-hour multimodal egocentric dataset capturing daily human activities (e.g., shopping, cooking, socializing), paired with EgoLifeQA tasks that test agents’ long-term memory retrieval, health habit monitoring, and personalized recommendation capabilities in dynamic environments. GTA \cite{wang2024gta} further integrates real-world deployed tools and multi-modal inputs (images, web pages) to evaluate real-world problem-solving capabilities. 

\subsubsection{Collaborative Evaluation of Complex Systems}
As agency systems evolve toward organizational complexity, evaluation frameworks must quantify emergent coordination patterns and collective intelligence. Recent approaches shift evaluation from isolated agent proficiency to system-level cognitive collaboration, revealing scalability challenges in multi-agent workflows.

\paratitle{Multi-Agent System Benchmarking.}
TheAgentCompany \cite{xu2024theagentcompany} pioneered enterprise-level assessments using simulated software company environments to test web interaction and code collaboration capabilities. Comparative analysis like AutoGen and CrewAI \cite{barbarroxa2024benchmarking} establishes methodological standards through ML code generation challenges. Large Visual Language Model Survey \cite{li2025benchmark} systematizes over 200 multi-modal benchmarks. For multi-agent collaboration, MLRB \cite{kenney2024ml} designs 7 competition-level ML research tasks, and MLE-Bench \cite{chan2024mle} evaluates Kaggle-style model engineering through 71 real-world competitions. These efforts collectively establish rigorous evaluation protocols for emergent agent coordination capabilities.

\subsection{Tools}\label{sec:tools}

Tools are an important part of LLM agents. When dealing with complex tasks, LLM agents can call on external tools to generate more precise answers. Depending on their creativity, they can also create tools to solve tasks. In addition, LLM agents need corresponding tools for deployment, maintenance, and data acquisition.

\subsubsection{Tools used by LLM agents}
Since LLM agents do not perform well in handling some specific tasks, such aas those requiring real-time information and accurate calculations, external tools are introduced to help the LLM agents perform these tasks more effectively. These external tools can be categorized into three main groups.

\paratitle{Knowledge Retrieval.} For those real-time information that LLM agents are not aware of, knowledge retrieval tools, such as search engines, can help LLM agents to quickly access up-to-date knowledge so that they are no longer limited to the knowledge base they had during training. WebGPT~\cite{nakano2022webgpt} successfully combines online search engines and LLMs with the incorporation of the commercial API\footnote{\hyperlink{https://www.microsoft.com/en-us/bing/apis/bing-web-search-api}{https://www.microsoft.com/en-us/bing/apis/bing-web-search-api}}. WebCPM~\cite{qin2023webcpm}, inspired by WebGPT, develops a web search interface and uses it to construct the first Chinese long-form question answer (LFQA) dataset. ToolCoder~\cite{zhang2023toolcode} uses DuckDuckgo\footnote{\hyperlink{https://duckduckgo.com}{https://duckduckgo.com}} as the search engine for those frequently used public libraries and employs the BM25~\cite{robertson2009probabilistic} score for those less-known or private libraries.

\paratitle{Computation.} LLM agents may suffer hallucinations when dealing with tasks requiring precise computation. Computational tools like Python interpreters and math calculators can help LLM agents with complex code execution or computational tasks. AutoCoder~\cite{lei2024autocoder} designs a dataset with the interaction with coding execution results to facilitate LLM-based code generation. RLEF~\cite{gehring2025rlef} improves code generation performance through an end-to-end reinforcement learning framework that enables LLMs to learn feedback from code executors. CodeActAgent~\cite{Wang2024ExecutableCA} is an automatic agentic system which can update the actions based on the interaction with the code interpreter.
Toolformer~\cite{schick2023toolformer} integrates a range of tools, including calculators, to significantly improve the performance of models in tasks such as mathematical calculations without compromising the model's generality. ART~\cite{paranjape2023art} enables LLM to invoke external tools, such as calculators, when solving complex tasks and excels in mathematical reasoning and complex computational tasks.

\paratitle{API Interactions.} Building on external APIs, such as REST APT, can enable LLM agents to call external services and extend their functionality, such as manipulating databases and implementing end-to-end automated processes. RestGPT~\cite{song2023restgpt} explores more realistic scenarios by combining LLM with RESTful APIs and presents RestBench to evaluate the performance of RestGPT. GraphQLRestBench~\cite{saha2024sequential} builds a dataset consisting of sequences of natural language statements, and function calls to review existing open-source LLMs, exploring the capabilities of LLMs for API calls.

\subsubsection{Tools created by LLM agents}
Since the users of traditional tools tend to be humans, LLM agents often have limitations when making calls. In addition, the limitations of existing tools make it difficult to effectively handle new problems. In recent years, many studies have explored how LLM agents can create their tools. CRAFRT~\cite{yuan2023craft} provides a flexible framework for tool creation and retrieval by collecting GPT-4 code solutions for specific tasks and abstracting them into code snippets to create specialized tool sets for the tasks. Toolink~\cite{qian2024toolink} performs task resolution by creating a toolset and then integrating the planning and invocation of tools through a Chain of Solutions (CoS) approach. CREATOR~\cite{qian2023creator} proposes a four-phase framework--Creation, Decision, Execution, and Reflection--to enable LLM agents to create tools and improve the robustness of the output. LATM~\cite{cai2024largelanguagemodelstool} proposes a two-stage framework that allows LLMs to act as tool makers and tool users, respectively and proposes a tool caching mechanism that improves the efficiency of task solving and reduces the cost while maintaining performance by assigning different models to different tasks with different levels of difficulty.

\subsubsection{Tools for deploying LLM agents}

LLM tools are essential for the deployment, development, operation, and maintenance of LLM agents and for the secure transmission of data. According to their role, these tools can be categorized into three types.

\paratitle{Productionization.}
The main purpose of the productionization tools is to make it easy for users to deploy LLM agents in production environments. AutoGen~\cite{wu2023autogen} is an open-source framework that enables developers to build LLM applications with customizable, conversational multiple agents. LangChain~\cite{LangChian_2023} is an open-source framework for building LLM applications that is highly extensible and allows users to create custom modules and workflows to meet their specific needs. LlamaIndex~\cite{Liu_LlamaIndex_2022} is a data framework serving large model applications, allowing users to build LLM applications based on local data. It also provides a rich toolbox for accessing and indexing data, retrieving and reordering, and building custom query engines. Dify~\cite{Dify_2023} is an open-source LLM application development platform that differs from other platforms in that it allows users to build and test powerful AI workflows on canvas.

\paratitle{Operation and Maintenance.}
After deploying LLM agents, the O\&M tool ensures that the model performs well during training and remains reliable during production. Ollama~\cite{Ollama_2023} is a platform for building LLM agents that also offers observability and monitoring support, allowing teams to track their models' performance in real-time. Dify~\cite{Dify_2023} enables users to monitor and analyze application logs and performance over time, allowing for continuous improvements in prompts, datasets, and models based on production data and annotations.

\paratitle{Model Context Protocol.} MCP\footnote{\hyperlink{https://modelcontextprotocol.io/introduction}{https://modelcontextprotocol.io/introduction}} is an open protocol that standardizes how applications provide context to LLMs. It is used to create secure links between LLMs and data sources as well as to build LLM agents and workflows. MCP-Agent~\cite{MCP_Agent_2025} is a simple framework to build agents using MCP. As more services become MCP-aware, users will be able to take full advantage of them.

\section{Real-World Issues}\label{sec:realworld}

\begin{figure}[t]
\centering
\includegraphics[width=0.5\textwidth]{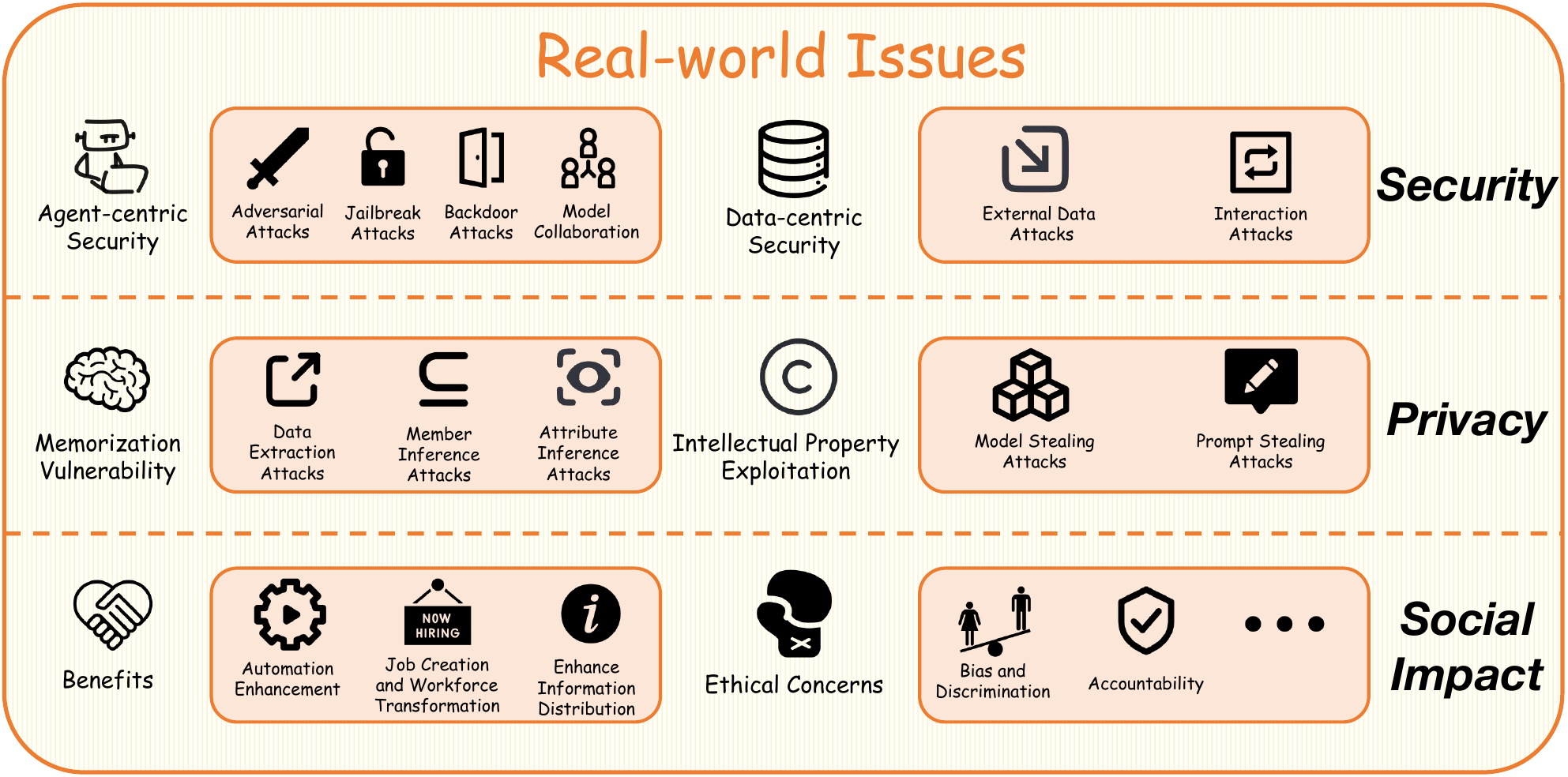}
\caption{An overview of real-world issues in LLM agent systems, organized into three domains: security challenges (including agent-centric and data-centric threats), privacy concerns (covering memorization vulnerabilities and intellectual property exploitation), and social impact considerations (highlighting both benefits and ethical challenges).}
\vspace{-5mm}
\label{fig:real-world-issuse}
\end{figure}

As LLM agents become increasingly integrated into various aspects of society, they bring forth significant real-world challenges that must be addressed for responsible deployment. Figure~\ref{fig:real-world-issuse} provides an overview of these challenges, categorized into three primary domains: security, privacy, and social impact. Security concerns encompass both agent-centric threats~(Section~\ref{subsec:security_1}) that target model components and data-centric threats~(Section~\ref{subsec:security_2}) that contaminate input data. Privacy issues~(Section~\ref{subsec:privacy}) include memorization vulnerabilities and intellectual property exploitation. Beyond technical concerns, LLM agents raise important ethical considerations and have broad societal implications~(Section~\ref{subsec:ethics}), including both potential benefits and risks to society. Understanding these challenges is crucial for developing robust, trustworthy agent systems.

\subsection{Agent-centric Security}\label{subsec:security_1}

Agent-centric security targets defending different types of attacks on the agent models, where attacks aim to manipulate, tamper, and steal critical components of the weights, architecture, and inference process of the agent models. These agent-centric attacks may lead to performance degradation, maliciously manipulated outputs, and privacy leaks within agent systems. 
Li et al.~\cite{li2025commercial} analyze the security vulnerabilities of LLM agents under attacks categorized by threat actors, objectives, entry points, and so on. They also conduct experiments on certain popular agents to demonstrate their security vulnerabilities.
Agent security bench~\cite{zhang2024agent} introduces a comprehensive framework to evaluate attacks and defenses for LLM-based agents across 10 scenarios, 10 agents, 400+ tools, 23 attack/defense methods, and 8 metrics, revealing significant vulnerabilities and limited defense effectiveness of current LLM agents.
We summarize the agent-centric security issues in the blow categories.

\subsubsection{Adversarial Attacks and Defense}
Adversarial attacks aim to compromise the reliability of the agents, rendering them ineffective in specific tasks.
Mo et al.~\cite{mo2024trembling} categorize adversarial attacks into three components, i.e., \textit{Perception, Brain, and Action}. 
AgentDojo~\cite{NEURIPS2024_97091a51} provides an evaluation framework designed to measure the adversarial robustness of AI agents by testing them on 97 realistic tasks and 629 security test cases.
ARE~\cite{wu2024adversarial} evaluates multimodal agent robustness under adversarial attacks.
For adversarial attack methods,
CheatAgent~\cite{ning2024cheatagent} uses an LLM-based agent to attack black-box LLM-empowered recommender systems by identifying optimal insertion positions, generating adversarial perturbations, and refining attacks through iterative prompt tuning and feedback.
GIGA~\cite{yuinfecting} introduces generalizable infectious gradient attacks to propagate adversarial inputs across multi-agent, multi-round LLM-powered systems by finding self-propagating inputs that generalize well across contexts.
For adversarial attacks defense methods,
LLAMOS~\cite{lin2024large} introduces a defense technique for adversarial attacks by purifying adversarial inputs using agent instruction and defense guidance before they are input into the LLM.
Chern et al.~\cite{chern2024combating} introduce a multi-agent debate method to reduce the susceptibility of agents to adversarial attacks.

\subsubsection{Jailbreaking Attacks and Defense}
Jailbreaking attacks attempt to break through the protection of the model and obtain unauthorized functionality or information.
For jailbreaking attack methods,
RLTA~\cite{wang2024reinforcement} uses reinforcement learning to automatically generate attacks that produce malicious prompts, triggering LLM agents' jailbreaking to produce specific output. These can be adapted to both white box and black box scenarios.
Atlas~\cite{dong2024jailbreaking} jailbreaks text-to-image models with safety filters using a mutation agent and a selection agent, enhanced by in-context learning and chain-of-thought techniques.
RLbreaker~\cite{chenllm} is a black-box jailbreaking attack using deep reinforcement learning to model jailbreaking as a search problem, featuring a customized reward function and PPO algorithm.
PathSeeker~\cite{lin2024pathseeker} also uses multi-agent reinforcement learning to guide smaller models in modifying inputs based on the target LLM's feedback, with a reward mechanism leveraging vocabulary richness to weaken security constraints.
For jailbreaking defense methods,
AutoDefense~\cite{zeng2024autodefense} proposes a multi-agent defense framework that uses LLM agents with specialized roles to collaboratively filter harmful responses, effectively resisting jailbreak attacks.
Guardians~\cite{barua2025guardians} uses three examination methods—reverse Turing Tests, multi-agent simulations, and tool-mediated adversarial scenarios—to detect rogue agents and counter jailbreaking attacks.
ShieldLearner~\cite{ni2025shieldlearner} proposes a novel defense paradigm for jailbreak attacks by autonomously learning attack patterns and synthesizing defense heuristics through trial and error.

\subsubsection{Backdoor Attacks and Defense}
Backdoor attacks implant specific triggers to cause the model to produce preset errors when encountering these triggers while performing normally under normal inputs.
For backdoor attack methods,
DemonAgent~\cite{zhu2025demonagent} proposes a dynamically encrypted muti-backdoor implantation attack method by using dynamic encryption to map and decompose backdoors into multiple fragments to avoid safety audits.
Yang et al.~\cite{yang2025watch} investigate and implement diverse forms of backdoor attacks on LLM-based agents, demonstrating their vulnerability through experiments on tasks like web shopping and tool utilization.
BadAgent~\cite{wang2024badagent} attacks LLM-based intelligent agents to trigger harmful operations through specific inputs or environment cues as backdoors.
BadJudge~\cite{tong2025badjudge} introduces a backdoor threat specific to the LLM-as-a-judge agent system, where adversaries manipulate evaluator models to inflate scores for malicious candidates, demonstrating significant score inflation across various data access levels.
DarkMind~\cite{guo2025darkmind} is a latent backdoor attack that exploits the reasoning processes of customized LLM agents by covertly altering outcomes during the reasoning chain without requiring trigger injection in user inputs.



\subsubsection{Model Collaboration Attacks and Defense}
Model collaboration attack is an emerging type of attack that mainly targets scenarios where multiple models work together. In this type of attack, attackers manipulate the interaction or collaboration mechanisms between multiple models to disrupt the overall functionality of the system.
For model collaboration attack methods,
CORBA~\cite{zhou2025corba} introduces a novel yet simple attack method for the LLM multi-agent system. It exploits contagion and recursion, which are hard to mitigate via alignment, disrupting agent interactions. 
AiTM~\cite{he2025red} introduces an attack method to the LLM multi-agent system by intercepting and manipulating inter-agent messages using an adversarial agent with a reflection mechanism.
For the defense methods,
Netsafe~\cite{yu2024netsafe} identifies critical safety phenomena and topological properties that influence the safety of multi-agent networks against adversarial attacks.
G-Safeguard~\cite{wang2025g} is also based on topology guidance and leverages graph neural networks to detect anomalies in the LLM multi-agent system. 
Trustagent~\cite{hua2024trustagent} aims to enhance the planning safety of LLM agentic framework in three different planning stages. 
PsySafe~\cite{zhang2024psysafe} is grounded in agent psychology to identify, evaluate, and mitigate safety risks in multi-agent systems by analyzing dark personality traits, assessing psychological and behavioral safety, and devising risk mitigation strategies.

\begin{table}[t]
\caption{Summary of agent-centric attacks and defense in LLM agents.}
\label{tab:security_1_literature}
\centering
\resizebox{0.5\textwidth}{!}
{
\begin{tabular}{ll}
\toprule
\textbf{Reference} & \textbf{Description} \\
\midrule
\multicolumn{2}{c}{\textbf{Adversarial Attacks and Defense}} \\
\midrule
Mo et al.~\cite{mo2024trembling}& \textbf{Attack:} Adversarial attack benchmark \\
AgentDojo~\cite{NEURIPS2024_97091a51} & \textbf{Attack:}  Adversarial attack framework\\
ARE~\cite{wu2024adversarial} & \textbf{Attack:} Adversarial attack evaluation for multimodal agents\\
GIGA~\cite{yuinfecting} & \textbf{Attack:} Generalizable infectious gradient attacks \\
CheatAgent~\cite{ning2024cheatagent} & \textbf{Attack:}  Adversarial attack agent for recommender systems\\
LLAMOS~\cite{lin2024large}  & \textbf{Defense:} Purifying adversarial attack input\\
Chern et al.~\cite{chern2024combating}  & \textbf{Defense:} Defense via multi-agent debate \\
\midrule
\multicolumn{2}{c}{\textbf{Jailbreaking Attacks and Defense}} \\
\midrule
RLTA~\cite{wang2024reinforcement} & \textbf{Attack:} Produce jailbreaking prompts via reinforcement learning  \\
Atlas~\cite{dong2024jailbreaking}  & \textbf{Attack:} Jailbreaks text-to-image models with safety filters \\ 
RLbreaker~\cite{chenllm}  & \textbf{Attack:} Model jailbreaking as a search problem \\
PathSeeker~\cite{lin2024pathseeker} & \textbf{Attack:} Use multi-agent reinforcement learning to jailbreak \\
AutoDefense~\cite{zeng2024autodefense}  & \textbf{Defense:} Multi-agent defense to filter harmful responses\\
Guardians~\cite{barua2025guardians}  & \textbf{Defense:} Detect rogue agents to counter jailbreaking attacks.\\
ShieldLearner~\cite{ni2025shieldlearner}  & \textbf{Defense:} Learn attack jailbreaking patterns.\\
\midrule
\multicolumn{2}{c}{\textbf{Backdoor Attacks and Defense}} \\
\midrule
DemonAgent~\cite{zhu2025demonagent} & \textbf{Attack:} Encrypted muti-backdoor implantation attack \\
Yang et al.~\cite{yang2025watch}  & \textbf{Attack:} Backdoor attacks evaluations on LLM-based agents\\
BadAgent~\cite{wang2024badagent}  & \textbf{Attack:} Inputs or environment cues as backdoors\\
BadJudge~\cite{tong2025badjudge}  & \textbf{Attack:} Backdoor to the LLM-as-a-judge agent system\\
DarkMind~\cite{guo2025darkmind}  & \textbf{Attack:} latent backdoor attack to customized LLM agents\\
\midrule
\multicolumn{2}{c}{\textbf{Agent Collaboration Attacks and Defense}} \\
\midrule
CORBA~\cite{zhou2025corba} & \textbf{Attack:} Multi-agent attack via multi-agent \\
AiTM~\cite{he2025red} & \textbf{Attack:} Intercepte and manipulate inter-agent messages\\
Netsafe~\cite{yu2024netsafe}  & \textbf{Defense:} Identify critical safety phenomena in multi-agent networks\\
G-Safeguard~\cite{wang2025g} & \textbf{Defense:} leverages graph neural networks to detect anomalies\\
Trustagent~\cite{hua2024trustagent} & \textbf{Defense:}  Agent constitution in task planning.\\
PsySafe~\cite{zhang2024psysafe} & \textbf{Defense:} Mitigate safety risks via agent psychology \\
\bottomrule
\end{tabular}
}
\end{table}

\subsection{Data-centric Security}\label{subsec:security_2}
The goal of data-centric attacks is to contaminate the input data of LLM agents, ultimately leading to unreasonable tool calling, aggressive outputs and resource depletion, etc~\cite{deng2024ai}. In data-centric attacks, any components in LLM agent systems or default parameters are not allowed to be modified. Based on the data type, we categorize attacks into external data attacks and execution data attacks. Corresponding defense strategies are summarized to counter these agent attacks.

\subsubsection{External Data Attack and Defense}

\paratitle{User Input Falsifying.} Modifying the user input is the most straightforward and widely used data-centric attacks. These injections~\cite{zhang2024agent} can lead to uncontrolled and dangerous outputs. Though it is simple, it always achieves the highest Attack Success Rate (ASR)~\cite{debenedetti2025agentdojo,zhang2024agent}. Li et al.~\cite{li2024targeting} propose malicious prefix prompts, such as ``ignore the document".  
InjectAgent~\cite{zhan2024injecagent} and Agentdojo~\cite{debenedetti2025agentdojo}  are two prompt injection benchmarks, which test the single and multi-turn attacks in LLM agents. As the widespread effect of injections on user inputs increases, various defense models have been  designed. Mantis~\cite{pasquini2024hacking} defenses through hacking back to attackers' own systems. \cite{abdelnabi2025firewalls} offers a defense module called the Input Firewall, which extracts key points from users' natural language and converts them into a structured JSON format. RTBAS~\cite{zhong2025rtbas} and TaskShield~\cite{jia2024task} check the every step of information flow and agent process, including function calls and tool execution, to make sure the execution aligns with the original instructions and intentions. In the ASB~\cite{zhang2024agent} benchmark, a sandwich defend strategy adds additional guarding instructions to help LLM agents ignore malicious injections.

\paratitle{Dark Psychological Guidance.} Attackers can carry out dark psychological guidance in the prompts, \textit{e.g.,} use ``cheating" instead of ``care",  ``betrayal" instead of ``fairness",  ``subversion" instead of ``authority". Then LLM agents are guided to be aggressive and antisocial, which may cause serious social impacts. \cite{tian2023evil} proposes the ``Evil Geniuses" to generate prompts to put agents into specific role-playing states. Its prompts are optimized through the red-blue exercises. \cite{zhang2024psysafe} injects the dark psychological traits into the user inputs. To defense dark psychological injections, doctor and police agents~\cite{zhang2024psysafe} are incorporated into the agents systems. The doctor agents conduct the psychological assessment, while the police agents supervise the safety of agent systems. They work together to guard the healthy psychology at any time.

\paratitle{External Source Poisoning.} Many attackers pay their attention to the RAG-based LLM agents, as they have been proven to be more reliable than general memory-based LLM agents~\cite{wang2024biorag}. The attackers inject poisoning samples into the knowledge databases~\cite{li2025commercial,gan2024navigating}. Based on this, the Indirect Prompt Injection (IPI) attack embeds malicious instructions into other external knowledge sources~\cite{xiang2024clas}, such as the websites, support literature, emails, online BBS, which can manipulate agents and cause them to deviate from the original intentions. WIPI~\cite{wu2024wipi} controls the agents through a public web page to indirectly poison instructions. \cite{nakash2024breaking} describes a Foot-in-the-Door (FITD) attack, which begins with inconspicuous, unrelated requests and gradually incorporates harmless ones. This approach increases the likelihood of the agent executing subsequent actions, leading to resource consumption that could have been avoided.
AgentPoison~\cite{chen2025agentpoison} is a typical red teaming work, which achieves a high success rate in knowledge-intensive QA agent. \cite{chern2024combating} employs a multi-agent debate for defense, where each agent acts as a domain expert to verify the facticity of external knowledge.

\subsubsection{Interaction Attack and Defense}

\paratitle{Interaction between user and agent interface.}
Some LLM agents store the private user-agent interactions in users' computer memory to enhance dialogue performance. During these interactions, LLM agents are usually black-box to attackers.
\cite{wang2025unveiling} is a private memory extraction attack that aggregates multiple levels of knowledge from the stored memory. \cite{EmbraceTheRed} presents an attack that occurs at the interface between users and LLM agents, where it solicits information from users.

\paratitle{Interaction among LLM agents.}
In multi-agent LLM systems, the interactions among agents are frequent and essential~\cite{xu2024survey}. Attackers poison a single agent, which then infects other agents~\cite{li2024personal}. This recursive attack can ultimately deplete the computational resources. AgentSmith~\cite{gu2024agent} concludes that the infectious spread occurs exponentially fast.
The Contagious Recursive Blocking Attack (CORBA)~\cite{zhou2025corba} is designed to disrupt the communications among agents, allowing the infection to propagate across the entire communication network. 
\cite{he2025red} incorporates a reflection mechanism to finish disruptions based on the semantic understanding of communications. \cite{lee2024prompt} injects malicious instructions into one agent, enabling them to self-replicate across the agent network, resembling the spread of a computer virus. Additionally, \cite{lee2024prompt}  develops a tagging strategy to control the infection spread. To defend against Byzantine attacks during the agent interactions, BlockAgents~\cite{chen2024blockagents} introduces a consensus mechanism based on blockchain and proof-of-thought (PoT) techniques. The agent that contributes the most to the planning process is granted the accounting rights.

\paratitle{Interaction between agents and tools.} 
To call appropriate tools, the agents first make a plan, and then finish the action. The interaction between agents and tools is vulnerable. Some attackers maliciously modify planning thoughts, and thus alter the agent actions. The agent may call unconvincing or harmful tools to complete the task, and further cause unexpected consequences. AgentHarm~\cite{andriushchenko2024agentharm} adds harmful distractions during multi-step execution tasks.  InjectAgent~\cite{zhan2024injecagent} conducts attacks during the agent planning process. The multi-layer agent firewall~\cite{abdelnabi2025firewalls} incorporates a self-correction mechanism, known as the trajectory firewall layer, to correct the deviated trajectory of agents. This firewall layer verifies the generated responses to ensure compliance with security rules. 

\begin{table}[t]
\caption{Summary of data-centric attack and defense in LLM agents.}
\label{tab:security_2_literature}
\centering
\resizebox{0.5\textwidth}{!}
{
\begin{tabular}{ll}
\toprule
\textbf{Reference} & \textbf{Description} \\
\midrule
\multicolumn{2}{c}{\textbf{External Data Attacks and Security}} \\
\midrule
Li et al.~\cite{li2024targeting} & \textbf{Attack:} Malicious prefix injection\\
Psysafe~\cite{zhang2024psysafe} & \textbf{Attack:} A dark psychological injection benchmark  \\
Tian et al.~\cite{tian2023evil} & \textbf{Attack:} Guide agents into specific role-playing states  \\
InjectAgent~\cite{zhan2024injecagent}  & \textbf{Attack:} A prompting injection benchmark\\
Agentdojo~\cite{debenedetti2025agentdojo}  & \textbf{Attack:} A user  injection benchmark \\
AgentPoison~\cite{chen2025agentpoison} & \textbf{Attack:} Poisoning samples in knowledge databases\\
Nakash et al.\cite{nakash2024breaking} & \textbf{Attack:} Indirect prompt injection through FITD attack \\
WIPI~\cite{wu2024wipi} & \textbf{Attack:} control agents through a public web page\\
ASB~\cite{zhang2024agent}  & \textbf{Attack:} A multi-type attack benchmark \\
AgentHarm~\cite{andriushchenko2024agentharm} & \textbf{Attack:} A multi-type attack benchmark \\
Mantis~\cite{pasquini2024hacking}  & \textbf{Defense:} Hacking back to attackers \\
Chern et al.\cite{chern2024combating} & \textbf{Defense:} Employ multi-agent debate to verify external knowledge\\
RTBAS~\cite{zhong2025rtbas} & \textbf{Defense:} Check every step of agent information flow \\
TaskShield~\cite{jia2024task} & \textbf{Defense:} Check every step of agent process \\
Zhang et al.~\cite{zhang2024psysafe} & \textbf{Defense:} Doctor and police agents guard the healthy psychology
\\
\midrule
\multicolumn{2}{c}{\textbf{Interaction Attacks and Security}} \\
\midrule
Wang et al.~\cite{wang2025unveiling} & \textbf{Attack:}  Private memory extraction attack \\
CORBA~\cite{zhou2025corba} & \textbf{Attack:} Disrupt the communications among agents\\
AgentSmith~\cite{gu2024agent} & \textbf{Attack:} Poison one agent to infectious other agents\\
Lee et al.~\cite{lee2024prompt} & \textbf{Attack:} Conduct injections to self-replicate among agents\\
He et al.~\cite{he2025red}  & \textbf{Attack:} Inject semantic disruptions to agent communications\\
BlockAgents~\cite{chen2024blockagents} & \textbf{Defense:} Incorporate blockchain and PoT against byzantine attacks \\
Abdelnabi et al.~\cite{abdelnabi2025firewalls} & \textbf{Defense:} A multi-layer agent firewall\\
\bottomrule
\end{tabular}
}
\end{table}


\subsection{Privacy}\label{subsec:privacy}
The widespread use of LLMs in multi-agent systems has also raised several privacy concerns. These issues are mainly caused by the memory capacity of LLMs, which may lead to the leakage of private information during conversations or when completing tasks. In addition, LLM agents are vulnerable to attacks involving model and prompt theft, along with other forms of intellectual property theft. This section explores the privacy threats posed by \textbf{LLM Memorization Vulnerabilities} and \textbf{LLM Intellectual Property Exploitation} emphasizing the importance of ensuring the safe and secure deployment of LLMs in collaborative environments. Additionally, it discusses potential countermeasures to mitigate these risks.

\subsubsection{LLM Memorization Vulnerabilities}
It has been shown that LLMs are able to generate text similar to humans. However, such generated text may be retained training data, which poses serious privacy protection issues. These risks are particularly severe in multi-agent systems, where LLMs may leak sensitive information when collaborating to solve complex tasks. This section explores the privacy threats posed by LLM memory and discusses protection measures against these threats.

\paratitle{Data Extraction Attacks.} 
They exploit the memory capacity of LLMs to extract sensitive information from training data. Carlini et al.~\cite{carlini2021extracting} show that an attacker can extract personally identifiable information (PII) such as name, email, and phone number from a GPT-2 model through specific queries. The risk of data extraction increases with model size, frequency of repeated data, and context length~\cite{carlini2022quantifying}. Huang et al.~\cite{huang2022large} further study data extraction attacks against pre-trained LLMs such as GPT-neo, highlighting the feasibility of such attacks in practical applications.

\paratitle{Member Inference Attacks.} 
Their purpose is to determine whether a particular data sample has been part of the LLM training data. Mireshghallah et al.~\cite{mireshghallah2022quantifying} empirically analyze the vulnerability of fine-tuned LLMs to membership inference attacks and find that fine-tuning the model head makes it more vulnerable to such attacks. Fu et al.~\cite{fu2023practical} propose a self-calibrated membership inference attack method based on probability changes, which provides a more reliable membership signal through these variations. This type of attack is particularly dangerous in multi-agent systems, as the training data may originate from multiple sources of sensitive information. In response to these risks, protection strategies such as differential privacy (DP) and knowledge distillation have been developed~\cite{hoory2021learning,kang2023knowledge}.

\paratitle{Attribute Inference Attacks.} 
The goal of attribute inference attacks is to infer a certain feature or characteristic of a data sample using training data. To confirm the existence of sensitive attribute inference in LLMs, Pan et al.~\cite{pan2020privacy} conduct an in-depth study of privacy issues related to attribute inference attacks in LLMs. Wang et al.~\cite{wang2024property} study attribute existence inference attacks on generative models and find that most generative models are vulnerable to such attacks.

\paratitle{Protective Measures.}
Several protective strategies have been proposed to reduce the chance of LLM memorization. Data cleaning strategies can successfully reduce the risk of memorization by locating and eliminating sensitive information in training data~\cite{kandpal2022deduplicating}. Another effective way to minimize privacy leakage is to introduce differential privacy noise into model gradients and training data~\cite{hoory2021learning} during pre-training and fine-tuning. Knowledge distillation techniques have become an intuitive means of privacy protection by transferring knowledge from private teacher models to public student models~\cite{kang2023knowledge}. In addition, privacy leakage detection tools such as ProPILE can help service providers assess the extent of their PII leakage before deploying LLM agents~\cite{kim2023propile}.

\subsubsection{LM Intellectual Property Exploitation}

LLM agents are subject to memory concerns as well as privacy risks associated with intellectual property (IP), such as model theft and prompt theft. These attacks put both individuals and organizations at serious danger by taking advantage of the LLMs's economic value and signaling. 

\paratitle{Model Stealing Attacks.}
Model theft attacks attempt to extract model information (such as parameters or hyperparameters) by querying the model and observing its responses. Krishna et al.~\cite{krishna2019thieves} show that an attacker can steal information from language models such as BERT through multiple queries without accessing the original training data. Naseh et al.~\cite{naseh2023stealing} demonstrate that attackers can steal the types and hyperparameters of LLM decoding algorithms at a low cost. Li et al.~\cite{li2024extracting} investigate the feasibility of extracting specialized code from LLMs, highlighting the risk of model theft in multi-agent systems. In response to these attacks, protective measures such as model watermarking~\cite{kirchenbauer2023watermark} and blockchain-based IP authentication~\cite{lin2024blockchain} have been proposed. 

\paratitle{Prompt Stealing Attacks.}
Prompt theft attacks involve inferring original hints from generated content that may have significant business value. Shen et al.~\cite{shen2024prompt} conduct the first study of prompt stealer attacks against text-to-image generation models and propose an effective attack method called PromptStealer. Sha et al.~\cite{sha2024prompt} extend this study to LLMs, using a parameter extractor to determine the properties of the original prompt. Hui et al.~\cite{hui2024pleak} propose PLEAK, a closed-box prompt extraction framework that extracts system prompts for LLM applications by optimizing adversarial queries. To prevent prompt theft, adversarial samples have been proposed as an effective method to obstruct attackers from inferring the original prompt by introducing disturbance to the generated content~\cite{shen2024prompt}. 

The privacy challenges for LLM agents are multifaceted, ranging from memory threats to risks related to intellectual property. As LLMs continue to evolve, robust privacy protection technologies must be developed to mitigate these privacy risks while ensuring that LLMs play an effective role in multi-agent systems.

\begin{table}[t]
\caption{Summary of privacy threats and countermeasures in LLM agents.}
\label{tab:privacy_literature}
\centering
\resizebox{0.49\textwidth}{!}
{
\begin{tabular}{ll}
\toprule
\textbf{Reference} & \textbf{Description} \\
\midrule
\multicolumn{2}{c}{\textbf{LM Memorization Vulnerabilities}} \\
\midrule
Carlini et al.~\cite{carlini2021extracting} & \textbf{Attack:} Data Extraction \\
Huang et al.~\cite{huang2022large} & \textbf{Attack:} Data Extraction on Pretrained LLMs \\
Mireshghallah et al.~\cite{mireshghallah2022quantifying} & \textbf{Attack:} Membership Inference on Fine-Tuned LLMs \\
Fu et al.~\cite{fu2023practical} & \textbf{Attack:} Self-Calibrated Membership Inference \\
Pan et al.~\cite{pan2020privacy} & \textbf{Attack:} Attribute Inference in General-Purpose LLMs \\
Wang et al.~\cite{wang2024property} & \textbf{Attack:} Property Existence Inference in Generative Models \\
Kandpal et al.~\cite{kandpal2022deduplicating} & \textbf{Defense:} Data Sanitization to Mitigate Memorization \\
Hoory et al.~\cite{hoory2021learning} & \textbf{Defense:} Differential Privacy for Pre-Trained LLMs \\
Kang et al.~\cite{kang2023knowledge} & \textbf{Defense:} Knowledge Distillation for Privacy Preservation \\
Kim et al.~\cite{kim2023propile} & \textbf{Defense:} Privacy Leakage Assessment Tool \\
\midrule
\multicolumn{2}{c}{\textbf{LM Intellectual Property Exploitation}} \\
\midrule
Krishna et al.~\cite{krishna2019thieves} & \textbf{Attack:} Model Stealing via Query APIs \\
Naseh et al.~\cite{naseh2023stealing} & \textbf{Attack:} Stealing Decoding Algorithms of LLMs \\
Li et al.~\cite{li2024extracting} & \textbf{Attack:} Extracting Specialized Code Abilities from LLMs \\
Shen et al.~\cite{shen2024prompt} & \textbf{Attack:} Prompt Stealing in Text-to-Image Models \\
Sha et al.~\cite{sha2024prompt} & \textbf{Attack:} Prompt Stealing in LLMs \\
Hui et al.~\cite{hui2024pleak} & \textbf{Attack:} Closed-Box Prompt Extraction \\
Kirchenbauer et al.~\cite{kirchenbauer2023watermark} & \textbf{Defense:} Model Watermarking for IP Protection \\
Lin et al.~\cite{lin2024blockchain} & \textbf{Defense:} Blockchain for IP Verification \\
\bottomrule
\end{tabular}
}
\end{table}
\subsection{Social Impact and Ethical Concerns}\label{subsec:ethics}

LLM agents profoundly impact society, driving automation, industrial innovation, and productivity gains. However, ethical concerns remain. The following section explores both the benefits and challenges associated with their use. We summarize the content in Table~\ref{tab:social_impacts}.

\subsubsection{Benefits to Sociaty}

LLM agents have significantly impacted human society, offering numerous benefits across various domains.

\paratitle{Automation Enhancement.}
LLM agents have found applications across diverse fields, including healthcare, biomedicine, law, and education~\cite{bommasani2021opportunities}. By automating labor-intensive tasks, they reduce time costs and enhance efficacy. In healthcare, for example, they assist in interpreting clinical symptoms, explaining lab results, and even drafting medical documentation. In legal and educational settings, they streamline administrative work, generate summaries, and provide instant, context-aware responses~\cite{bommasani2021opportunities, floridi2020gpt, touvron2023llama}. Their ability to alleviate repetitive workloads allows professionals to focus on more complex, high-stake tasks, ultimately improving productivity and accessibility across industries.

\paratitle{Job Creation and Workforce Transformation.}
While researchers acknowledge the potential for AI agents to replace human jobs and disrupt the job market~\cite{bommasani2021opportunities}, others argue that their advancements will reshape workforce demands~\cite{tadas2024redefining}. 
The rise of LLM agents is transforming the job market, not only expanding technical roles such as machine learning engineers and data scientists but also driving demand for managerial positions like AI project managers and business strategists. Given their growing economic impact, governments are encouraged to support AI-focused training programs to equip individuals for this evolving landscape.
Unlike LLMs, which often require specialized expertise to use effectively, LLM agents are designed for accessibility, attracting a broader user base and enabling wider applications across various industries. As a result, their societal impact is expected to surpass that of LLMs or other AI models alone, bringing both challenges and unprecedented opportunities.

\paratitle{Enhance Information Distribution.}
Businesses reliant on large-scale text generation, such as online advertising, benefit significantly from LLM agents. However, their misuse is a growing concern, particularly regarding the proliferation of fake news and misinformation~\cite{floridi2020gpt,touvron2023llama}.
Beyond accelerating advertisement distribution, enhanced information dissemination offers broader societal benefits. For instance, the global shortage of patient, experienced, and knowledgeable teachers has long been a challenge. LLM agents introduce transformative solutions, such as intelligent online tutoring systems, revolutionizing education accessibility~\cite{moore2023empowering}.

\subsubsection{Ethical Concerns}

Although LLM agents bring numerous benefits to society, they also pose potential risks that cannot be overlooked. These challenges raise significant ethical concerns, including bias in decision-making, misinformation propagation, and privacy issues, highlighting the need for responsible development and regulation.

\paratitle{Bias and Discrimination.}
LLM agents inherently inherit biases present in their training datasets and may even amplify them during the learning process, leading to skewed outputs and reinforcing existing stereotypes~\cite{liu2025culturevlm}. Recognizing this issue, many existing works have implemented strategies to mitigate harmful content generation. These methods include filtering sensitive topics, applying reinforcement learning with human feedback, and refining model training processes to promote fairness and reduce bias~\cite{bommasani2021opportunities,floridi2020gpt,touvron2023llama}.
The pursuit of fairness has become a critical focus in studies on LLM agents, as researchers strive to develop models that minimize bias, promote inclusivity, and ensure ethical AI deployment in real-world applications~\cite{henderson2023foundation,lemley2020fair}.

\paratitle{Accountability.}
Despite efforts to mitigate toxic content in LLM agents, the risk of harmful outputs persists~\cite{touvron2023llama,floridi2020gpt,oh2024uniguard}. Accountability remains a key challenge, as documented datasets provide limited oversight, while vast amounts of undocumented data can be easily integrated into training. Rigorous dataset documentation is essential, despite its costs~\cite{bender2021dangers}. Additionally, proper governance frameworks are necessary to ensure accountability in LLM agents~\cite{brundage2020toward,ganguli2022predictability}.

\paratitle{Copyright.}
Copyright concerns are closely linked to privacy and accountability. Some argue that AI should adhere to the same legal and ethical standards as humans, ensuring fair use and intellectual property protection~\cite{lemley2020fair}. Many creators oppose their work being used to train models that could replace them, yet the absence of clear regulations and the growing demand for data lead to widespread misuse~\cite{deng2024deconstructing}. This issue is often underestimated and requires urgent attention, as it threatens human creators, increases the prevalence of AI-generated content over human-produced work in certain domains, and risks content degradation, particularly when large AI models are increasingly trained on AI-generated data~\cite{shumailov2024ai}.
Addressing these issues is particularly crucial in the use of LLM agents, where users often lack direct awareness of the training data sources. This opacity increases the risk of unintended consequences, as individuals may unknowingly rely on models trained on controversial datasets, potentially resulting in reputational harm or even legal repercussions.



\paratitle{Others.}
Some ethical concerns in the use of LLM agents, such as privacy~\cite{bommasani2021opportunities,weidinger2021ethical,xiao2024large}, data manipulation~\cite{alber2025medical}, and misinformation~\cite{floridi2020gpt,jin2022towards}, are so critical that we provide a thorough discussion in Sections~\ref{subsec:security_1},~\ref{subsec:security_2} and~\ref{subsec:privacy}.
Beyond these, additional ethical concerns remain.
One major issue is that LLM agents lack true semantic and contextual understanding, relying purely on statistical word associations. This limitation is often misinterpreted and overestimated, leading to undue reliance on these models~\cite{floridi2020gpt}, especially when their behavior may not align well with human intentions~\cite{shen2023large}.
Moreover, concerns have been raised about the significant carbon footprint of LLM agents, posing environmental challenges~\cite{luccioni2023estimating}, alongside the high computational costs associated with training large models~\cite{strubell2020energy}.

\begin{table}[t]
\caption{Overview of Social Impacts and Ethical Considerations in LLM Agents.}
\label{tab:social_impacts}
\centering
\resizebox{0.49\textwidth}{!}
{
\begin{tabular}{ll}
\toprule
\textbf{Impact} & \textbf{Reference} \\
\midrule
\multicolumn{2}{c}{\textbf{Benefits to Society}} \\
\midrule
Automation Enhancement & Foundation Models~\cite{bommasani2021opportunities}, GPT-3~\cite{floridi2020gpt}, LLaMA~\cite{touvron2023llama}   \\
Workforce Transformation & Foundation Models~\cite{bommasani2021opportunities}, Redefining Work~\cite{tadas2024redefining}  \\
Enhance Information Distribution & GPT-3~\cite{floridi2020gpt}, LLaMa~\cite{touvron2023llama}, Empower Online Education~\cite{moore2023empowering}  \\
\midrule
\multicolumn{2}{c}{\textbf{Ethical Concerns}} \\
\midrule
Bias and Discrimination & 
Fair Use~\cite{henderson2023foundation}, Fair Learning~\cite{lemley2020fair} \\
Accountability & Stochastic Parrots~\cite{bender2021dangers}, Governance~\cite{brundage2020toward,ganguli2022predictability}  \\
Copyright & Fair Learning~\cite{lemley2020fair}, Ethics of LLMs~\cite{deng2024deconstructing}, AI collapse~\cite{shumailov2024ai} \\
Data Privacy & Foundation Models~\cite{bommasani2021opportunities}, Ethical and Social Risks~\cite{weidinger2021ethical} \\
Manipulation \& Misinformation & Data-Poisoning Attacks~\cite{alber2025medical} \\
Others & Overreliance~\cite{floridi2020gpt}, Alignment~\cite{shen2023large}, Carbon Footprint~\cite{luccioni2023estimating}, Expenses~\cite{strubell2020energy} \\
\bottomrule
\end{tabular}
}
\end{table}

\section{Applications}\label{sec:app}

The versatility of LLM agents has led to their adoption across diverse domains, transforming how complex tasks are approached in both research and industry settings. This section surveys the broad spectrum of LLM agent applications, from accelerating scientific discovery~(Section~\ref{subsec:scientific}) to enhancing interactive gaming experiences~(Section~\ref{subsec:gaming}), modeling complex social phenomena~(Section~\ref{subsec:social}), and boosting productivity~(Section~\ref{subsec:productivity}). These applications demonstrate how the integration of LLM-based agent systems enables enhanced problem-solving capabilities through specialized knowledge application, multi-agent collaboration, and human-AI interaction paradigms.

\subsection{Scientific Discovery}\label{subsec:scientific}
By leveraging multiple specialized LLM agents that communicate and coordinate, LLM-based multi-agent AI systems can combine diverse expertise, access external tools, and decompose tasks, thereby extending the capabilities of single LLMs~\cite{zhou2024awesome,AAAIpp}. 
In this part, we survey advances in applying LLM-driven multi-agent systems to scientific research over the past three years. 

\subsubsection{Agentic AI Across Scientific Disciplines}
LLM-based multi-agent systems are increasingly applied across scientific disciplines to emulate human collaborative workflows and tackle complex, interdisciplinary problems that require diverse knowledge and skills. 
For example, the SciAgents~\cite{sciagents} framework uses distinct LLM agents such as ``Ontologist,'' ``Scientist,'' and ``Critic'' to collectively generate and refine scientific hypotheses. 
Centered on an ontological knowledge graph that encodes relationships between scientific concepts, SciAgents orchestrates ChatGPT-4-based agents to generate novel research ideas and experimental plans.
In a case study on bio-inspired materials, one agent generated a proposal to integrate silk with novel pigments; another agent suggested simulation experiments to test the idea, and a critical agent identified weaknesses and prompted improvements. 
Beyond hypothesis generation, LLM-based agents are being used to plan and execute experimental research. 
For instance, Curie~\cite{kon2025curierigorousautomatedscientific} developed an AI agent framework for rigorous automated experimentation.
In Curie, an Architect agent first designs high-level experimental plans to answer a scientific question, then multiple Technician agents carry out specific experimental steps. 
In tests on questions derived from computer science research papers, Curie's structured multi-agent approach improved the correctness of experimental results, outperforming more straightforward prompt-based automation by a notable margin.
This indicates that multi-agent systems can bring not just creativity but also discipline and reliability.
Aside from scientific findings, LLMs are also used to improve the generation pipeline of academic works.
AgentReview~\cite{jin2024agentreview} proposes an LLM-agent-based framework for simulating academic peer review processes, offering valuable insights to improve the design of evaluation protocols for academic papers.

\subsubsection{Agentic AI in Chemistry, Materials Science and Astronomy}
Due to the abundance of digital tools and data in these fields, chemistry, materials science, and Astronomy have been early adopters of LLM-based agentic AI.
In the chemistry domain, ChemCrow~\cite{bran2023chemcrowaugmentinglargelanguagemodels} exemplifies an LLM-driven chemistry agent designed to foster scientific advancement by bridging the gap between experimental and computational chemistry. 
ChemCrow integrates an LLM with a suite of 18 expert-designed chemistry tools, such as molecule property predictors, reaction planners and databases, enabling it to plan and execute chemical syntheses autonomously. 
Materials science problems, which often span multiple scales and modalities (from atomic simulations to empirical data), also benefit from multi-agent AI.
AtomAgents~\cite{ghafarollahi2024atomagentsalloydesigndiscovery} framework is a physics-aware multi-agent system for automating alloy design. 
In this system, a Planner agent (GPT-4) decomposes a complex materials design challenge into a sequence of tasks, which are then verified by a Critic agent and delegated to specialist modules. 
Similar principles are being applied in physics and astronomy. 
For example, an AI copilot agent has been developed for the Cherenkov Telescope Array in astronomy~\cite{kostunin2025aiagentsgroundbasedgamma}, using an instruction-tuned LLM to autonomously manage telescope configuration databases and even generate code for data analysis workflows. 
Although still experimental, these efforts indicate that LLM-based agents could soon be used in physics labs and astronomical observatories. They could handle routine decision-making and free human scientists to focus on high-level insights.

\subsubsection{Agentic AI in Biology}\label{subsec:biology}

The life sciences are likewise beginning to embrace LLM-based multi-agent systems for hypothesis generation and data analysis~\cite{qi2024largelanguagemodelsbiomedical}. 
One notable direction is using LLM agents to propose biological experiments or interpret multi-omics data.
BioDiscoveryAgent~\cite{roohani2024biodiscoveryagent} proposed an AI agent to design genetic perturbation experiments in molecular biology. 
By parsing literature and gene databases, an LLM agent can suggest which gene knockouts or edits might elucidate a certain biological pathway.
Another system, GeneAgent~\cite{wang2024geneagentselfverificationlanguageagent}, uses a self-refinement loop to discover gene associations from biomedical databases, improving the reliability of findings by cross-checking against known gene sets. 
RiGPS~\cite{xiao2025knowledge} developed a multi-agent system with an experiment-based self-verified reinforcement learning framework, enhancing the biomarker identification task in the single-cell dataset. 
BioRAG~\cite{wang2024biorag} developed a multi-agent-based RAG system to handle biology-related QA, where several agents are designed to retrieve information using multiple tools, and one agent is specifically used to self-evaluate the retrieval results. 
These examples illustrate the methodology of self-questioning or self-verification in multi-agent AI: one or more agents propose a scientific insight, and another evaluates its plausibility with known knowledge, thereby reducing errors. 

\subsubsection{Agentic AI in Scientific Dataset Construction}
Multi-agent systems also accelerate the construction of scientific datasets. 
For instance, PathGen-1.6M~\cite{sun2024pathgen16m16millionpathology} generated a massive pathology image dataset via multi-agent collaboration, where multiple AI models played different roles: one vision model scanned whole-slide histology images to select representative regions, another (an LLM or multimodal model) generated descriptive captions for each region, and additional agents iteratively refined the captions for accuracy. 
KALIN~\cite{cai2025knowledge} developed a multi-agent collaborative framework to generate a high-quality domain LLM training corpus. 
Specifically, two distinct LLMs are trained to generate scientific questions with input chunked research articles as context. 
Then, KAILIN utilizes a knowledge hierarchy to self-evaluate the alignment of generated questions with the input context, then self-evolving to more in-depth questions. 
GeneSUM~\cite{chen2024genesum} is designed to maintain the gene function description knowledge dataset automatically. 
Specifically, a single description agent serves as a reader for gene ontology, a retrieval agent functions as a reader for related literature, and a summarization agent acts as the generator. 
GeneSUM thus can automatically read emerging gene-function-related research articles and renew the database of gene function descriptions. 
These approaches demonstrate a virtuous cycle: AI systems can consume scientific data and create it, improving the next generation of models.

\subsubsection{Agentic AI in Medical}
Digitization of medical records~\cite{keshavjee2006best,ye2023needed} brings great potential in applying agentic AI in medical service.
One line of research has created simulated clinical environments in which autonomous doctors and patient agents interact.
AgentHospital~\cite{li2024agenthospital} is a virtual hospital populated by LLM-driven doctors, nurses, and patient agents, modeling the full cycle of care from triage to diagnosis to treatment.
In this system, each patient agent presents symptoms, and doctor agents must converse with the patient, order virtual tests, make a diagnosis, and prescribe treatment.
In parallel, other work focuses on aligning multi-agent AI directly with clinical decision support in real scenarios.
ClinicalLab~\cite{yan2024clinicallab}introduced a comprehensive benchmark and an agent for multi-department medical diagnostics, which involved 150 diseases across 24 medical specialties, reflecting the breadth of knowledge required in hospital settings. 
Multi-agent systems can also enhance conversational applications by introducing roles and simulations.
AIPatient~\cite{yu2024aipatientsimulatingpatientsehrs} is a system that creates realistic patient simulators powered by LLMs.
It leverages a structured knowledge graph of medical information as a source of ground truth about a patient's conditions, and a Reasoning RAG workflow that allows the patient agent to retrieve relevant details and respond to a doctor's questions in a convincing manner. 
Medical imaging is another domain ripe for multi-agent AI integration.
For instance, CXR-Agent~\cite{sharma2024cxr} uses a vision-language model together with an LLM to interpret chest X-rays and generate radiology reports with uncertainty estimates.
MedRAX~\cite{fallahpour2025medraxmedicalreasoningagent} integrates several specialized tools, such as an optical character reader for reading prior reports, a segmentation model for highlighting image regions, and an LLM for clinical reasoning, to solve complex chest X-ray cases that require referring to patient history and imaging simultaneously. 
Evaluations of these approaches on standard chest X-ray benchmarks~\cite{lee2024comparative} showed that it could achieve diagnostic accuracy on par with state-of-the-art standalone models while also providing an uncertainty score that correlates with its correctness.
In summary, the multi-agent paradigm in medicine holds promise for improving AI reliability by introducing redundancy, specialization, and oversight. However, it also complicates the system, requiring rigorous validation.

\begin{table}[t]
\caption{Overview of Applications in LLM Agents.}
\label{table:application}
\centering
\setlength{\tabcolsep}{1pt}
\resizebox{0.5\textwidth}{!}{
\begin{tabular}{lcc}
\toprule
Method & Domain & Core Idea \\
\midrule
\multicolumn{3}{c}{\textbf{Scientific Discovery}} \\
\midrule
SciAgents~\cite{sciagents}& General Sciences & Collaborative hypothesis generation \\
Curie~\cite{kon2025curierigorousautomatedscientific}& General Sciences & Automated experimentation \\
ChemCrow~\cite{bran2023chemcrowaugmentinglargelanguagemodels}& Chemistry & Tool-augmented synthesis planning \\
AtomAgents~\cite{ghafarollahi2024atomagentsalloydesigndiscovery}& Materials Science & Physics-aware alloy design \\
D. Kostunin el al~\cite{kostunin2025aiagentsgroundbasedgamma}& Astronomy & Telescope configuration management \\
BioDiscoveryAgent~\cite{roohani2024biodiscoveryagent}& Biology & Genetic perturbation design \\
GeneAgent~\cite{wang2024geneagentselfverificationlanguageagent}& Biology & Self-verifying gene association discovery \\
RiGPS~\cite{xiao2025knowledge}& Biology & Biomarker identification \\
BioRAG~\cite{wang2024biorag}& Biology & Biology-focused retrieval augmentation \\
PathGen-1.6M~\cite{sun2024pathgen16m16millionpathology} & Medical Dataset & Pathology image dataset generation \\
KALIN~\cite{cai2025knowledge}& Biology Dataset & Scientific question corpus generation \\
GeneSUM~\cite{chen2024genesum}& Biology Dataset & Gene function knowledge maintenance \\
AgentHospital~\cite{li2024agenthospital} & Medical & Virtual hospital simulation \\
ClinicalLab~\cite{yan2024clinicallab}& Medical & Multi-department diagnostics \\
AIPatient~\cite{yu2024aipatientsimulatingpatientsehrs} & Medical & Patient simulation \\ 
CXR-Agent~\cite{sharma2024cxr}& Medical & Chest X-ray interpretation \\
MedRAX~\cite{fallahpour2025medraxmedicalreasoningagent}& Medical & Multimodal medical reasoning \\
\midrule
\multicolumn{3}{c}{\textbf{Gaming}} \\
\midrule
ReAct~\cite{yao2023react} & Game Playing & Reasoning and acting in text environments \\
Voyager~\cite{wang2023voyager} & Game Playing & Lifelong learning in Minecraft \\
ChessGPT~\cite{feng2023chessgpt} & Game Playing & Chess gameplay evaluation \\
GLAM~\cite{carta2023grounding} & Game Playing & Reinforcement learning in text environments \\ 
CALYPSO~\cite{zhu2023calypso} & Game Generation & Narrative generation for D\&D \\
GameGPT~\cite{chen2023gamegpt} & Game Generation & Automated game development \\
Sun et al.~\cite{sun2023language} & Game Generation & Interactive storytelling experience \\
\midrule
\multicolumn{3}{c}{\textbf{Social Science}} \\
\midrule
Econagent~\cite{li2023econagent} & Economy & Economic decision simulation \\
TradingGPT~\cite{li2023tradinggpt} & Economy & Financial trading simulation \\
CompeteAI~\cite{zhao2024competeai} & Economy & Market competition modeling \\
Ma et al.~\cite{ma2024understanding} & Psychology & Mental health support analysis \\
Zhang et al.~\cite{zhang2024exploring} & Psychology & Social behavior simulation \\
TE~\cite{aher2023using} & Psychology & Psychological experiment simulation \\
Generative agents~\cite{park2023generative} & Social Simulation & Human behavior emulation \\
Liu et al.~\cite{liu2024training} & Social Simulation & Learning from social interactions \\
S$^3$~\cite{gao2023s3} & Social Simulation & Social network behavior modeling \\
\midrule
\multicolumn{3}{c}{\textbf{Productivity Tools}} \\
\midrule
SDM~\cite{dong2024self} & Software Development & Self-collaboration for code generation \\
ChatDev~\cite{qian2024communicative} & Software Development & Chat-powered development framework \\
MetaGPT~\cite{hong2024metagpt} & Software Development & Meta-programming for collaboration \\
Agent4Rec~\cite{zhang2024generative} & Recommender Systems & User behavior modeling \\
AgentCF~\cite{zhang2024agentcf} & Recommender Systems & User-item interaction modeling \\
MACRec~\cite{wang2024macrec} & Recommender Systems & Multi-agent recommendation \\
RecMind~\cite{wang2023recmind} & Recommender Systems & Knowledge-enhanced recommendation \\
\bottomrule
\end{tabular}}
\end{table}

\subsection{Gaming}\label{subsec:gaming}
The development of LLM agents offers an unprecedented opportunity in gaming, enabling agents to take on diverse roles and exhibit human-like decision-making skills in intricate game environments. Based on the different characteristics of the games and roles of the agent, the applications can be categorized into game playing and game generation.

\paratitle{Game Playing.} In role-playing games, LLM agents can assume various character roles, both as player-controlled characters and non-player characters (NPCs). ReAct~\cite{yao2023react} prompts LLMs to integrate reasoning and reflection into action generation, enhancing decision-making in the embodied environment. Voyager~\cite{wang2023voyager} introduces an LLM-powered lifelong learning agent in Minecraft that persistently explores the game world. ChessGPT~\cite{feng2023chessgpt} presents an autonomous agent on mixed game-language data to facilitate board state evaluation and chess gameplay. GLAM~\cite{carta2023grounding} builds an agent in the BabyAI-text environment, where a policy is used to select the next action, with training conducted through online reinforcement learning. 

\paratitle{Game Generation.} In game generation, LLMs are used to create dynamic and interactive game content. CALYPSO~\cite{zhu2023calypso} creates LLM agents as the assistants to help build a compelling
narrative to present in the context of playing Dungeons \& Dragons. GameGPT~\cite{chen2023gamegpt} leverages dual-agent collaboration and a hierarchical approach, using multiple internal dictionaries to automate and enhance the game development process. Sun et al.~\cite{sun2023language} create an interactive storytelling game experience in 1001 Nights, where instructive language models and image generation are combined to shape the narrative and world.



\subsection{Social Science}\label{subsec:social}
The application of LLM agents in social science has seen significant advancements, providing new opportunities for understanding and simulating complex human behaviors and interactions. These models facilitate insights into various domains, including economics, psychology and social simulation. Below, we explore how LLM agents are being applied across these three critical areas.

\paratitle{Economy.} In economics, LLM agents are utilized to analyze financial data and simulate financial activities. Econagent~\cite{li2023econagent} employs prompt engineering to create agents that mimic human-like decisions or macroeconomic simulations. TradingGPT~\cite{li2023tradinggpt} presents a multi-agent framework for financial trading, which simulates human decision processes by incorporating hierarchical memory structures and debate mechanisms with individualized trading profiles. CompeteAI~\cite{zhao2024competeai} leverages LLM agents to model a virtual town where restaurants and customers interact, providing insights consistent with sociological and economic theories. 

\paratitle{Psychology.} In psychological research, LLM agents are utilized to model human behavior with diverse traits and cognitive processes. Ma et al.~\cite{ma2024understanding} investigate the psychological effects and potential benefits of using LLM-based conversational agents for mental health support. Zhang et al.~\cite{zhang2024exploring} examine how LLM agents with unique traits and thought processes replicate human-like social behaviors, including conformity and majority influence. TE~\cite{aher2023using} uses LLM agents to simulate psychological experiments, potentially revealing consistent distortions in how language models replicate specific human behaviors.

\paratitle{Social Simulation.}
In societal simulation, LLM agents are employed to model complex societal behaviors. These simulations help in understanding real-world phenomena, such as social influence, information diffusion, and collective decision-making. Generative agents~\cite{park2023generative} introduce a multi-agent interaction model within an interactive sandbox environment, leveraging LLM agents to simulate realistic human behavior in a variety of contexts. Building on this, Liu et al.~\cite{liu2024training} introduce a training paradigm that enables LLMs to learn from these simulated social interactions involving multiple LLM agents. S$^3$~\cite{gao2023s3} develops an LLM-based multi-agent system to ensure the agents' behaviors closely mimic those of real humans within social networks.

\subsection{Productivity Tools}\label{subsec:productivity}
LLM agents are increasingly leveraged to boost productivity by automating diverse tasks, facilitating collaboration in solving complex problems, and optimizing efficiency across multiple domains. Below, we highlight their applications in software development and recommender systems.

\paratitle{Software Development.} Since software development involves multiple roles, such as product managers, developers, and testers, all working together to deliver high-quality products, LLM agents are increasingly being used to streamline various aspects of the process. SDM~\cite{dong2024self} introduces a self-collaboration framework that guides multiple LLM agents to work together on code generation tasks, enhancing their ability to tackle complex software development challenges collaboratively. ChatDev~\cite{qian2024communicative} proposes a chat-powered software development framework, where agents are guided on both what to communicate and how to communicate effectively. MetaGPT~\cite{hong2024metagpt} further incorporates human workflows (i.e., Standardized Operating Procedures) into LLM-powered multi-agent collaboration through a meta-programming approach to enhance coordination and streamline the collaborative process.

\paratitle{Recommender Systems.} In the realm of recommender systems, LLM agents are increasingly utilized to simulate user behaviors. Agent4Rec~\cite{zhang2024generative} employs LLM agents with integrated user profiling, memory, and action modules to model user behavior in recommender systems. AgentCF~\cite{zhang2024agentcf} treats both users and items as LLM agents, introducing a collaborative learning framework to model user-item interactions in recommender systems. MACRec~\cite{wang2024macrec} directly develops multiple agents to tackle the recommendation task. RecMind~\cite{wang2023recmind} employs LLM agents to incorporate external knowledge and carefully plans the utilization of tools for zero-shot personalized recommendations.

\section{Challenges and Future Trends}\label{sec:future}

Advancements in LLM-based multi-agent systems bring significant opportunities but also present pressing challenges in scalability, memory, reliability, and evaluation. 
This section outlines key obstacles and emerging trends shaping the future of agentic AI. 

\subsection{Scalability and Coordination} 
Scaling LLM-based multi-agent systems remains challenging due to high computational demands,  inefficiencies in coordination, and resource utilization~\cite{qian2024scaling,chan2023chateval}. 
Existing multi-agent frameworks, designed for lightweight agents like function calls and rule-based systems~\cite{rana2000scalability, deters2001scalable}, lack system-level optimization for LLM agents with billion-scale parameters~\cite{wu2023autogen}. 
Future directions include \emph{hierarchical structuring}, where high-level LLM agents delegate subtasks to specialized lower-level agents, and \emph{decentralized planning}, which enables agents to plan concurrently and synchronize periodically to mitigate bottlenecks. 
Advancements in robust communication protocols and efficient scheduling mechanisms are needed to enhance coordination, real-time decision-making, and system robustness~\cite{qian2024scaling, chan2023chateval}. 

\subsection{Memory Constraints and Long-Term Adaptation.} 
Maintaining coherence across multi-turn dialogues and the longitudinal accumulation of knowledge requires effective memory mechanisms~\cite{verma2024adaptagent}. However, as LLMs possess very limited effective context~\cite{jiang2023llmlingua, jin2024mm}, integrating sufficient historical information into prompts becomes challenging. This hinders the models' contextual awareness over extended interactions. 
Ensuring interaction continuity requires efficient memory scalability and relevance management~\cite{yao2024velo} beyond current practice such as vector databases, memory caches, context window management, and retrieval-augmented generation (RAG)~\cite{lewis2020retrieval}. 
Future directions include \emph{hierarchical memory architectures} that combine \emph{episodic memory} for short-term planning with \emph{semantic memory} for long-term retention, as well as autonomous knowledge compression~\cite{cheng2024xrag} to refine memory dynamically and enhance reasoning over extended interactions.

\subsection{Reliability and Scientific Rigor} 
LLMs, while knowledge-rich, are neither comprehensive nor up-to-date, thus potentially unsuitable as standalone replacements for structured databases. Their stochastic nature makes outputs highly sensitive to minor variations in prompts~\cite{jin2024better}, causing hallucinations~\cite{agarwal2024medhalu} and compounding uncertainty in multi-agent systems, such as agentic frameworks for medical applications and autonomous scientific discovery~\cite{lu2024ai}, where unreliable outputs can mislead high-stake decision-making. 
Addressing these challenges necessitates the development of rigorous validation mechanisms and structured verification pipelines, including \emph{knowledge-graph-based verification}, where outputs are cross-checked against structured databases~\cite{agrawal2024can}, and \emph{cross-referencing via retrieval}, which grounds responses in cited source like web pages as in WebGPT~\cite{nakano2021webgpt}.  
Along this direction, future work can explore LLMs capable of direct citation generation, as well as up-to-date and comprehensive knowledge sources readily available for LLM applications. 
Meanwhile, in high-stakes domains like healthcare, law, or scientific research, pure automation remains risky. \emph{AI-human verification loops} are becoming standard for ensuring safety, reliability, and accountability~\cite{agarwal2024medhalu}. 
Future works can enhance cross-referencing mechanisms~\cite{gao2023enabling}, self-consistency~\cite{wangself}, and standardized AI auditing frameworks, such as fact-checking logs, to improve accountability. For example, one critical challenge is determining optimal intervention points amid the vast scale of LLM-generated content. 

\subsection{Multi-turn, Multi-agent Dynamic Evaluation} 
Traditional AI evaluation frameworks, designed for static datasets and single-turn tasks, fail to capture the complexities of LLM agents in dynamic, multi-turn, and multi-agent environments~\cite{verma2024adaptagent}. 
Current benchmarks primarily assess task execution such as code completion~\cite{zhou2023codebertscore,wang2023execution} and dialogue generation~\cite{lykov2023llm} in isolated settings, overlooking emergent agent behaviors, long-term adaptation, and collaborative reasoning that unfold across multi-turn interactions. 
Additionally, static benchmarks struggle to keep pace with evolving LLM capabilities~\cite{zhudyval}. Concerns persist regarding potential data contamination, where model performance may stem from memorization rather than genuine reasoning. 
Future research should focus on dynamic evaluation methodologies, integrating multi-agent interaction scenarios, structured performance metrics, and adaptive sample generation algorithms~\cite{zhu2024dynamic} to create more robust and reliable assessment frameworks.

\subsection{Regulatory Measures for Safe Deployment}
As agentic AI systems gain autonomy, regulatory frameworks must evolve to ensure accountability, transparency, and safety. A key challenge is mitigating algorithmic bias--agents may inadvertently discriminate based on gender, age, ethnicity, or other sensitive attributes, often in ways imperceptible to developers~\cite{yi2023unpacking,liu2025culturevlm}. 
Addressing this requires standardized auditing protocols to systematically identify and correct biases, alongside traceability mechanisms that log decision-making pathways and model confidence for post-hoc accountability. Future work can explore multidisciplinary approaches combining fairness-aware training pipelines with legal and ethical safeguards. Collaboration between policymakers, researchers, and industry stakeholders will be critical to ensuring AI-driven systems operate safely and equitably in alignment with societal values~\cite{wang2023evaluating}.

\subsection{Role-playing Scenarios}
LLM agents can simulate roles such as researchers, debators, and instructors~\cite{chan2023chateval,ChatArena}, but their effectiveness is constrained by training data limitations and an incomplete understanding of human cognition~\cite{wang2023evaluating,yao2024value}. Since LLMs are predominantly trained on web-based corpora, they struggle to emulate roles with insufficient representation online~\cite{nguyen2024large} and often produce conversations lacking diversity~\cite{jin2024agentreview}. Future research should focus on enhancing role-play fidelity by improving multi-agent coordination, incorporating real-world reasoning frameworks, and refining dialogue diversity to better support complex human-AI interactions.




\section{Conclusion}\label{sec:con}

This survey has presented a systematic taxonomy of LLM agents, deconstructing their methodological components across construction, collaboration, and evolution dimensions. We have advanced a unified architectural perspective that bridges individual agent design principles with multi-agent collaborative systems—an approach that distinguishes our work from previous surveys. Despite remarkable progress, significant challenges remain, including scalability limitations, memory constraints, reliability concerns, and inadequate evaluation frameworks. Looking forward, we anticipate transformative developments in coordination protocols, hybrid architectures, self-supervised learning, and safety mechanisms that will enhance agent capabilities across diverse domains. By providing this foundational understanding and identifying promising research directions, we hope to contribute to the responsible advancement of LLM agent technologies that may fundamentally reshape human-machine collaboration.

\IEEEdisplaynontitleabstractindextext

%




\ifCLASSOPTIONcaptionsoff
  \newpage
\fi



%

\bibliographystyle{IEEEtran}
\bibliography{rec,ref_application_sci_med}

\begin{thebibliography}{100}
\providecommand{\url}[1]{#1}
\csname url@samestyle\endcsname
\providecommand{\newblock}{\relax}
\providecommand{\bibinfo}[2]{#2}
\providecommand{\BIBentrySTDinterwordspacing}{\spaceskip=0pt\relax}
\providecommand{\BIBentryALTinterwordstretchfactor}{4}
\providecommand{\BIBentryALTinterwordspacing}{\spaceskip=\fontdimen2\font plus
\BIBentryALTinterwordstretchfactor\fontdimen3\font minus \fontdimen4\font\relax}
\providecommand{\BIBforeignlanguage}[2]{{%
\expandafter\ifx\csname l@#1\endcsname\relax
\typeout{** WARNING: IEEEtran.bst: No hyphenation pattern has been}%
\typeout{** loaded for the language `#1'. Using the pattern for}%
\typeout{** the default language instead.}%
\else
\language=\csname l@#1\endcsname
\fi
#2}}
\providecommand{\BIBdecl}{\relax}
\BIBdecl

\bibitem{xi2025rise}
Z.~Xi, W.~Chen, X.~Guo, W.~He, Y.~Ding, B.~Hong, M.~Zhang, J.~Wang, S.~Jin, E.~Zhou \emph{et~al.}, ``The rise and potential of large language model based agents: A survey,'' \emph{Science China Information Sciences}, vol.~68, no.~2, p. 121101, 2025.

\bibitem{wooldridge1995intelligent}
M.~Wooldridge and N.~R. Jennings, ``Intelligent agents: Theory and practice,'' \emph{The knowledge engineering review}, vol.~10, no.~2, pp. 115--152, 1995.

\bibitem{zheng2024large}
D.~Zheng, M.~Lapata, and J.~Z. Pan, ``Large language models as reliable knowledge bases?'' \emph{arXiv preprint arXiv:2407.13578}, 2024.

\bibitem{lotfi2023non}
S.~Lotfi, M.~Finzi, Y.~Kuang, T.~G. Rudner, M.~Goldblum, and A.~G. Wilson, ``Non-vacuous generalization bounds for large language models,'' \emph{arXiv preprint arXiv:2312.17173}, 2023.

\bibitem{fei2024multimodal}
H.~Fei, Y.~Yao, Z.~Zhang, F.~Liu, A.~Zhang, and T.-S. Chua, ``From multimodal llm to human-level ai: Modality, instruction, reasoning, efficiency and beyond,'' in \emph{COLING}, 2024, pp. 1--8.

\bibitem{huang2022towards}
J.~Huang and K.~C.-C. Chang, ``Towards reasoning in large language models: A survey,'' \emph{arXiv preprint arXiv:2212.10403}, 2022.

\bibitem{wang2024tool}
C.~Wang, W.~Luo, Q.~Chen, H.~Mai, J.~Guo, S.~Dong, Z.~Li, L.~Ma, S.~Gao \emph{et~al.}, ``Tool-lmm: A large multi-modal model for tool agent learning,'' \emph{arXiv e-prints}, pp. arXiv--2401, 2024.

\bibitem{zhang2024survey}
Z.~Zhang, X.~Bo, C.~Ma, R.~Li, X.~Chen, Q.~Dai, J.~Zhu, Z.~Dong, and J.-R. Wen, ``A survey on the memory mechanism of large language model based agents,'' \emph{arXiv preprint arXiv:2404.13501}, 2024.

\bibitem{zhao2023depth}
P.~Zhao, Z.~Jin, and N.~Cheng, ``An in-depth survey of large language model-based artificial intelligence agents,'' \emph{arXiv preprint arXiv:2309.14365}, 2023.

\bibitem{sumers2023cognitive}
T.~Sumers, S.~Yao, K.~Narasimhan, and T.~Griffiths, ``Cognitive architectures for language agents,'' \emph{TMLR}, 2023.

\bibitem{hu2024survey}
S.~Hu, T.~Huang, F.~Ilhan, S.~Tekin, G.~Liu, R.~Kompella, and L.~Liu, ``A survey on large language model-based game agents,'' \emph{arXiv preprint arXiv:2404.02039}, 2024.

\bibitem{xu2024survey}
X.~Xu, Y.~Wang, C.~Xu, Z.~Ding, J.~Jiang, Z.~Ding, and B.~F. Karlsson, ``A survey on game playing agents and large models: Methods, applications, and challenges,'' \emph{arXiv preprint arXiv:2403.10249}, 2024.

\bibitem{xu2024unleashing}
M.~Xu, H.~Du, D.~Niyato, J.~Kang, Z.~Xiong, S.~Mao, Z.~Han, A.~Jamalipour, D.~I. Kim, X.~Shen \emph{et~al.}, ``Unleashing the power of edge-cloud generative ai in mobile networks: A survey of aigc services,'' \emph{IEEE Communications Surveys \& Tutorials}, vol.~26, no.~2, pp. 1127--1170, 2024.

\bibitem{qu2025mobile}
G.~Qu, Q.~Chen, W.~Wei, Z.~Lin, X.~Chen, and K.~Huang, ``Mobile edge intelligence for large language models: A contemporary survey,'' \emph{IEEE Communications Surveys \& Tutorials}, 2025.

\bibitem{durante2024agent}
Z.~Durante, Q.~Huang, N.~Wake, R.~Gong, J.~S. Park, B.~Sarkar, R.~Taori, Y.~Noda, D.~Terzopoulos, Y.~Choi \emph{et~al.}, ``Agent ai: Surveying the horizons of multimodal interaction,'' \emph{arXiv preprint arXiv:2401.03568}, 2024.

\bibitem{wang2024large}
Y.~Wang, Y.~Pan, Q.~Zhao, Y.~Deng, Z.~Su, L.~Du, and T.~H. Luan, ``Large model agents: State-of-the-art, cooperation paradigms, security and privacy, and future trends,'' \emph{arXiv preprint arXiv:2409.14457}, 2024.

\bibitem{wang2024survey}
L.~Wang, C.~Ma, X.~Feng, Z.~Zhang, H.~Yang, J.~Zhang, Z.~Chen, J.~Tang, X.~Chen, Y.~Lin \emph{et~al.}, ``A survey on large language model based autonomous agents,'' \emph{Frontiers of Computer Science}, vol.~18, no.~6, p. 186345, 2024.

\bibitem{li2024survey}
X.~Li, S.~Wang, S.~Zeng, Y.~Wu, and Y.~Yang, ``A survey on llm-based multi-agent systems: workflow, infrastructure, and challenges,'' \emph{Vicinagearth}, vol.~1, no.~1, p.~9, 2024.

\bibitem{li2024review}
X.~Li, ``A review of prominent paradigms for llm-based agents: Tool use (including rag), planning, and feedback learning,'' \emph{arXiv preprint arXiv:2406.05804}, 2024.

\bibitem{jin2025comprehensive}
W.~Jin, H.~Du, B.~Zhao, X.~Tian, B.~Shi, and G.~Yang, ``A comprehensive survey on multi-agent cooperative decision-making: Scenarios, approaches, challenges and perspectives,'' \emph{arXiv preprint arXiv:2503.13415}, 2025.

\bibitem{ma2024survey}
Y.~Ma, Z.~Song, Y.~Zhuang, J.~Hao, and I.~King, ``A survey on vision-language-action models for embodied ai,'' \emph{arXiv preprint arXiv:2405.14093}, 2024.

\bibitem{guo2024large}
T.~Guo, X.~Chen, Y.~Wang, R.~Chang, S.~Pei, N.~V. Chawla, O.~Wiest, and X.~Zhang, ``Large language model based multi-agents: A survey of progress and challenges,'' \emph{arXiv preprint arXiv:2402.01680}, 2024.

\bibitem{masterman2024landscape}
T.~Masterman, S.~Besen, M.~Sawtell, and A.~Chao, ``The landscape of emerging ai agent architectures for reasoning, planning, and tool calling: A survey,'' \emph{arXiv preprint arXiv:2404.11584}, 2024.

\bibitem{cheng2024exploring}
Y.~Cheng, C.~Zhang, Z.~Zhang, X.~Meng, S.~Hong, W.~Li, Z.~Wang, Z.~Wang, F.~Yin, J.~Zhao \emph{et~al.}, ``Exploring large language model based intelligent agents: Definitions, methods, and prospects,'' \emph{arXiv preprint arXiv:2401.03428}, 2024.

\bibitem{li2023camel}
G.~Li, H.~A. A.~K. Hammoud, H.~Itani, D.~Khizbullin, and B.~Ghanem, ``Camel: Communicative agents for "mind" exploration of large language model society,'' in \emph{NeurIPS}, 2023.

\bibitem{wu2023autogen}
Q.~Wu, G.~Bansal, J.~Zhang, Y.~Wu, B.~Li, E.~Zhu, L.~Jiang, X.~Zhang, S.~Zhang, J.~Liu, A.~H. Awadallah, R.~W. White, D.~Burger, and C.~Wang, ``Autogen: Enabling next-gen llm applications via multi-agent conversation,'' 2023.

\bibitem{hong2024metagpt}
S.~Hong, X.~Zheng, J.~Chen, Y.~Cheng, J.~Wang, C.~Zhang, Z.~Wang, S.~K.~S. Yau, Z.~Lin, L.~Zhou \emph{et~al.}, ``Metagpt: Meta programming for a multi-agent collaborative framework,'' in \emph{ICLR}, 2024.

\bibitem{qian2024chatdev}
C.~Qian, W.~Liu, H.~Liu, N.~Chen, Y.~Dang, J.~Li, C.~Yang, W.~Chen, Y.~Su, X.~Cong \emph{et~al.}, ``Chatdev: Communicative agents for software development,'' in \emph{ACL}, 2024, pp. 15\,174--15\,186.

\bibitem{zhang2025aflow}
J.~Zhang, J.~Xiang, Z.~Yu, F.~Teng, X.-H. Chen, J.~Chen, M.~Zhuge, X.~Cheng, S.~Hong, J.~Wang, B.~Liu, Y.~Luo, and C.~Wu, ``{AF}low: Automating agentic workflow generation,'' in \emph{ICLR}, 2025.

\bibitem{park2023generative}
J.~S. Park, J.~O'Brien, C.~J. Cai, M.~R. Morris, P.~Liang, and M.~S. Bernstein, ``Generative agents: Interactive simulacra of human behavior,'' in \emph{UIST}, 2023, pp. 1--22.

\bibitem{wang2025user}
L.~Wang, J.~Zhang, H.~Yang, Z.-Y. Chen, J.~Tang, Z.~Zhang, X.~Chen, Y.~Lin, H.~Sun, R.~Song \emph{et~al.}, ``User behavior simulation with large language model-based agents,'' \emph{ACM Transactions on Information Systems}, vol.~43, no.~2, pp. 1--37, 2025.

\bibitem{khattab2024dspy}
O.~Khattab, A.~Singhvi, P.~Maheshwari, Z.~Zhang, K.~Santhanam, S.~Vardhamanan, S.~Haq, A.~Sharma, T.~T. Joshi, H.~Moazam, H.~Miller, M.~Zaharia, and C.~Potts, ``Dspy: Compiling declarative language model calls into self-improving pipelines,'' in \emph{ICLR}, 2024.

\bibitem{yao2023react}
S.~Yao, J.~Zhao, D.~Yu, N.~Du, I.~Shafran, K.~Narasimhan, and Y.~Cao, ``React: Synergizing reasoning and acting in language models,'' in \emph{ICLR}, 2023.

\bibitem{besta2024graphofthoughts}
M.~Besta, N.~Blach, A.~Kubicek, R.~Gerstenberger, M.~Podstawski, L.~Gianinazzi, J.~Gajda, T.~Lehmann, H.~Niewiadomski, P.~Nyczyk \emph{et~al.}, ``Graph of thoughts: Solving elaborate problems with large language models,'' in \emph{AAAI}, vol.~38, no.~16, 2024, pp. 17\,682--17\,690.

\bibitem{wang2023voyager}
G.~Wang, Y.~Xie, Y.~Jiang, A.~Mandlekar, C.~Xiao, Y.~Zhu, L.~Fan, and A.~Anandkumar, ``Voyager: An open-ended embodied agent with large language models,'' \emph{TMLR}, 2023.

\bibitem{zhu2023ghost}
X.~Zhu, Y.~Chen, H.~Tian, C.~Tao, W.~Su, C.~Yang, G.~Huang, B.~Li, L.~Lu, X.~Wang \emph{et~al.}, ``Ghost in the minecraft: Generally capable agents for open-world environments via large language models with text-based knowledge and memory,'' \emph{arXiv preprint arXiv:2305.17144}, 2023.

\bibitem{zhao2024expel}
A.~Zhao, D.~Huang, Q.~Xu, M.~Lin, Y.-J. Liu, and G.~Huang, ``Expel: Llm agents are experiential learners,'' in \emph{AAAI}, 2024, pp. 19\,632--19\,642.

\bibitem{shinn2023reflexion}
N.~Shinn, F.~Cassano, A.~Gopinath, K.~Narasimhan, and S.~Yao, ``Reflexion: Language agents with verbal reinforcement learning,'' \emph{NeurIPS}, vol.~36, pp. 8634--8652, 2023.

\bibitem{ruan2023tptu}
J.~Ruan, Y.~Chen, B.~Zhang, Z.~Xu, T.~Bao, H.~Mao, Z.~Li, X.~Zeng, R.~Zhao \emph{et~al.}, ``Tptu: Task planning and tool usage of large language model-based ai agents,'' in \emph{NeurIPS}, 2023.

\bibitem{xie2023openagents}
T.~Xie, F.~Zhou, Z.~Cheng, P.~Shi, L.~Weng, Y.~Liu, T.~J. Hua, J.~Zhao, Q.~Liu, C.~Liu \emph{et~al.}, ``Openagents: An open platform for language agents in the wild,'' \emph{arXiv preprint arXiv:2310.10634}, 2023.

\bibitem{wang2024lego}
H.~Wang, H.~Xin, C.~Zheng, Z.~Liu, Q.~Cao, Y.~Huang, J.~Xiong, H.~Shi, E.~Xie, J.~Yin \emph{et~al.}, ``Lego-prover: Neural theorem proving with growing libraries,'' in \emph{ICLR}, 2024.

\bibitem{packer2023memgpt}
C.~Packer, V.~Fang, S.~G. Patil, K.~Lin, S.~Wooders, and J.~E. Gonzalez, ``Memgpt: Towards llms as operating systems,'' \emph{CoRR}, 2023.

\bibitem{lewis2020retrieval}
P.~Lewis, E.~Perez, A.~Piktus, F.~Petroni, V.~Karpukhin, N.~Goyal, H.~K{\"u}ttler, M.~Lewis, W.-t. Yih, T.~Rockt{\"a}schel \emph{et~al.}, ``Retrieval-augmented generation for knowledge-intensive nlp tasks,'' \emph{NeurIPS}, vol.~33, pp. 9459--9474, 2020.

\bibitem{edge2024graphrag}
D.~Edge, H.~Trinh, N.~Cheng, J.~Bradley, A.~Chao, A.~Mody, S.~Truitt, D.~Metropolitansky, R.~O. Ness, and J.~Larson, ``From local to global: A graph rag approach to query-focused summarization,'' \emph{arXiv preprint arXiv:2404.16130}, 2024.

\bibitem{zhang2024chainofagents}
Y.~Zhang, R.~Sun, Y.~Chen, T.~Pfister, R.~Zhang, and S.~Arik, ``Chain of agents: Large language models collaborating on long-context tasks,'' \emph{Advances in Neural Information Processing Systems}, vol.~37, pp. 132\,208--132\,237, 2024.

\bibitem{trivedi2022interleaving}
H.~Trivedi, N.~Balasubramanian, T.~Khot, and A.~Sabharwal, ``Interleaving retrieval with chain-of-thought reasoning for knowledge-intensive multi-step questions,'' \emph{arXiv preprint arXiv:2212.10509}, 2022.

\bibitem{li2024llatrieval}
X.~Li, C.~Zhu, L.~Li, Z.~Yin, T.~Sun, and X.~Qiu, ``Llatrieval: Llm-verified retrieval for verifiable generation,'' in \emph{NAACL}, 2024, pp. 5453--5471.

\bibitem{wu2025graphaugreasoning}
W.~Wu, Y.~Jing, Y.~Wang, W.~Hu, and D.~Tao, ``Graph-augmented reasoning: Evolving step-by-step knowledge graph retrieval for llm reasoning,'' 2025.

\bibitem{guan2025deeprag}
X.~Guan, J.~Zeng, F.~Meng, C.~Xin, Y.~Lu, H.~Lin, X.~Han, L.~Sun, and J.~Zhou, ``Deeprag: Thinking to retrieval step by step for large language models,'' \emph{arXiv preprint arXiv:2502.01142}, 2025.

\bibitem{wang2023plan}
L.~Wang, W.~Xu, Y.~Lan, Z.~Hu, Y.~Lan, R.~K.-W. Lee, and E.-P. Lim, ``Plan-and-solve prompting: Improving zero-shot chain-of-thought reasoning by large language models,'' \emph{arXiv preprint arXiv:2305.04091}, 2023.

\bibitem{durfee2001distributed}
E.~H. Durfee, ``Distributed problem solving and planning,'' in \emph{ECCAI Advanced Course on Artificial Intelligence}.\hskip 1em plus 0.5em minus 0.4em\relax Springer, 2001, pp. 118--149.

\bibitem{tao2024chain}
M.~Tao, D.~Zhao, and Y.~Feng, ``Chain-of-discussion: A multi-model framework for complex evidence-based question answering,'' \emph{arXiv preprint arXiv:2402.16313}, 2024.

\bibitem{hu2023tree}
M.~Hu, Y.~Mu, X.~Yu, M.~Ding, S.~Wu, W.~Shao, Q.~Chen, B.~Wang, Y.~Qiao, and P.~Luo, ``Tree-planner: Efficient close-loop task planning with large language models,'' \emph{arXiv preprint arXiv:2310.08582}, 2023.

\bibitem{choireactree}
J.-W. Choi, H.~Kim, H.~Ong, Y.~Yoon, M.~Jang, J.~Kim \emph{et~al.}, ``Reactree: Hierarchical task planning with dynamic tree expansion using llm agent nodes,'' 2025.

\bibitem{long2023large}
J.~Long, ``Large language model guided tree-of-thought,'' \emph{arXiv preprint arXiv:2305.08291}, 2023.

\bibitem{zhang2024rest}
D.~Zhang, S.~Zhoubian, Z.~Hu, Y.~Yue, Y.~Dong, and J.~Tang, ``Rest-mcts*: Llm self-training via process reward guided tree search,'' \emph{NeurIPS}, vol.~37, pp. 64\,735--64\,772, 2024.

\bibitem{lykov2023llm}
A.~Lykov, M.~Dronova, N.~Naglov, M.~Litvinov, S.~Satsevich, A.~Bazhenov, V.~Berman, A.~Shcherbak, and D.~Tsetserukou, ``Llm-mars: Large language model for behavior tree generation and nlp-enhanced dialogue in multi-agent robot systems,'' \emph{arXiv preprint arXiv:2312.09348}, 2023.

\bibitem{ao2024llm}
J.~Ao, F.~Wu, Y.~Wu, A.~Swikir, and S.~Haddadin, ``Llm as bt-planner: Leveraging llms for behavior tree generation in robot task planning,'' \emph{arXiv preprint arXiv:2409.10444}, 2024.

\bibitem{rivera2024conceptagent}
C.~Rivera, G.~Byrd, W.~Paul, T.~Feldman, M.~Booker, E.~Holmes, D.~Handelman, B.~Kemp, A.~Badger, A.~Schmidt \emph{et~al.}, ``Conceptagent: Llm-driven precondition grounding and tree search for robust task planning and execution,'' \emph{arXiv preprint arXiv:2410.06108}, 2024.

\bibitem{bhat2024grounding}
V.~Bhat, A.~U. Kaypak, P.~Krishnamurthy, R.~Karri, and F.~Khorrami, ``Grounding llms for robot task planning using closed-loop state feedback,'' \emph{arXiv preprint arXiv:2402.08546}, 2024.

\bibitem{li2023traineragent}
H.~Li, H.~Jiang, T.~Zhang, Z.~Yu, A.~Yin, H.~Cheng, S.~Fu, Y.~Zhang, and W.~He, ``Traineragent: Customizable and efficient model training through llm-powered multi-agent system,'' \emph{arXiv preprint arXiv:2311.06622}, 2023.

\bibitem{wan2024dynamic}
G.~Wan, Y.~Wu, J.~Chen, and S.~Li, ``Dynamic self-consistency: Leveraging reasoning paths for efficient llm sampling,'' \emph{arXiv preprint arXiv:2408.17017}, 2024.

\bibitem{seo2024llm}
S.~Seo, J.~Lee, S.~Noh, and H.~Kang, ``Llm-based cooperative agents using information relevance and plan validation,'' \emph{arXiv preprint arXiv:2405.16751}, 2024.

\bibitem{sun2023adaplanner}
H.~Sun, Y.~Zhuang, L.~Kong, B.~Dai, and C.~Zhang, ``Adaplanner: Adaptive planning from feedback with language models,'' \emph{NeurIPS}, vol.~36, pp. 58\,202--58\,245, 2023.

\bibitem{jafaripour2025adaptive}
M.~Jafaripour, S.~Golestan, S.~Miwa, Y.~Mitsuka, and O.~Zaiane, ``Adaptive iterative feedback prompting for obstacle-aware path planning via llms,'' in \emph{AAAI Workshop}, 2025.

\bibitem{qiao2023making}
S.~Qiao, H.~Gui, C.~Lv, Q.~Jia, H.~Chen, and N.~Zhang, ``Making language models better tool learners with execution feedback,'' \emph{arXiv preprint arXiv:2305.13068}, 2023.

\bibitem{yang2023gpt4tools}
R.~Yang, L.~Song, Y.~Li, S.~Zhao, Y.~Ge, X.~Li, and Y.~Shan, ``Gpt4tools: Teaching large language model to use tools via self-instruction,'' \emph{NeurIPS}, vol.~36, pp. 71\,995--72\,007, 2023.

\bibitem{yuan2024easytool}
S.~Yuan, K.~Song, J.~Chen, X.~Tan, Y.~Shen, R.~Kan, D.~Li, and D.~Yang, ``Easytool: Enhancing llm-based agents with concise tool instruction,'' \emph{arXiv preprint arXiv:2401.06201}, 2024.

\bibitem{wu2025avatar}
S.~Wu, S.~Zhao, Q.~Huang, K.~Huang, M.~Yasunaga, K.~Cao, V.~Ioannidis, K.~Subbian, J.~Leskovec, and J.~Y. Zou, ``Avatar: Optimizing llm agents for tool usage via contrastive reasoning,'' \emph{NeurIPS}, vol.~37, pp. 25\,981--26\,010, 2025.

\bibitem{huang2024drivlme}
Y.~Huang, J.~Sansom, Z.~Ma, F.~Gervits, and J.~Chai, ``Drivlme: Enhancing llm-based autonomous driving agents with embodied and social experiences,'' in \emph{IROS}.\hskip 1em plus 0.5em minus 0.4em\relax IEEE, 2024, pp. 3153--3160.

\bibitem{zhang2024towards}
Y.~Zhang, S.~Yang, C.~Bai, F.~Wu, X.~Li, Z.~Wang, and X.~Li, ``Towards efficient llm grounding for embodied multi-agent collaboration,'' \emph{arXiv preprint arXiv:2405.14314}, 2024.

\bibitem{colle2024improving}
B.~Colle, ``Improving embodied llm agents capabilities through collaboration,'' 2024.

\bibitem{boiko2023autonomous}
D.~A. Boiko, R.~MacKnight, B.~Kline, and G.~Gomes, ``Autonomous chemical research with large language models,'' \emph{Nature}, vol. 624, no. 7992, pp. 570--578, 2023.

\bibitem{jiang2023llmlingua}
H.~Jiang, Q.~Wu, C.-Y. Lin, Y.~Yang, and L.~Qiu, ``Llmlingua: Compressing prompts for accelerated inference of large language models,'' in \emph{EMNLP}, 2023, pp. 13\,358--13\,376.

\bibitem{qiao2024autoact}
S.~Qiao, N.~Zhang, R.~Fang, Y.~Luo, W.~Zhou, Y.~E. Jiang, C.~Lv, and H.~Chen, ``Autoact: Automatic agent learning from scratch for qa via self-planning,'' \emph{arXiv preprint arXiv:2401.05268}, 2024.

\bibitem{suzgun2024meta}
M.~Suzgun and A.~T. Kalai, ``Meta-prompting: Enhancing language models with task-agnostic scaffolding,'' \emph{arXiv preprint arXiv:2401.12954}, 2024.

\bibitem{khan2024debating}
A.~Khan, J.~Hughes, D.~Valentine, L.~Ruis, K.~Sachan, A.~Radhakrishnan, E.~Grefenstette, S.~R. Bowman, T.~Rockt{\"a}schel, and E.~Perez, ``Debating with more persuasive llms leads to more truthful answers,'' \emph{arXiv preprint arXiv:2402.06782}, 2024.

\bibitem{tang2023medagents}
X.~Tang, A.~Zou, Z.~Zhang, Z.~Li, Y.~Zhao, X.~Zhang, A.~Cohan, and M.~Gerstein, ``Medagents: Large language models as collaborators for zero-shot medical reasoning,'' \emph{arXiv preprint arXiv:2311.10537}, 2023.

\bibitem{chen2023reconcile}
J.~C.-Y. Chen, S.~Saha, and M.~Bansal, ``Reconcile: Round-table conference improves reasoning via consensus among diverse llms,'' \emph{arXiv preprint arXiv:2309.13007}, 2023.

\bibitem{liang2023encouraging}
T.~Liang, Z.~He, W.~Jiao, X.~Wang, Y.~Wang, R.~Wang, Y.~Yang, S.~Shi, and Z.~Tu, ``Encouraging divergent thinking in large language models through multi-agent debate,'' \emph{arXiv preprint arXiv:2305.19118}, 2023.

\bibitem{kim2024can}
K.~Kim, S.~Lee, K.-H. Huang, H.~P. Chan, M.~Li, and H.~Ji, ``Can llms produce faithful explanations for fact-checking? towards faithful explainable fact-checking via multi-agent debate,'' \emph{arXiv preprint arXiv:2402.07401}, 2024.

\bibitem{du2023improving}
Y.~Du, S.~Li, A.~Torralba, J.~B. Tenenbaum, and I.~Mordatch, ``Improving factuality and reasoning in language models through multiagent debate,'' in \emph{ICML}, 2023.

\bibitem{zhu2024knowagent}
Y.~Zhu, S.~Qiao, Y.~Ou, S.~Deng, N.~Zhang, S.~Lyu, Y.~Shen, L.~Liang, J.~Gu, and H.~Chen, ``Knowagent: Knowledge-augmented planning for llm-based agents,'' \emph{arXiv preprint arXiv:2403.03101}, 2024.

\bibitem{qiao2024agent}
S.~Qiao, R.~Fang, N.~Zhang, Y.~Zhu, X.~Chen, S.~Deng, Y.~Jiang, P.~Xie, F.~Huang, and H.~Chen, ``Agent planning with world knowledge model,'' \emph{NeurIPS}, vol.~37, pp. 114\,843--114\,871, 2024.

\bibitem{fang2024refining}
R.~Fang, S.~Qiao, and Z.~Xi, ``Refining guideline knowledge for agent planning using textgrad,'' in \emph{ICKG}.\hskip 1em plus 0.5em minus 0.4em\relax IEEE, 2024, pp. 102--103.

\bibitem{zhong2023self}
Q.~Zhong, L.~Ding, J.~Liu, B.~Du, and D.~Tao, ``Self-evolution learning for discriminative language model pretraining,'' in \emph{ACL Findings}, 2023, pp. 4130--4145.

\bibitem{akiba2025evolutionary}
T.~Akiba, M.~Shing, Y.~Tang, Q.~Sun, and D.~Ha, ``Evolutionary optimization of model merging recipes,'' \emph{Nature Machine Intelligence}, pp. 1--10, 2025.

\bibitem{wu2023self}
S.~Wu, K.~Lu, B.~Xu, J.~Lin, Q.~Su, and C.~Zhou, ``Self-evolved diverse data sampling for efficient instruction tuning,'' \emph{arXiv preprint arXiv:2311.08182}, 2023.

\bibitem{madaan2023self}
A.~Madaan, N.~Tandon, P.~Gupta, S.~Hallinan, L.~Gao, S.~Wiegreffe, U.~Alon, N.~Dziri, S.~Prabhumoye, Y.~Yang \emph{et~al.}, ``Self-refine: Iterative refinement with self-feedback,'' \emph{NeurIPS}, vol.~36, pp. 46\,534--46\,594, 2023.

\bibitem{zelikman2024star}
E.~Zelikman, Y.~Wu, J.~Mu, and N.~D. Goodman, ``Star: Self-taught reasoner bootstrapping reasoning with reasoning,'' in \emph{NeurIPS}, vol. 1126, 2024.

\bibitem{hosseiniv}
A.~Hosseini, X.~Yuan, N.~Malkin, A.~Courville, A.~Sordoni, and R.~Agarwal, ``V-star: Training verifiers for self-taught reasoners,'' in \emph{COLM}, 2024.

\bibitem{weng2023large}
Y.~Weng, M.~Zhu, F.~Xia, B.~Li, S.~He, S.~Liu, B.~Sun, K.~Liu, and J.~Zhao, ``Large language models are better reasoners with self-verification,'' in \emph{EMNLP Findings}, 2023, pp. 2550--2575.

\bibitem{yuan2024selfrewardinglanguagemodels}
W.~Yuan, R.~Y. Pang, K.~Cho, X.~Li, S.~Sukhbaatar, J.~Xu, and J.~Weston, ``Self-rewarding language models,'' 2024.

\bibitem{yang2024rlcd}
K.~Yang, D.~Klein, A.~Celikyilmaz, N.~Peng, and Y.~Tian, ``Rlcd: Reinforcement learning from contrastive distillation for lm alignment,'' in \emph{ICLR}, 2024.

\bibitem{panglanguage}
J.-C. Pang, P.~Wang, K.~Li, X.-H. Chen, J.~Xu, Z.~Zhang, and Y.~Yu, ``Language model self-improvement by reinforcement learning contemplation,'' in \emph{ICLR}, 2024.

\bibitem{zhang2024proagent}
C.~Zhang, K.~Yang, S.~Hu, Z.~Wang, G.~Li, Y.~Sun, C.~Zhang, Z.~Zhang, A.~Liu, S.-C. Zhu \emph{et~al.}, ``Proagent: building proactive cooperative agents with large language models,'' in \emph{AAAI}, vol.~38, no.~16, 2024, pp. 17\,591--17\,599.

\bibitem{ma2024coevolving}
H.~Ma, T.~Hu, Z.~Pu, L.~Boyin, X.~Ai, Y.~Liang, and M.~Chen, ``Coevolving with the other you: Fine-tuning llm with sequential cooperative multi-agent reinforcement learning,'' \emph{NeurIPS}, vol.~37, pp. 15\,497--15\,525, 2024.

\bibitem{ma2023evolving}
C.~Ma, Z.~Yang, H.~Ci, J.~Gao, M.~Gao, X.~Pan, and Y.~Yang, ``Evolving diverse red-team language models in multi-round multi-agent games,'' \emph{arXiv preprint arXiv:2310.00322}, 2023.

\bibitem{liang2024encouraging}
T.~Liang, Z.~He, W.~Jiao, X.~Wang, Y.~Wang, R.~Wang, Y.~Yang, S.~Shi, and Z.~Tu, ``Encouraging divergent thinking in large language models through multi-agent debate,'' in \emph{EMNLP}, 2024, pp. 17\,889--17\,904.

\bibitem{goucritic}
Z.~Gou, Z.~Shao, Y.~Gong, Y.~Yang, N.~Duan, W.~Chen \emph{et~al.}, ``Critic: Large language models can self-correct with tool-interactive critiquing,'' in \emph{ICLR}, 2024.

\bibitem{song2024trial}
Y.~Song, D.~Yin, X.~Yue, J.~Huang, S.~Li, and B.~Y. Lin, ``Trial and error: Exploration-based trajectory optimization of llm agents,'' in \emph{ACL}, 2024, pp. 7584--7600.

\bibitem{jiang2023selfevolve}
S.~Jiang, Y.~Wang, and Y.~Wang, ``Selfevolve: A code evolution framework via large language models,'' \emph{arXiv preprint arXiv:2306.02907}, 2023.

\bibitem{huang2024understanding}
X.~Huang, W.~Liu, X.~Chen, X.~Wang, H.~Wang, D.~Lian, Y.~Wang, R.~Tang, and E.~Chen, ``Understanding the planning of llm agents: A survey,'' \emph{arXiv preprint arXiv:2402.02716}, 2024.

\bibitem{kojima2022large}
T.~Kojima, S.~S. Gu, M.~Reid, Y.~Matsuo, and Y.~Iwasawa, ``Large language models are zero-shot reasoners,'' \emph{NeurIPS}, vol.~35, pp. 22\,199--22\,213, 2022.

\bibitem{wei2022chain}
J.~Wei, X.~Wang, D.~Schuurmans, M.~Bosma, F.~Xia, E.~Chi, Q.~V. Le, D.~Zhou \emph{et~al.}, ``Chain-of-thought prompting elicits reasoning in large language models,'' \emph{NeurIPS}, vol.~35, pp. 24\,824--24\,837, 2022.

\bibitem{wang2022self}
X.~Wang, J.~Wei, D.~Schuurmans, Q.~Le, E.~Chi, S.~Narang, A.~Chowdhery, and D.~Zhou, ``Self-consistency improves chain of thought reasoning in language models,'' \emph{arXiv preprint arXiv:2203.11171}, 2022.

\bibitem{li2024enhancing}
W.~Li and W.~Pan, ``Enhancing chain-of-thought reasoning in large language models through text style diversity and prompt fusion,'' in \emph{EIBDCT}, vol. 13181.\hskip 1em plus 0.5em minus 0.4em\relax SPIE, 2024, pp. 226--232.

\bibitem{jiang2024technical}
J.~Jiang, Z.~Chen, Y.~Min, J.~Chen, X.~Cheng, J.~Wang, Y.~Tang, H.~Sun, J.~Deng, W.~X. Zhao \emph{et~al.}, ``Technical report: Enhancing llm reasoning with reward-guided tree search,'' \emph{arXiv preprint arXiv:2411.11694}, 2024.

\bibitem{browne2012survey}
C.~B. Browne, E.~Powley, D.~Whitehouse, S.~M. Lucas, P.~I. Cowling, P.~Rohlfshagen, S.~Tavener, D.~Perez, S.~Samothrakis, and S.~Colton, ``A survey of monte carlo tree search methods,'' \emph{IEEE Transactions on Computational Intelligence and AI in games}, vol.~4, no.~1, pp. 1--43, 2012.

\bibitem{guo2024can}
H.~Guo, Z.~Liu, Y.~Zhang, and Z.~Wang, ``Can large language models play games? a case study of a self-play approach,'' \emph{arXiv preprint arXiv:2403.05632}, 2024.

\bibitem{liu2024large}
Y.~Liu, P.~Sun, and H.~Li, ``Large language models as agents in two-player games,'' \emph{arXiv preprint arXiv:2402.08078}, 2024.

\bibitem{laleh2024survey}
A.~R. Laleh and M.~N. Ahmadabadi, ``A survey on enhancing reinforcement learning in complex environments: Insights from human and llm feedback,'' \emph{arXiv preprint arXiv:2411.13410}, 2024.

\bibitem{shen2024llm}
Z.~Shen, ``Llm with tools: A survey,'' \emph{arXiv preprint arXiv:2409.18807}, 2024.

\bibitem{kim2024understanding}
C.~Y. Kim, C.~P. Lee, and B.~Mutlu, ``Understanding large-language model (llm)-powered human-robot interaction,'' in \emph{HRI}, 2024, pp. 371--380.

\bibitem{li2025metal}
B.~Li, Y.~Wang, J.~Gu, K.-W. Chang, and N.~Peng, ``Metal: A multi-agent framework for chart generation with test-time scaling,'' \emph{arXiv preprint arXiv:2502.17651}, 2025.

\bibitem{guo2024ds}
S.~Guo, C.~Deng, Y.~Wen, H.~Chen, Y.~Chang, and J.~Wang, ``Ds-agent: Automated data science by empowering large language models with case-based reasoning,'' \emph{arXiv preprint arXiv:2402.17453}, 2024.

\bibitem{yin2023exchange}
Z.~Yin, Q.~Sun, C.~Chang, Q.~Guo, J.~Dai, X.~Huang, and X.~Qiu, ``Exchange-of-thought: Enhancing large language model capabilities through cross-model communication,'' \emph{arXiv preprint arXiv:2312.01823}, 2023.

\bibitem{li2021learning}
Y.~Li, S.~Ren, P.~Wu, S.~Chen, C.~Feng, and W.~Zhang, ``Learning distilled collaboration graph for multi-agent perception,'' \emph{NeurIPS}, vol.~34, pp. 29\,541--29\,552, 2021.

\bibitem{liu2024dynamic}
Z.~Liu, Y.~Zhang, P.~Li, Y.~Liu, and D.~Yang, ``A dynamic llm-powered agent network for task-oriented agent collaboration,'' in \emph{COLM}, 2024.

\bibitem{kim2024mdagents}
Y.~Kim, C.~Park, H.~Jeong, Y.~S. Chan, X.~Xu, D.~McDuff, H.~Lee, M.~Ghassemi, C.~Breazeal, H.~Park \emph{et~al.}, ``Mdagents: An adaptive collaboration of llms for medical decision-making,'' \emph{NeurIPS}, vol.~37, pp. 79\,410--79\,452, 2024.

\bibitem{ying2023inferring}
L.~Ying, T.~Zhi-Xuan, V.~Mansinghka, and J.~B. Tenenbaum, ``Inferring the goals of communicating agents from actions and instructions,'' in \emph{Proceedings of the AAAI Symposium Series}, vol.~2, no.~1, 2023, pp. 26--33.

\bibitem{vyas2024autonomous}
J.~Vyas and M.~Mercang{\"o}z, ``Autonomous industrial control using an agentic framework with large language models,'' \emph{arXiv preprint arXiv:2411.05904}, 2024.

\bibitem{dell2022data}
D.~Dell'Anna, N.~Alechina, F.~Dalpiaz, M.~Dastani, and B.~Logan, ``Data-driven revision of conditional norms in multi-agent systems,'' \emph{Journal of Artificial Intelligence Research}, vol.~75, pp. 1549--1593, 2022.

\bibitem{liu2023agentbench}
X.~Liu, H.~Yu, H.~Zhang, Y.~Xu, X.~Lei, H.~Lai, Y.~Gu, H.~Ding, K.~Men, K.~Yang \emph{et~al.}, ``Agentbench: Evaluating llms as agents,'' \emph{arXiv preprint arXiv:2308.03688}, 2023.

\bibitem{deng2023mind2web}
X.~Deng, Y.~Gu, B.~Zheng, S.~Chen, S.~Stevens, B.~Wang, H.~Sun, and Y.~Su, ``Mind2web: Towards a generalist agent for the web,'' \emph{NeurIPS}, vol.~36, pp. 28\,091--28\,114, 2023.

\bibitem{yin2024mmau}
G.~Yin, H.~Bai, S.~Ma, F.~Nan, Y.~Sun, Z.~Xu, S.~Ma, J.~Lu, X.~Kong, A.~Zhang \emph{et~al.}, ``Mmau: A holistic benchmark of agent capabilities across diverse domains,'' \emph{arXiv preprint arXiv:2407.18961}, 2024.

\bibitem{gu2024blade}
K.~Gu, R.~Shang, R.~Jiang, K.~Kuang, R.-J. Lin, D.~Lyu, Y.~Mao, Y.~Pan, T.~Wu, J.~Yu \emph{et~al.}, ``Blade: Benchmarking language model agents for data-driven science,'' \emph{arXiv preprint arXiv:2408.09667}, 2024.

\bibitem{liu2024visualagentbench}
X.~Liu, T.~Zhang, Y.~Gu, I.~L. Iong, Y.~Xu, X.~Song, S.~Zhang, H.~Lai, X.~Liu, H.~Zhao \emph{et~al.}, ``Visualagentbench: Towards large multimodal models as visual foundation agents,'' \emph{arXiv preprint arXiv:2408.06327}, 2024.

\bibitem{li2025embodied}
M.~Li, S.~Zhao, Q.~Wang, K.~Wang, Y.~Zhou, S.~Srivastava, C.~Gokmen, T.~Lee, E.~L. Li, R.~Zhang \emph{et~al.}, ``Embodied agent interface: Benchmarking llms for embodied decision making,'' \emph{NeurIPS}, vol.~37, pp. 100\,428--100\,534, 2025.

\bibitem{xu2024crab}
T.~Xu, L.~Chen, D.-J. Wu, Y.~Chen, Z.~Zhang, X.~Yao, Z.~Xie, Y.~Chen, S.~Liu, B.~Qian \emph{et~al.}, ``Crab: Cross-platfrom agent benchmark for multi-modal embodied language model agents,'' in \emph{NeurIPS Workshop}, 2024.

\bibitem{butt2024benchagents}
N.~Butt, V.~Chandrasekaran, N.~Joshi, B.~Nushi, and V.~Balachandran, ``Benchagents: Automated benchmark creation with agent interaction,'' \emph{arXiv preprint arXiv:2410.22584}, 2024.

\bibitem{wang2024benchmark}
S.~Wang, Z.~Long, Z.~Fan, Z.~Wei, and X.~Huang, ``Benchmark self-evolving: A multi-agent framework for dynamic llm evaluation,'' \emph{arXiv preprint arXiv:2402.11443}, 2024.

\bibitem{wang2024revisiting}
W.~Wang, Z.~Ma, P.~Liu, and M.~Chen, ``Revisiting benchmark and assessment: An agent-based exploratory dynamic evaluation framework for llms,'' \emph{arXiv preprint arXiv:2410.11507}, 2024.

\bibitem{wu2024seal}
M.~Wu, T.~Zhu, H.~Han, C.~Tan, X.~Zhang, and W.~Chen, ``Seal-tools: Self-instruct tool learning dataset for agent tuning and detailed benchmark,'' in \emph{NLPCC}.\hskip 1em plus 0.5em minus 0.4em\relax Springer, 2024, pp. 372--384.

\bibitem{guo2024ctooleval}
Z.~Guo, Y.~Huang, and D.~Xiong, ``Ctooleval: a chinese benchmark for llm-powered agent evaluation in real-world api interactions,'' in \emph{ACL Findings}, 2024, pp. 15\,711--15\,724.

\bibitem{jiang2025medagentbench}
Y.~Jiang, K.~C. Black, G.~Geng, D.~Park, A.~Y. Ng, and J.~H. Chen, ``Medagentbench: Dataset for benchmarking llms as agents in medical applications,'' \emph{arXiv preprint arXiv:2501.14654}, 2025.

\bibitem{fan2024ai}
Z.~Fan, J.~Tang, W.~Chen, S.~Wang, Z.~Wei, J.~Xi, F.~Huang, and J.~Zhou, ``Ai hospital: Benchmarking large language models in a multi-agent medical interaction simulator,'' \emph{arXiv preprint arXiv:2402.09742}, 2024.

\bibitem{ma2024lampilot}
Y.~Ma, C.~Cui, X.~Cao, W.~Ye, P.~Liu, J.~Lu, A.~Abdelraouf, R.~Gupta, K.~Han, A.~Bera \emph{et~al.}, ``Lampilot: An open benchmark dataset for autonomous driving with language model programs,'' in \emph{CVPR}, 2024, pp. 15\,141--15\,151.

\bibitem{zhang2024benchmarking}
Y.~Zhang, Q.~Jiang, X.~Han, N.~Chen, Y.~Yang, and K.~Ren, ``Benchmarking data science agents,'' \emph{arXiv preprint arXiv:2402.17168}, 2024.

\bibitem{huang2024code}
Y.~Huang, J.~Luo, Y.~Yu, Y.~Zhang, F.~Lei, Y.~Wei, S.~He, L.~Huang, X.~Liu, J.~Zhao \emph{et~al.}, ``Da-code: Agent data science code generation benchmark for large language models,'' \emph{arXiv preprint arXiv:2410.07331}, 2024.

\bibitem{huang2024dca}
B.~Huang, Y.~Yu, J.~Huang, X.~Zhang, and J.~Ma, ``Dca-bench: A benchmark for dataset curation agents,'' \emph{arXiv preprint arXiv:2406.07275}, 2024.

\bibitem{xie2024travelplanner}
J.~Xie, K.~Zhang, J.~Chen, T.~Zhu, R.~Lou, Y.~Tian, Y.~Xiao, and Y.~Su, ``Travelplanner: A benchmark for real-world planning with language agents,'' \emph{arXiv preprint arXiv:2402.01622}, 2024.

\bibitem{huang2023benchmarking}
Q.~Huang, J.~Vora, P.~Liang, and J.~Leskovec, ``Benchmarking large language models as ai research agents,'' in \emph{NeurIPS Workshop}, 2023.

\bibitem{chan2024mle}
J.~S. Chan, N.~Chowdhury, O.~Jaffe, J.~Aung, D.~Sherburn, E.~Mays, G.~Starace, K.~Liu, L.~Maksin, T.~Patwardhan \emph{et~al.}, ``Mle-bench: Evaluating machine learning agents on machine learning engineering,'' \emph{arXiv preprint arXiv:2410.07095}, 2024.

\bibitem{andriushchenkoagentharm}
M.~Andriushchenko, A.~Souly, M.~Dziemian, D.~Duenas, M.~Lin, J.~Wang, D.~Hendrycks, A.~Zou, J.~Z. Kolter, M.~Fredrikson \emph{et~al.}, ``Agentharm: Benchmarking robustness of llm agents on harmful tasks,'' in \emph{ICLR}, 2024.

\bibitem{xie2025osworld}
T.~Xie, D.~Zhang, J.~Chen, X.~Li, S.~Zhao, R.~Cao, J.~H. Toh, Z.~Cheng, D.~Shin, F.~Lei \emph{et~al.}, ``Osworld: Benchmarking multimodal agents for open-ended tasks in real computer environments,'' \emph{NeurIPS}, vol.~37, pp. 52\,040--52\,094, 2025.

\bibitem{xu2024tur}
K.~Xu, Y.~Kordi, T.~Nayak, A.~Asija, Y.~Wang, K.~Sanders, A.~Byerly, J.~Zhang, B.~Van~Durme, and D.~Khashabi, ``Tur [k] ingbench: A challenge benchmark for web agents,'' \emph{arXiv preprint arXiv:2403.11905}, 2024.

\bibitem{kapoor2024omniact}
R.~Kapoor, Y.~P. Butala, M.~Russak, J.~Y. Koh, K.~Kamble, W.~AlShikh, and R.~Salakhutdinov, ``Omniact: A dataset and benchmark for enabling multimodal generalist autonomous agents for desktop and web,'' in \emph{ECCV}.\hskip 1em plus 0.5em minus 0.4em\relax Springer, 2024, pp. 161--178.

\bibitem{yang2025egolife}
J.~Yang, S.~Liu, H.~Guo, Y.~Dong, X.~Zhang, S.~Zhang, P.~Wang, Z.~Zhou, B.~Xie, Z.~Wang \emph{et~al.}, ``Egolife: Towards egocentric life assistant,'' \emph{arXiv preprint arXiv:2503.03803}, 2025.

\bibitem{wang2024gta}
J.~Wang, M.~Zerun, Y.~Li, S.~Zhang, C.~Chen, K.~Chen, and X.~Le, ``Gta: a benchmark for general tool agents,'' in \emph{NeurIPS}, 2024.

\bibitem{xu2024theagentcompany}
F.~F. Xu, Y.~Song, B.~Li, Y.~Tang, K.~Jain, M.~Bao, Z.~Z. Wang, X.~Zhou, Z.~Guo, M.~Cao \emph{et~al.}, ``Theagentcompany: benchmarking llm agents on consequential real world tasks,'' \emph{arXiv preprint arXiv:2412.14161}, 2024.

\bibitem{barbarroxa2024benchmarking}
R.~Barbarroxa, L.~Gomes, and Z.~Vale, ``Benchmarking large language models for multi-agent systems: A comparative analysis of autogen, crewai, and taskweaver,'' in \emph{International Conference on Practical Applications of Agents and Multi-Agent Systems}.\hskip 1em plus 0.5em minus 0.4em\relax Springer, 2024, pp. 39--48.

\bibitem{li2025benchmark}
Z.~Li, X.~Wu, H.~Du, H.~Nghiem, and G.~Shi, ``Benchmark evaluations, applications, and challenges of large vision language models: A survey,'' \emph{arXiv preprint arXiv:2501.02189}, 2025.

\bibitem{kenney2024ml}
M.~Kenney, ``Ml research benchmark,'' \emph{arXiv preprint arXiv:2410.22553}, 2024.

\bibitem{nakano2022webgpt}
R.~Nakano, J.~Hilton, S.~Balaji, J.~Wu, L.~Ouyang, C.~Kim, C.~Hesse, S.~Jain, V.~Kosaraju, W.~Saunders, X.~Jiang, K.~Cobbe, T.~Eloundou, G.~Krueger, K.~Button, M.~Knight, B.~Chess, and J.~Schulman, ``Webgpt: Browser-assisted question-answering with human feedback,'' 2022.

\bibitem{qin2023webcpm}
Y.~Qin, Z.~Cai, D.~Jin, L.~Yan, S.~Liang, K.~Zhu, Y.~Lin, X.~Han, N.~Ding, H.~Wang, R.~Xie, F.~Qi, Z.~Liu, M.~Sun, and J.~Zhou, ``{W}eb{CPM}: Interactive web search for {C}hinese long-form question answering,'' in \emph{ACL}, A.~Rogers, J.~Boyd-Graber, and N.~Okazaki, Eds.\hskip 1em plus 0.5em minus 0.4em\relax Toronto, Canada: Association for Computational Linguistics, Jul. 2023, pp. 8968--8988.

\bibitem{zhang2023toolcode}
K.~Zhang, H.~Zhang, G.~Li, J.~Li, Z.~Li, and Z.~Jin, ``Toolcoder: Teach code generation models to use api search tools,'' 2023.

\bibitem{robertson2009probabilistic}
S.~Robertson, H.~Zaragoza \emph{et~al.}, ``The probabilistic relevance framework: Bm25 and beyond,'' \emph{Foundations and Trends{\textregistered} in Information Retrieval}, vol.~3, no.~4, pp. 333--389, 2009.

\bibitem{lei2024autocoder}
B.~Lei, Y.~Li, and Q.~Chen, ``Autocoder: Enhancing code large language model with \textsc{AIEV-Instruct},'' 2024.

\bibitem{gehring2025rlef}
J.~Gehring, K.~Zheng, J.~Copet, V.~Mella, Q.~Carbonneaux, T.~Cohen, and G.~Synnaeve, ``Rlef: Grounding code llms in execution feedback with reinforcement learning,'' 2025.

\bibitem{Wang2024ExecutableCA}
X.~Wang, Y.~Chen, L.~Yuan, Y.~Zhang, Y.~Li, H.~Peng, and H.~Ji, ``Executable code actions elicit better llm agents,'' \emph{ArXiv}, vol. abs/2402.01030, 2024.

\bibitem{schick2023toolformer}
T.~Schick, J.~Dwivedi-Yu, R.~Dess{\`\i}, R.~Raileanu, M.~Lomeli, E.~Hambro, L.~Zettlemoyer, N.~Cancedda, and T.~Scialom, ``Toolformer: Language models can teach themselves to use tools,'' \emph{Advances in Neural Information Processing Systems}, vol.~36, pp. 68\,539--68\,551, 2023.

\bibitem{paranjape2023art}
B.~Paranjape, S.~Lundberg, S.~Singh, H.~Hajishirzi, L.~Zettlemoyer, and M.~T. Ribeiro, ``Art: Automatic multi-step reasoning and tool-use for large language models,'' \emph{arXiv preprint arXiv:2303.09014}, 2023.

\bibitem{song2023restgpt}
Y.~Song, W.~Xiong, D.~Zhu, W.~Wu, H.~Qian, M.~Song, H.~Huang, C.~Li, K.~Wang, R.~Yao, Y.~Tian, and S.~Li, ``Restgpt: Connecting large language models with real-world restful apis,'' 2023.

\bibitem{saha2024sequential}
A.~Saha, L.~Mandal, B.~Ganesan, S.~Ghosh, R.~Sindhgatta, C.~Eberhardt, D.~Debrunner, and S.~Mehta, ``Sequential {API} function calling using {G}raph{QL} schema,'' in \emph{EMNLP}, Y.~Al-Onaizan, M.~Bansal, and Y.-N. Chen, Eds., Miami, Florida, USA, Nov. 2024, pp. 19\,452--19\,458.

\bibitem{yuan2023craft}
L.~Yuan, Y.~Chen, X.~Wang, Y.~R. Fung, H.~Peng, and H.~Ji, ``Craft: Customizing llms by creating and retrieving from specialized toolsets,'' \emph{arXiv preprint arXiv:2309.17428}, 2023.

\bibitem{qian2024toolink}
C.~Qian, C.~Xiong, Z.~Liu, and Z.~Liu, ``Toolink: Linking toolkit creation and using through chain-of-solving on open-source model,'' in \emph{NAACL}, 2024, pp. 831--854.

\bibitem{qian2023creator}
C.~Qian, C.~Han, Y.~Fung, Y.~Qin, Z.~Liu, and H.~Ji, ``{CREATOR}: Tool creation for disentangling abstract and concrete reasoning of large language models,'' in \emph{EMNLP Findings}, H.~Bouamor, J.~Pino, and K.~Bali, Eds., Singapore, Dec. 2023, pp. 6922--6939.

\bibitem{cai2024largelanguagemodelstool}
T.~Cai, X.~Wang, T.~Ma, X.~Chen, and D.~Zhou, ``Large language models as tool makers,'' 2024.

\bibitem{LangChian_2023}
\BIBentryALTinterwordspacing
``{LangChain},'' 1 2023. [Online]. Available: \url{https://github.com/langchain-ai/langchain}
\BIBentrySTDinterwordspacing

\bibitem{Liu_LlamaIndex_2022}
\BIBentryALTinterwordspacing
``{LlamaIndex},'' 11 2022. [Online]. Available: \url{https://github.com/jerryjliu/llama_index}
\BIBentrySTDinterwordspacing

\bibitem{Dify_2023}
\BIBentryALTinterwordspacing
``{Dify},'' 5 2023. [Online]. Available: \url{https://github.com/langgenius/dify}
\BIBentrySTDinterwordspacing

\bibitem{Ollama_2023}
\BIBentryALTinterwordspacing
``{Ollama},'' 7 2023. [Online]. Available: \url{https://github.com/ollama/ollama}
\BIBentrySTDinterwordspacing

\bibitem{MCP_Agent_2025}
\BIBentryALTinterwordspacing
``{MCP Agent},'' 2 2025. [Online]. Available: \url{https://github.com/lastmile-ai/mcp-agent}
\BIBentrySTDinterwordspacing

\bibitem{li2025commercial}
A.~Li, Y.~Zhou, V.~C. Raghuram, T.~Goldstein, and M.~Goldblum, ``Commercial llm agents are already vulnerable to simple yet dangerous attacks,'' \emph{arXiv preprint arXiv:2502.08586}, 2025.

\bibitem{zhang2024agent}
W.~Zhang, K.~Tang, H.~Wu, M.~Wang, Y.~Shen, G.~Hou, Z.~Tan, P.~Li, Y.~Zhuang, and W.~Lu, ``Agent-pro: Learning to evolve via policy-level reflection and optimization,'' in \emph{ACL}, 2024, pp. 5348--5375.

\bibitem{mo2024trembling}
L.~Mo, Z.~Liao, B.~Zheng, Y.~Su, C.~Xiao, and H.~Sun, ``A trembling house of cards? mapping adversarial attacks against language agents,'' \emph{arXiv preprint arXiv:2402.10196}, 2024.

\bibitem{NEURIPS2024_97091a51}
E.~Debenedetti, J.~Zhang, M.~Balunovic, L.~Beurer-Kellner, M.~Fischer, and F.~Tramer, ``Agentdojo: A dynamic environment to evaluate prompt injection attacks and defenses for llm agents,'' in \emph{NeurIPS}, vol.~37, 2024, pp. 82\,895--82\,920.

\bibitem{wu2024adversarial}
C.~H. Wu, J.~Y. Koh, R.~Salakhutdinov, D.~Fried, and A.~Raghunathan, ``Adversarial attacks on multimodal agents,'' \emph{arXiv preprint arXiv:2406.12814}, 2024.

\bibitem{ning2024cheatagent}
L.-b. Ning, S.~Wang, W.~Fan, Q.~Li, X.~Xu, H.~Chen, and F.~Huang, ``Cheatagent: Attacking llm-empowered recommender systems via llm agent,'' in \emph{KDD}, 2024, pp. 2284--2295.

\bibitem{yuinfecting}
W.~Yu, K.~Hu, T.~Pang, C.~Du, M.~Lin, and M.~Fredrikson, ``Infecting llm agents via generalizable adversarial attack,'' in \emph{NeurIPS Workshop}, 2024.

\bibitem{lin2024large}
G.~Lin and Q.~Zhao, ``Large language model sentinel: Llm agent for adversarial purification,'' \emph{arXiv preprint arXiv:2405.20770}, 2024.

\bibitem{chern2024combating}
S.~Chern, Z.~Fan, and A.~Liu, ``Combating adversarial attacks with multi-agent debate,'' \emph{arXiv preprint arXiv:2401.05998}, 2024.

\bibitem{wang2024reinforcement}
X.~Wang, J.~Peng, K.~Xu, H.~Yao, and T.~Chen, ``Reinforcement learning-driven llm agent for automated attacks on llms,'' in \emph{ACL Findings}, 2024, pp. 170--177.

\bibitem{dong2024jailbreaking}
Y.~Dong, Z.~Li, X.~Meng, N.~Yu, and S.~Guo, ``Jailbreaking text-to-image models with llm-based agents,'' \emph{arXiv preprint arXiv:2408.00523}, 2024.

\bibitem{chenllm}
X.~Chen, Y.~Nie, W.~Guo, and X.~Zhang, ``When llm meets drl: Advancing jailbreaking efficiency via drl-guided search,'' in \emph{NeurIPS}, 2024.

\bibitem{lin2024pathseeker}
Z.~Lin, W.~Ma, M.~Zhou, Y.~Zhao, H.~Wang, Y.~Liu, J.~Wang, and L.~Li, ``Pathseeker: Exploring llm security vulnerabilities with a reinforcement learning-based jailbreak approach,'' \emph{arXiv preprint arXiv:2409.14177}, 2024.

\bibitem{zeng2024autodefense}
Y.~Zeng, Y.~Wu, X.~Zhang, H.~Wang, and Q.~Wu, ``Autodefense: Multi-agent llm defense against jailbreak attacks,'' \emph{arXiv preprint arXiv:2403.04783}, 2024.

\bibitem{barua2025guardians}
S.~Barua, M.~Rahman, M.~J. Sadek, R.~Islam, S.~Khaled, and A.~Kabir, ``Guardians of the agentic system: Preventing many shots jailbreak with agentic system,'' \emph{arXiv preprint arXiv:2502.16750}, 2025.

\bibitem{ni2025shieldlearner}
Z.~Ni, H.~Wang, and H.~Wang, ``Shieldlearner: A new paradigm for jailbreak attack defense in llms,'' \emph{arXiv preprint arXiv:2502.13162}, 2025.

\bibitem{zhu2025demonagent}
P.~Zhu, Z.~Zhou, Y.~Zhang, S.~Yan, K.~Wang, and S.~Su, ``Demonagent: Dynamically encrypted multi-backdoor implantation attack on llm-based agent,'' \emph{arXiv preprint arXiv:2502.12575}, 2025.

\bibitem{yang2025watch}
W.~Yang, X.~Bi, Y.~Lin, S.~Chen, J.~Zhou, and X.~Sun, ``Watch out for your agents! investigating backdoor threats to llm-based agents,'' \emph{NeurIPS}, vol.~37, pp. 100\,938--100\,964, 2025.

\bibitem{wang2024badagent}
Y.~Wang, D.~Xue, S.~Zhang, and S.~Qian, ``Badagent: Inserting and activating backdoor attacks in llm agents,'' in \emph{ACL}, 2024, pp. 9811--9827.

\bibitem{tong2025badjudge}
T.~Tong, F.~Wang, Z.~Zhao, and M.~Chen, ``Badjudge: Backdoor vulnerabilities of llm-as-a-judge,'' in \emph{ICLR}, 2025.

\bibitem{guo2025darkmind}
Z.~Guo and R.~Tourani, ``Darkmind: Latent chain-of-thought backdoor in customized llms,'' \emph{arXiv preprint arXiv:2501.18617}, 2025.

\bibitem{zhou2025corba}
Z.~Zhou, Z.~Li, J.~Zhang, Y.~Zhang, K.~Wang, Y.~Liu, and Q.~Guo, ``Corba: Contagious recursive blocking attacks on multi-agent systems based on large language models,'' \emph{arXiv preprint arXiv:2502.14529}, 2025.

\bibitem{he2025red}
P.~He, Y.~Lin, S.~Dong, H.~Xu, Y.~Xing, and H.~Liu, ``Red-teaming llm multi-agent systems via communication attacks,'' \emph{arXiv preprint arXiv:2502.14847}, 2025.

\bibitem{yu2024netsafe}
M.~Yu, S.~Wang, G.~Zhang, J.~Mao, C.~Yin, Q.~Liu, Q.~Wen, K.~Wang, and Y.~Wang, ``Netsafe: Exploring the topological safety of multi-agent networks,'' \emph{arXiv preprint arXiv:2410.15686}, 2024.

\bibitem{wang2025g}
S.~Wang, G.~Zhang, M.~Yu, G.~Wan, F.~Meng, C.~Guo, K.~Wang, and Y.~Wang, ``G-safeguard: A topology-guided security lens and treatment on llm-based multi-agent systems,'' \emph{arXiv preprint arXiv:2502.11127}, 2025.

\bibitem{hua2024trustagent}
W.~Hua, X.~Yang, M.~Jin, Z.~Li, W.~Cheng, R.~Tang, and Y.~Zhang, ``Trustagent: Towards safe and trustworthy llm-based agents through agent constitution,'' in \emph{EMNLP Findings}, 2024.

\bibitem{zhang2024psysafe}
Z.~Zhang, Y.~Zhang, L.~Li, H.~Gao, L.~Wang, H.~Lu, F.~Zhao, Y.~Qiao, and J.~Shao, ``Psysafe: A comprehensive framework for psychological-based attack, defense, and evaluation of multi-agent system safety,'' \emph{arXiv preprint arXiv:2401.11880}, 2024.

\bibitem{deng2024ai}
Z.~Deng, Y.~Guo, C.~Han, W.~Ma, J.~Xiong, S.~Wen, and Y.~Xiang, ``Ai agents under threat: A survey of key security challenges and future pathways,'' \emph{ACM Computing Surveys}, 2024.

\bibitem{debenedetti2025agentdojo}
E.~Debenedetti, J.~Zhang, M.~Balunovic, L.~Beurer-Kellner, M.~Fischer, and F.~Tram{\`e}r, ``Agentdojo: A dynamic environment to evaluate prompt injection attacks and defenses for llm agents,'' \emph{NeurIPS}, vol.~37, pp. 82\,895--82\,920, 2025.

\bibitem{li2024targeting}
X.~Li, Z.~Li, Y.~Kosuga, Y.~Yoshida, and V.~Bian, ``Targeting the core: A simple and effective method to attack rag-based agents via direct llm manipulation,'' \emph{arXiv preprint arXiv:2412.04415}, 2024.

\bibitem{zhan2024injecagent}
Q.~Zhan, Z.~Liang, Z.~Ying, and D.~Kang, ``Injecagent: Benchmarking indirect prompt injections in tool-integrated large language model agents,'' \emph{arXiv preprint arXiv:2403.02691}, 2024.

\bibitem{pasquini2024hacking}
D.~Pasquini, E.~M. Kornaropoulos, and G.~Ateniese, ``Hacking back the ai-hacker: Prompt injection as a defense against llm-driven cyberattacks,'' \emph{arXiv preprint arXiv:2410.20911}, 2024.

\bibitem{abdelnabi2025firewalls}
S.~Abdelnabi, A.~Gomaa, E.~Bagdasarian, P.~O. Kristensson, and R.~Shokri, ``Firewalls to secure dynamic llm agentic networks,'' \emph{arXiv preprint arXiv:2502.01822}, 2025.

\bibitem{zhong2025rtbas}
P.~Y. Zhong, S.~Chen, R.~Wang, M.~McCall, B.~L. Titzer, and H.~Miller, ``Rtbas: Defending llm agents against prompt injection and privacy leakage,'' \emph{arXiv preprint arXiv:2502.08966}, 2025.

\bibitem{jia2024task}
F.~Jia, T.~Wu, X.~Qin, and A.~Squicciarini, ``The task shield: Enforcing task alignment to defend against indirect prompt injection in llm agents,'' \emph{arXiv preprint arXiv:2412.16682}, 2024.

\bibitem{tian2023evil}
Y.~Tian, X.~Yang, J.~Zhang, Y.~Dong, and H.~Su, ``Evil geniuses: Delving into the safety of llm-based agents,'' \emph{arXiv preprint arXiv:2311.11855}, 2023.

\bibitem{wang2024biorag}
C.~Wang, Q.~Long, X.~Meng, X.~Cai, C.~Wu, Z.~Meng, X.~Wang, and Y.~Zhou, ``Biorag: A rag-llm framework for biological question reasoning,'' \emph{arXiv preprint arXiv:2408.01107}, 2024.

\bibitem{gan2024navigating}
Y.~Gan, Y.~Yang, Z.~Ma, P.~He, R.~Zeng, Y.~Wang, Q.~Li, C.~Zhou, S.~Li, T.~Wang \emph{et~al.}, ``Navigating the risks: A survey of security, privacy, and ethics threats in llm-based agents,'' \emph{arXiv preprint arXiv:2411.09523}, 2024.

\bibitem{xiang2024clas}
Z.~Xiang, Y.~Zeng, M.~Kang, C.~Xu, J.~Zhang, Z.~Yuan, Z.~Chen, C.~Xie, F.~Jiang, M.~Pan \emph{et~al.}, ``Clas 2024: The competition for llm and agent safety,'' in \emph{NeurIPS Workshop}, 2024.

\bibitem{wu2024wipi}
F.~Wu, S.~Wu, Y.~Cao, and C.~Xiao, ``Wipi: A new web threat for llm-driven web agents,'' \emph{arXiv preprint arXiv:2402.16965}, 2024.

\bibitem{nakash2024breaking}
I.~Nakash, G.~Kour, G.~Uziel, and A.~Anaby-Tavor, ``Breaking react agents: Foot-in-the-door attack will get you in,'' \emph{arXiv preprint arXiv:2410.16950}, 2024.

\bibitem{chen2025agentpoison}
Z.~Chen, Z.~Xiang, C.~Xiao, D.~Song, and B.~Li, ``Agentpoison: Red-teaming llm agents via poisoning memory or knowledge bases,'' \emph{NeurIPS}, vol.~37, pp. 130\,185--130\,213, 2025.

\bibitem{wang2025unveiling}
B.~Wang, W.~He, P.~He, S.~Zeng, Z.~Xiang, Y.~Xing, and J.~Tang, ``Unveiling privacy risks in llm agent memory,'' \emph{arXiv preprint arXiv:2502.13172}, 2025.

\bibitem{EmbraceTheRed}
E.~T. Red, ``Malicious chatgpt agents: How gpts can quietly grab your data (demo),'' \emph{Embrace The Red}, 2023.

\bibitem{li2024personal}
Y.~Li, H.~Wen, W.~Wang, X.~Li, Y.~Yuan, G.~Liu, J.~Liu, W.~Xu, X.~Wang, Y.~Sun \emph{et~al.}, ``Personal llm agents: Insights and survey about the capability, efficiency and security,'' \emph{arXiv preprint arXiv:2401.05459}, 2024.

\bibitem{gu2024agent}
X.~Gu, X.~Zheng, T.~Pang, C.~Du, Q.~Liu, Y.~Wang, J.~Jiang, and M.~Lin, ``Agent smith: A single image can jailbreak one million multimodal llm agents exponentially fast,'' \emph{arXiv preprint arXiv:2402.08567}, 2024.

\bibitem{lee2024prompt}
D.~Lee and M.~Tiwari, ``Prompt infection: Llm-to-llm prompt injection within multi-agent systems,'' \emph{arXiv preprint arXiv:2410.07283}, 2024.

\bibitem{chen2024blockagents}
B.~Chen, G.~Li, X.~Lin, Z.~Wang, and J.~Li, ``Blockagents: Towards byzantine-robust llm-based multi-agent coordination via blockchain,'' in \emph{ACM Turing Award Celebration Conference}, 2024, pp. 187--192.

\bibitem{andriushchenko2024agentharm}
M.~Andriushchenko, A.~Souly, M.~Dziemian, D.~Duenas, M.~Lin, J.~Wang, D.~Hendrycks, A.~Zou, Z.~Kolter, M.~Fredrikson \emph{et~al.}, ``Agentharm: A benchmark for measuring harmfulness of llm agents,'' \emph{arXiv preprint arXiv:2410.09024}, 2024.

\bibitem{carlini2021extracting}
N.~Carlini, F.~Tramer, E.~Wallace, M.~Jagielski, A.~Herbert-Voss, K.~Lee, A.~Roberts, T.~Brown, D.~Song, U.~Erlingsson \emph{et~al.}, ``Extracting training data from large language models,'' in \emph{USENIX}, 2021, pp. 2633--2650.

\bibitem{carlini2022quantifying}
N.~Carlini, D.~Ippolito, M.~Jagielski, K.~Lee, F.~Tramer, and C.~Zhang, ``Quantifying memorization across neural language models,'' in \emph{ICLR}, 2022.

\bibitem{huang2022large}
J.~Huang, H.~Shao, and K.~C.-C. Chang, ``Are large pre-trained language models leaking your personal information?'' \emph{arXiv preprint arXiv:2205.12628}, 2022.

\bibitem{mireshghallah2022quantifying}
F.~Mireshghallah, K.~Goyal, A.~Uniyal, T.~Berg-Kirkpatrick, and R.~Shokri, ``Quantifying privacy risks of masked language models using membership inference attacks,'' \emph{arXiv preprint arXiv:2203.03929}, 2022.

\bibitem{fu2023practical}
W.~Fu, H.~Wang, C.~Gao, G.~Liu, Y.~Li, and T.~Jiang, ``Practical membership inference attacks against fine-tuned large language models via self-prompt calibration,'' \emph{arXiv preprint arXiv:2311.06062}, 2023.

\bibitem{hoory2021learning}
S.~Hoory, A.~Feder, A.~Tendler, S.~Erell, A.~Peled-Cohen, I.~Laish, H.~Nakhost, U.~Stemmer, A.~Benjamini, A.~Hassidim \emph{et~al.}, ``Learning and evaluating a differentially private pre-trained language model,'' in \emph{EMNLP Findings}, 2021, pp. 1178--1189.

\bibitem{kang2023knowledge}
M.~Kang, S.~Lee, J.~Baek, K.~Kawaguchi, and S.~J. Hwang, ``Knowledge-augmented reasoning distillation for small language models in knowledge-intensive tasks,'' \emph{NeurIPS}, vol.~36, pp. 48\,573--48\,602, 2023.

\bibitem{pan2020privacy}
X.~Pan, M.~Zhang, S.~Ji, and M.~Yang, ``Privacy risks of general-purpose language models,'' in \emph{IEEE Symposium on Security and Privacy (SP)}.\hskip 1em plus 0.5em minus 0.4em\relax IEEE, 2020, pp. 1314--1331.

\bibitem{wang2024property}
L.~Wang, J.~Wang, J.~Wan, L.~Long, Z.~Yang, and Z.~Qin, ``Property existence inference against generative models,'' in \emph{USENIX}, 2024, pp. 2423--2440.

\bibitem{kandpal2022deduplicating}
N.~Kandpal, E.~Wallace, and C.~Raffel, ``Deduplicating training data mitigates privacy risks in language models,'' in \emph{ICML}.\hskip 1em plus 0.5em minus 0.4em\relax PMLR, 2022, pp. 10\,697--10\,707.

\bibitem{kim2023propile}
S.~Kim, S.~Yun, H.~Lee, M.~Gubri, S.~Yoon, and S.~J. Oh, ``Propile: Probing privacy leakage in large language models,'' \emph{NeurIPS}, vol.~36, pp. 20\,750--20\,762, 2023.

\bibitem{krishna2019thieves}
K.~Krishna, G.~S. Tomar, A.~P. Parikh, N.~Papernot, and M.~Iyyer, ``Thieves on sesame street! model extraction of bert-based apis,'' \emph{arXiv preprint arXiv:1910.12366}, 2019.

\bibitem{naseh2023stealing}
A.~Naseh, K.~Krishna, M.~Iyyer, and A.~Houmansadr, ``Stealing the decoding algorithms of language models,'' in \emph{ACM SIGSAC}, 2023, pp. 1835--1849.

\bibitem{li2024extracting}
Z.~Li, C.~Wang, P.~Ma, C.~Liu, S.~Wang, D.~Wu, C.~Gao, and Y.~Liu, ``On extracting specialized code abilities from large language models: A feasibility study,'' in \emph{ICSE}, 2024, pp. 1--13.

\bibitem{kirchenbauer2023watermark}
J.~Kirchenbauer, J.~Geiping, Y.~Wen, J.~Katz, I.~Miers, and T.~Goldstein, ``A watermark for large language models,'' in \emph{ICML}.\hskip 1em plus 0.5em minus 0.4em\relax PMLR, 2023, pp. 17\,061--17\,084.

\bibitem{lin2024blockchain}
Y.~Lin, Z.~Gao, H.~Du, D.~Niyato, J.~Kang, Z.~Xiong, and Z.~Zheng, ``Blockchain-based efficient and trustworthy aigc services in metaverse,'' \emph{IEEE Transactions on Services Computing}, 2024.

\bibitem{shen2024prompt}
X.~Shen, Y.~Qu, M.~Backes, and Y.~Zhang, ``Prompt stealing attacks against $\{$Text-to-Image$\}$ generation models,'' in \emph{USENIX}, 2024, pp. 5823--5840.

\bibitem{sha2024prompt}
Z.~Sha and Y.~Zhang, ``Prompt stealing attacks against large language models,'' \emph{arXiv preprint arXiv:2402.12959}, 2024.

\bibitem{hui2024pleak}
B.~Hui, H.~Yuan, N.~Gong, P.~Burlina, and Y.~Cao, ``Pleak: Prompt leaking attacks against large language model applications,'' in \emph{ACM SIGSAC}, 2024, pp. 3600--3614.

\bibitem{bommasani2021opportunities}
R.~Bommasani, D.~A. Hudson, E.~Adeli, R.~Altman, S.~Arora, S.~von Arx, M.~S. Bernstein, J.~Bohg, A.~Bosselut, E.~Brunskill \emph{et~al.}, ``{On the Opportunities and Risks of Foundation Models},'' \emph{arXiv preprint arXiv:2108.07258}, 2021.

\bibitem{floridi2020gpt}
L.~Floridi and M.~Chiriatti, ``{GPT-3: Its Nature, Scope, Limits, and Consequences},'' \emph{Minds and Machines}, vol.~30, pp. 681--694, 2020.

\bibitem{touvron2023llama}
H.~Touvron, T.~Lavril, G.~Izacard, X.~Martinet, M.-A. Lachaux, T.~Lacroix, B.~Rozi{\`e}re, N.~Goyal, E.~Hambro, F.~Azhar \emph{et~al.}, ``{LLaMA: Open and Efficient Foundation Language Models},'' \emph{arXiv preprint arXiv:2302.13971}, 2023.

\bibitem{tadas2024redefining}
P.~Tadas and S.~Agarmore, ``{Redefining Work in the Age of AI: Challenges and Pathways to Opportunities},'' in \emph{SPICES}.\hskip 1em plus 0.5em minus 0.4em\relax IEEE, 2024, pp. 1--5.

\bibitem{moore2023empowering}
S.~Moore, R.~Tong, A.~Singh, Z.~Liu, X.~Hu, Y.~Lu, J.~Liang, C.~Cao, H.~Khosravi, P.~Denny \emph{et~al.}, ``{Empowering Education with LLMs - The Next-Gen Interface and Content Generation},'' in \emph{International Conference on Artificial Intelligence in Education}.\hskip 1em plus 0.5em minus 0.4em\relax Springer, 2023, pp. 32--37.

\bibitem{liu2025culturevlm}
S.~Liu, Y.~Jin, C.~Li, D.~F. Wong, Q.~Wen, L.~Sun, H.~Chen, X.~Xie, and J.~Wang, ``Culturevlm: Characterizing and improving cultural understanding of vision-language models for over 100 countries,'' \emph{arXiv:2501.01282}, 2025.

\bibitem{henderson2023foundation}
P.~Henderson, X.~Li, D.~Jurafsky, T.~Hashimoto, M.~A. Lemley, and P.~Liang, ``{Foundation Models and Fair Use},'' \emph{JMLR}, vol.~24, no. 400, pp. 1--79, 2023.

\bibitem{lemley2020fair}
M.~A. Lemley and B.~Casey, ``{Fair Learning},'' \emph{Tex. L. Rev.}, vol.~99, p. 743, 2020.

\bibitem{oh2024uniguard}
S.~Oh, Y.~Jin, M.~Sharma, D.~Kim, E.~Ma, G.~Verma, and S.~Kumar, ``Uniguard: Towards universal safety guardrails for jailbreak attacks on multimodal large language models,'' \emph{arXiv:2411.01703}, 2024.

\bibitem{bender2021dangers}
E.~M. Bender, T.~Gebru, A.~McMillan-Major, and S.~Shmitchell, ``{On the Dangers of Stochastic Parrots: Can Language Models Be Too Big?}'' in \emph{FAccT}, 2021, pp. 610--623.

\bibitem{brundage2020toward}
M.~Brundage, S.~Avin, J.~Wang, H.~Belfield, G.~Krueger, G.~Hadfield, H.~Khlaaf, J.~Yang, H.~Toner, R.~Fong \emph{et~al.}, ``{Toward Trustworthy AI Development: Mechanisms for Supporting Verifiable Claims},'' \emph{arXiv preprint arXiv:2004.07213}, 2020.

\bibitem{ganguli2022predictability}
D.~Ganguli, D.~Hernandez, L.~Lovitt, A.~Askell, Y.~Bai, A.~Chen, T.~Conerly, N.~Dassarma, D.~Drain, N.~Elhage \emph{et~al.}, ``{Predictability and Surprise in Large Generative Models},'' in \emph{FAccT}, 2022, pp. 1747--1764.

\bibitem{deng2024deconstructing}
C.~Deng, Y.~Duan, X.~Jin, H.~Chang, Y.~Tian, H.~Liu, H.~P. Zou, Y.~Jin, Y.~Xiao, Y.~Wang \emph{et~al.}, ``{Deconstructing The Ethics of Large Language Models from Long-standing Issues to New-emerging Dilemmas: A Survey},'' \emph{arXiv e-prints}, pp. arXiv--2406, 2024.

\bibitem{shumailov2024ai}
I.~Shumailov, Z.~Shumaylov, Y.~Zhao, N.~Papernot, R.~Anderson, and Y.~Gal, ``{AI models collapse when trained on recursively generated data},'' \emph{Nature}, vol. 631, no. 8022, pp. 755--759, 2024.

\bibitem{weidinger2021ethical}
L.~Weidinger, J.~Mellor, M.~Rauh, C.~Griffin, J.~Uesato, P.-S. Huang, M.~Cheng, M.~Glaese, B.~Balle, A.~Kasirzadeh \emph{et~al.}, ``{Ethical and social risks of harm from Language Models},'' \emph{arXiv preprint arXiv:2112.04359}, 2021.

\bibitem{xiao2024large}
Y.~Xiao, Y.~Jin, Y.~Bai, Y.~Wu, X.~Yang, X.~Luo, W.~Yu, X.~Zhao, Y.~Liu, Q.~Gu \emph{et~al.}, ``Large language models can be contextual privacy protection learners,'' in \emph{EMNLP}, 2024, pp. 14\,179--14\,201.

\bibitem{alber2025medical}
D.~A. Alber, Z.~Yang, A.~Alyakin, E.~Yang, S.~Rai, A.~A. Valliani, J.~Zhang, G.~R. Rosenbaum, A.~K. Amend-Thomas, D.~B. Kurland \emph{et~al.}, ``{Medical large language models are vulnerable to data-poisoning attacks},'' \emph{Nature Medicine}, pp. 1--9, 2025.

\bibitem{jin2022towards}
Y.~Jin, X.~Wang, R.~Yang, Y.~Sun, W.~Wang, H.~Liao, and X.~Xie, ``Towards fine-grained reasoning for fake news detection,'' in \emph{AAAI}, vol.~36, no.~5, 2022, pp. 5746--5754.

\bibitem{shen2023large}
T.~Shen, R.~Jin, Y.~Huang, C.~Liu, W.~Dong, Z.~Guo, X.~Wu, Y.~Liu, and D.~Xiong, ``{Large Language Model Alignment: A Survey},'' \emph{arXiv preprint arXiv:2309.15025}, 2023.

\bibitem{luccioni2023estimating}
A.~S. Luccioni, S.~Viguier, and A.-L. Ligozat, ``{Estimating the Carbon Footprint of BLOOM, a 176B Parameter Language Model},'' \emph{JMLR}, vol.~24, no. 253, pp. 1--15, 2023.

\bibitem{strubell2020energy}
E.~Strubell, A.~Ganesh, and A.~McCallum, ``{Energy and Policy Considerations for Deep Learning in NLP},'' in \emph{AAAI}, vol.~34, no.~09, 2020, pp. 13\,693--13\,696.

\bibitem{zhou2024awesome}
J.~Zhou, ``Awesome ai agents for scientific discovery,'' \url{https://github.com/zhoujieli/Awesome-LLM-Agents-Scientific-Discovery}, 2024.

\bibitem{AAAIpp}
\BIBentryALTinterwordspacing
AAAI, ``Aaai 2025 presidential panel: Future of ai research,'' 2025. [Online]. Available: \url{https://aaai.org/wp-content/uploads/2025/03/AAAI-2025-PresPanel-Report-FINAL.pdf}
\BIBentrySTDinterwordspacing

\bibitem{sciagents}
\BIBentryALTinterwordspacing
A.~Ghafarollahi and M.~J. Buehler, ``Sciagents: Automating scientific discovery through bioinspired multi-agent intelligent graph reasoning,'' \emph{Advanced Materials}, vol. n/a, no. n/a, p. 2413523. [Online]. Available: \url{https://advanced.onlinelibrary.wiley.com/doi/abs/10.1002/adma.202413523}
\BIBentrySTDinterwordspacing

\bibitem{kon2025curierigorousautomatedscientific}
\BIBentryALTinterwordspacing
P.~T.~J. Kon, J.~Liu, Q.~Ding, Y.~Qiu, Z.~Yang, Y.~Huang, J.~Srinivasa, M.~Lee, M.~Chowdhury, and A.~Chen, ``Curie: Toward rigorous and automated scientific experimentation with ai agents,'' 2025. [Online]. Available: \url{https://arxiv.org/abs/2502.16069}
\BIBentrySTDinterwordspacing

\bibitem{jin2024agentreview}
Y.~Jin, Q.~Zhao, Y.~Wang, H.~Chen, K.~Zhu, Y.~Xiao, and J.~Wang, ``Agentreview: Exploring peer review dynamics with llm agents,'' in \emph{EMNLP}, 2024, pp. 1208--1226.

\bibitem{bran2023chemcrowaugmentinglargelanguagemodels}
\BIBentryALTinterwordspacing
A.~M. Bran, S.~Cox, O.~Schilter, C.~Baldassari, A.~D. White, and P.~Schwaller, ``Chemcrow: Augmenting large-language models with chemistry tools,'' 2023. [Online]. Available: \url{https://arxiv.org/abs/2304.05376}
\BIBentrySTDinterwordspacing

\bibitem{ghafarollahi2024atomagentsalloydesigndiscovery}
\BIBentryALTinterwordspacing
A.~Ghafarollahi and M.~J. Buehler, ``Atomagents: Alloy design and discovery through physics-aware multi-modal multi-agent artificial intelligence,'' 2024. [Online]. Available: \url{https://arxiv.org/abs/2407.10022}
\BIBentrySTDinterwordspacing

\bibitem{kostunin2025aiagentsgroundbasedgamma}
\BIBentryALTinterwordspacing
D.~Kostunin, V.~Sotnikov, S.~Golovachev, and A.~Strube, ``Ai agents for ground-based gamma astronomy,'' 2025. [Online]. Available: \url{https://arxiv.org/abs/2503.00821}
\BIBentrySTDinterwordspacing

\bibitem{qi2024largelanguagemodelsbiomedical}
\BIBentryALTinterwordspacing
B.~Qi, K.~Zhang, K.~Tian, H.~Li, Z.-R. Chen, S.~Zeng, E.~Hua, H.~Jinfang, and B.~Zhou, ``Large language models as biomedical hypothesis generators: A comprehensive evaluation,'' 2024. [Online]. Available: \url{https://arxiv.org/abs/2407.08940}
\BIBentrySTDinterwordspacing

\bibitem{roohani2024biodiscoveryagent}
Y.~Roohani, A.~Lee, Q.~Huang, J.~Vora, Z.~Steinhart, K.~Huang, A.~Marson, P.~Liang, and J.~Leskovec, ``Biodiscoveryagent: An ai agent for designing genetic perturbation experiments,'' \emph{arXiv preprint arXiv:2405.17631}, 2024.

\bibitem{wang2024geneagentselfverificationlanguageagent}
\BIBentryALTinterwordspacing
Z.~Wang, Q.~Jin, C.-H. Wei, S.~Tian, P.-T. Lai, Q.~Zhu, C.-P. Day, C.~Ross, and Z.~Lu, ``Geneagent: Self-verification language agent for gene set knowledge discovery using domain databases,'' 2024. [Online]. Available: \url{https://arxiv.org/abs/2405.16205}
\BIBentrySTDinterwordspacing

\bibitem{xiao2025knowledge}
M.~Xiao, W.~Zhang, X.~Huang, H.~Zhu, M.~Wu, X.~Li, and Y.~Zhou, ``Knowledge-guided biomarker identification for label-free single-cell rna-seq data: A reinforcement learning perspective,'' \emph{arXiv preprint arXiv:2501.04718}, 2025.

\bibitem{sun2024pathgen16m16millionpathology}
\BIBentryALTinterwordspacing
Y.~Sun, Y.~Zhang, Y.~Si, C.~Zhu, Z.~Shui, K.~Zhang, J.~Li, X.~Lyu, T.~Lin, and L.~Yang, ``Pathgen-1.6m: 1.6 million pathology image-text pairs generation through multi-agent collaboration,'' 2024. [Online]. Available: \url{https://arxiv.org/abs/2407.00203}
\BIBentrySTDinterwordspacing

\bibitem{cai2025knowledge}
X.~Cai, C.~Wang, Q.~Long, Y.~Zhou, and M.~Xiao, ``Knowledge hierarchy guided biological-medical dataset distillation for domain llm training,'' \emph{arXiv preprint arXiv:2501.15108}, 2025.

\bibitem{chen2024genesum}
Z.~Chen, C.~Hu, M.~Wu, Q.~Long, X.~Wang, Y.~Zhou, and M.~Xiao, ``Genesum: Large language model-based gene summary extraction,'' in \emph{2024 IEEE International Conference on Bioinformatics and Biomedicine (BIBM)}.\hskip 1em plus 0.5em minus 0.4em\relax IEEE, 2024, pp. 1438--1443.

\bibitem{keshavjee2006best}
K.~Keshavjee, J.~Bosomworth, J.~Copen, J.~Lai, B.~Kucukyazici, R.~Lilani, and A.~M. Holbrook, ``Best practices in emr implementation: a systematic review,'' in \emph{AMIA Annual Symposium Proceedings}, vol. 2006, 2006, p. 982.

\bibitem{ye2023needed}
X.~Ye, M.~Xiao, Z.~Ning, W.~Dai, W.~Cui, Y.~Du, and Y.~Zhou, ``Needed: Introducing hierarchical transformer to eye diseases diagnosis,'' in \emph{Proceedings of the 2023 SIAM International Conference on Data Mining (SDM)}.\hskip 1em plus 0.5em minus 0.4em\relax SIAM, 2023, pp. 667--675.

\bibitem{li2024agenthospital}
J.~Li, Y.~Lai, W.~Li, J.~Ren, M.~Zhang, X.~Kang, S.~Wang, P.~Li, Y.-Q. Zhang, W.~Ma \emph{et~al.}, ``Agent hospital: A simulacrum of hospital with evolvable medical agents,'' \emph{arXiv preprint arXiv:2405.02957}, 2024.

\bibitem{yan2024clinicallab}
W.~Yan, H.~Liu, T.~Wu, Q.~Chen, W.~Wang, H.~Chai, J.~Wang, W.~Zhao, Y.~Zhang, R.~Zhang \emph{et~al.}, ``Clinicallab: Aligning agents for multi-departmental clinical diagnostics in the real world,'' \emph{arXiv preprint arXiv:2406.13890}, 2024.

\bibitem{yu2024aipatientsimulatingpatientsehrs}
\BIBentryALTinterwordspacing
H.~Yu, J.~Zhou, L.~Li, S.~Chen, J.~Gallifant, A.~Shi, X.~Li, W.~Hua, M.~Jin, G.~Chen, Y.~Zhou, Z.~Li, T.~Gupte, M.-L. Chen, Z.~Azizi, Y.~Zhang, T.~L. Assimes, X.~Ma, D.~S. Bitterman, L.~Lu, and L.~Fan, ``Aipatient: Simulating patients with ehrs and llm powered agentic workflow,'' 2024. [Online]. Available: \url{https://arxiv.org/abs/2409.18924}
\BIBentrySTDinterwordspacing

\bibitem{sharma2024cxr}
N.~Sharma, ``Cxr-agent: Vision-language models for chest x-ray interpretation with uncertainty aware radiology reporting,'' \emph{arXiv preprint arXiv:2407.08811}, 2024.

\bibitem{fallahpour2025medraxmedicalreasoningagent}
\BIBentryALTinterwordspacing
A.~Fallahpour, J.~Ma, A.~Munim, H.~Lyu, and B.~Wang, ``Medrax: Medical reasoning agent for chest x-ray,'' 2025. [Online]. Available: \url{https://arxiv.org/abs/2502.02673}
\BIBentrySTDinterwordspacing

\bibitem{lee2024comparative}
R.~W. Lee, K.~H. Lee, J.~S. Yun, M.~S. Kim, and H.~S. Choi, ``Comparative analysis of m4cxr, an llm-based chest x-ray report generation model, and chatgpt in radiological interpretation,'' \emph{Journal of Clinical Medicine}, vol.~13, no.~23, p. 7057, 2024.

\bibitem{feng2023chessgpt}
X.~Feng, Y.~Luo, Z.~Wang, H.~Tang, M.~Yang, K.~Shao, D.~Mguni, Y.~Du, and J.~Wang, ``Chessgpt: Bridging policy learning and language modeling,'' in \emph{NeurIPS}, 2023, pp. 7216--7262.

\bibitem{carta2023grounding}
T.~Carta, C.~Romac, T.~Wolf, S.~Lamprier, O.~Sigaud, and P.-Y. Oudeyer, ``Grounding large language models in interactive environments with online reinforcement learning,'' in \emph{ICML}, 2023, pp. 3676--3713.

\bibitem{zhu2023calypso}
A.~Zhu, L.~Martin, A.~Head, and C.~Callison-Burch, ``Calypso: Llms as dungeon master's assistants,'' in \emph{AAAI}, 2023, pp. 380--390.

\bibitem{chen2023gamegpt}
D.~Chen, H.~Wang, Y.~Huo, Y.~Li, and H.~Zhang, ``Gamegpt: Multi-agent collaborative framework for game development,'' \emph{arXiv preprint arXiv:2310.08067}, 2023.

\bibitem{sun2023language}
Y.~Sun, Z.~Li, K.~Fang, C.~H. Lee, and A.~Asadipour, ``Language as reality: a co-creative storytelling game experience in 1001 nights using generative ai,'' in \emph{AAAI}, 2023, pp. 425--434.

\bibitem{li2023econagent}
N.~Li, C.~Gao, M.~Li, Y.~Li, and Q.~Liao, ``Econagent: large language model-empowered agents for simulating macroeconomic activities,'' \emph{ACL}, pp. 15\,523--15\,536, 2024.

\bibitem{li2023tradinggpt}
Y.~Li, Y.~Yu, H.~Li, Z.~Chen, and K.~Khashanah, ``Tradinggpt: Multi-agent system with layered memory and distinct characters for enhanced financial trading performance,'' \emph{arXiv preprint arXiv:2309.03736}, 2023.

\bibitem{zhao2024competeai}
Q.~Zhao, J.~Wang, Y.~Zhang, Y.~Jin, K.~Zhu, H.~Chen, and X.~Xie, ``Competeai: Understanding the competition dynamics in large language model-based agents,'' in \emph{ICML}, 2024, pp. 61\,092--61\,107.

\bibitem{ma2024understanding}
Z.~Ma, Y.~Mei, and Z.~Su, ``Understanding the benefits and challenges of using large language model-based conversational agents for mental well-being support,'' in \emph{AMIA Annual Symposium Proceedings}, vol. 2023, 2024, p. 1105.

\bibitem{zhang2024exploring}
J.~Zhang, X.~Xu, N.~Zhang, R.~Liu, B.~Hooi, and S.~Deng, ``Exploring collaboration mechanisms for llm agents: A social psychology view,'' in \emph{ACL}, 2024, pp. 14\,544--14\,607.

\bibitem{aher2023using}
G.~V. Aher, R.~I. Arriaga, and A.~T. Kalai, ``Using large language models to simulate multiple humans and replicate human subject studies,'' in \emph{ICML}, 2023, pp. 337--371.

\bibitem{liu2024training}
R.~Liu, R.~Yang, C.~Jia, G.~Zhang, D.~Zhou, A.~M. Dai, D.~Yang, and S.~Vosoughi, ``Training socially aligned language models on simulated social interactions,'' in \emph{ICLR}, 2024.

\bibitem{gao2023s3}
C.~Gao, X.~Lan, Z.~Lu, J.~Mao, J.~Piao, H.~Wang, D.~Jin, and Y.~Li, ``S3: Social-network simulation system with large language model-empowered agents,'' \emph{arXiv preprint arXiv:2307.14984}, 2023.

\bibitem{dong2024self}
Y.~Dong, X.~Jiang, Z.~Jin, and G.~Li, ``Self-collaboration code generation via chatgpt,'' \emph{ACM Transactions on Software Engineering and Methodology}, vol.~33, no.~7, pp. 1--38, 2024.

\bibitem{qian2024communicative}
C.~Qian, X.~Cong, C.~Yang, W.~Chen, Y.~Su, J.~Xu, Z.~Liu, and M.~Sun, ``Chatdev: Communicative agents for software development,'' in \emph{ACL}, 2024, pp. 15\,174--15\,186.

\bibitem{zhang2024generative}
A.~Zhang, Y.~Chen, L.~Sheng, X.~Wang, and T.-S. Chua, ``On generative agents in recommendation,'' in \emph{SIGIR}, 2024, pp. 1807--1817.

\bibitem{zhang2024agentcf}
J.~Zhang, Y.~Hou, R.~Xie, W.~Sun, J.~McAuley, W.~X. Zhao, L.~Lin, and J.-R. Wen, ``Agentcf: Collaborative learning with autonomous language agents for recommender systems,'' in \emph{WWW}, 2024, pp. 3679--3689.

\bibitem{wang2024macrec}
Z.~Wang, Y.~Yu, W.~Zheng, W.~Ma, and M.~Zhang, ``Macrec: A multi-agent collaboration framework for recommendation,'' in \emph{SIGIR}, 2024, pp. 2760--2764.

\bibitem{wang2023recmind}
Y.~Wang, Z.~Jiang, Z.~Chen, F.~Yang, Y.~Zhou, E.~Cho, X.~Fan, X.~Huang, Y.~Lu, and Y.~Yang, ``Recmind: Large language model powered agent for recommendation,'' \emph{arXiv preprint arXiv:2308.14296}, 2023.

\bibitem{qian2024scaling}
C.~Qian, Z.~Xie, Y.~Wang, W.~Liu, Y.~Dang, Z.~Du, W.~Chen, C.~Yang, Z.~Liu, and M.~Sun, ``Scaling large-language-model-based multi-agent collaboration,'' \emph{arXiv:2406.07155}, 2024.

\bibitem{chan2023chateval}
C.-M. Chan, W.~Chen, Y.~Su, J.~Yu, W.~Xue, S.~Zhang, J.~Fu, and Z.~Liu, ``Chateval: Towards better llm-based evaluators through multi-agent debate,'' \emph{arXiv preprint arXiv:2308.07201}, 2023.

\bibitem{rana2000scalability}
O.~F. Rana and K.~Stout, ``What is scalability in multi-agent systems?'' in \emph{Proceedings of the fourth international conference on Autonomous agents}, 2000, pp. 56--63.

\bibitem{deters2001scalable}
R.~Deters, ``Scalable multi-agent systems,'' in \emph{Proceedings of the 2001 joint ACM-ISCOPE conference on Java Grande}, 2001, p. 182.

\bibitem{verma2024adaptagent}
G.~Verma, R.~Kaur, N.~Srishankar, Z.~Zeng, T.~Balch, and M.~Veloso, ``Adaptagent: Adapting multimodal web agents with few-shot learning from human demonstrations,'' \emph{arXiv preprint arXiv:2411.13451}, 2024.

\bibitem{jin2024mm}
Y.~Jin, M.~Choi, G.~Verma, J.~Wang, and S.~Kumar, ``Mm-soc: Benchmarking multimodal large language models in social media platforms,'' in \emph{ACL Findings}, 2024.

\bibitem{yao2024velo}
Z.~Yao, Z.~Tang, J.~Lou, P.~Shen, and W.~Jia, ``Velo: A vector database-assisted cloud-edge collaborative llm qos optimization framework,'' in \emph{ICWS}.\hskip 1em plus 0.5em minus 0.4em\relax IEEE, 2024, pp. 865--876.

\bibitem{cheng2024xrag}
X.~Cheng, X.~Wang, X.~Zhang, T.~Ge, S.-Q. Chen, F.~Wei, H.~Zhang, and D.~Zhao, ``xrag: Extreme context compression for retrieval-augmented generation with one token,'' in \emph{NeurIPS}, 2024.

\bibitem{jin2024better}
Y.~Jin, M.~Chandra, G.~Verma, Y.~Hu, M.~De~Choudhury, and S.~Kumar, ``Better to ask in english: Cross-lingual evaluation of large language models for healthcare queries,'' in \emph{WWW}, 2024, pp. 2627--2638.

\bibitem{agarwal2024medhalu}
V.~Agarwal, Y.~Jin, M.~Chandra, M.~De~Choudhury, S.~Kumar, and N.~Sastry, ``Medhalu: Hallucinations in responses to healthcare queries by large language models,'' \emph{arXiv:2409.19492}, 2024.

\bibitem{lu2024ai}
C.~Lu, C.~Lu, R.~T. Lange, J.~Foerster, J.~Clune, and D.~Ha, ``The ai scientist: Towards fully automated open-ended scientific discovery,'' \emph{arXiv preprint arXiv:2408.06292}, 2024.

\bibitem{agrawal2024can}
G.~Agrawal, T.~Kumarage, Z.~Alghamdi, and H.~Liu, ``Can knowledge graphs reduce hallucinations in llms?: A survey,'' in \emph{NAACL}, 2024, pp. 3947--3960.

\bibitem{nakano2021webgpt}
R.~Nakano, J.~Hilton, S.~Balaji, J.~Wu, L.~Ouyang, C.~Kim, C.~Hesse, S.~Jain, V.~Kosaraju, W.~Saunders \emph{et~al.}, ``Webgpt: Browser-assisted question-answering with human feedback,'' \emph{arXiv preprint arXiv:2112.09332}, 2021.

\bibitem{gao2023enabling}
T.~Gao, H.~Yen, J.~Yu, and D.~Chen, ``Enabling large language models to generate text with citations,'' in \emph{EMNLP}, 2024.

\bibitem{wangself}
X.~Wang, J.~Wei, D.~Schuurmans, Q.~V. Le, E.~H. Chi, S.~Narang, A.~Chowdhery, and D.~Zhou, ``Self-consistency improves chain of thought reasoning in language models,'' in \emph{ICLR}, 2023.

\bibitem{zhou2023codebertscore}
S.~Zhou, U.~Alon, S.~Agarwal, and G.~Neubig, ``Codebertscore: Evaluating code generation with pretrained models of code,'' in \emph{EMNLP}, 2023, pp. 13\,921--13\,937.

\bibitem{wang2023execution}
Z.~Wang, S.~Zhou, D.~Fried, and G.~Neubig, ``Execution-based evaluation for open-domain code generation,'' in \emph{EMNLP}, 2023, pp. 1271--1290.

\bibitem{zhudyval}
K.~Zhu, J.~Chen, J.~Wang, N.~Z. Gong, D.~Yang, and X.~Xie, ``Dyval: Dynamic evaluation of large language models for reasoning tasks,'' in \emph{ICLR}, 2024.

\bibitem{zhu2024dynamic}
K.~Zhu, J.~Wang, Q.~Zhao, R.~Xu, and X.~Xie, ``Dynamic evaluation of large language models by meta probing agents,'' in \emph{ICML}.\hskip 1em plus 0.5em minus 0.4em\relax PMLR, 2024, pp. 62\,599--62\,617.

\bibitem{yi2023unpacking}
X.~Yi, J.~Yao, X.~Wang, and X.~Xie, ``Unpacking the ethical value alignment in big models,'' \emph{arXiv preprint arXiv:2310.17551}, 2023.

\bibitem{wang2023evaluating}
X.~Wang, L.~Jiang, J.~Hernandez-Orallo, D.~Stillwell, L.~Sun, F.~Luo, and X.~Xie, ``Evaluating general-purpose ai with psychometrics,'' \emph{arXiv preprint arXiv:2310.16379}, 2023.

\bibitem{ChatArena}
Y.~Wu, Z.~Jiang, A.~Khan, Y.~Fu, L.~Ruis, E.~Grefenstette, and T.~Rocktäschel, ``Chatarena: Multi-agent language game environments for large language models,'' 2023.

\bibitem{yao2024value}
J.~Yao, X.~Yi, Y.~Gong, X.~Wang, and X.~Xie, ``Value fulcra: Mapping large language models to the multidimensional spectrum of basic human value,'' in \emph{NAACL}, 2024, pp. 8754--8777.

\bibitem{nguyen2024large}
V.~C. Nguyen, M.~Taher, D.~Hong, V.~K. Possobom, V.~T. Gopalakrishnan, E.~Raj, Z.~Li, H.~J. Soled, M.~L. Birnbaum, S.~Kumar \emph{et~al.}, ``Do large language models align with core mental health counseling competencies?'' in \emph{NAACL}, 2025.

\end{thebibliography}

\end{document}